\documentclass[10pt,twocolumn,letterpaper]{article}

\usepackage{cvpr}              

\newcommand{\taskname}[1]{Event-aided Domain Generalization}
\newcommand{\taskacronym}[1]{E-DG}
\newcommand{\methodname}[1]{Privileged Event-based Predictive Regularization}
\newcommand{\method}[1]{PEPR}
\usepackage{graphicx} 
\usepackage{tikz}     

\usepackage{multirow}
\usepackage{multicol}
\usepackage{array}
\newcolumntype{H}{>{\setbox0=\hbox\bgroup}c<{\egroup}@{}}
\usepackage{makecell}
\usepackage{xcolor}
\newcommand{\highlight}{\rowcolor{cyan!17}}
\usepackage[table]{xcolor}
\usepackage[accsupp]{axessibility}  
\definecolor{cvprblue}{rgb}{0.21,0.49,0.74}
\usepackage[pagebackref,breaklinks,colorlinks,allcolors=cvprblue]{hyperref}
\usepackage{arydshln}

\def\paperID{11132} 
\def\confName{CVPR}
\def\confYear{2026}

\titleheader{\darkgrayed{This paper has been accepted for publication at the\\IEEE Conference on Computer Vision and Pattern Recognition (CVPR) Findings, Denver, 2026
\copyright IEEE}}

\title{PEPR: Privileged Event-based Predictive Regularization for Domain Generalization}

\author{Gabriele Magrini\\
University of Florence\\
Florence, Italy\\
{\tt\small gabriele.magrini@unifi.it}
\and
Federico Becattini\\
University of Siena\\
Siena, Italy\\
{\tt\small federico.becattini@unisi.it}
\and
Niccolò Biondi\\
University of Trento\\
Trento, Italy\\
{\tt\small niccolo.biondi@unitn.it}
\and
Pietro Pala\\
University of Florence\\
Florence, Italy\\
{\tt\small pietro.pala@unifi.it}
}

\begin{document}
\maketitle

\begin{abstract}
Deep neural networks for visual perception are highly susceptible to domain shift, which poses a critical challenge for real-world deployment under conditions that differ from the training data.
To address this domain generalization challenge, we propose a cross-modal framework under the learning using privileged information (LUPI) paradigm for training a robust, single-modality RGB model.
We leverage event cameras as a source of privileged information, available only during training.
The two modalities exhibit complementary characteristics: the RGB stream is semantically dense but domain-dependent, whereas the event stream is sparse yet more domain-invariant.
Direct feature alignment between them is therefore suboptimal, as it forces the RGB encoder to mimic the sparse event representation, thereby losing semantic detail.
To overcome this, we introduce \methodname{} (\method{}), which reframes LUPI as a predictive problem in a shared latent space.
Instead of enforcing direct cross-modal alignment, we train the RGB encoder with \method{} to predict event-based latent features, distilling robustness without sacrificing semantic richness.
The resulting standalone RGB model consistently improves robustness to day-to-night and other domain shifts, outperforming alignment-based baselines across object detection and semantic segmentation.
\end{abstract}


\begin{figure}
    \centering
    \begin{subfigure}{\linewidth}
        \includegraphics[width=\linewidth]{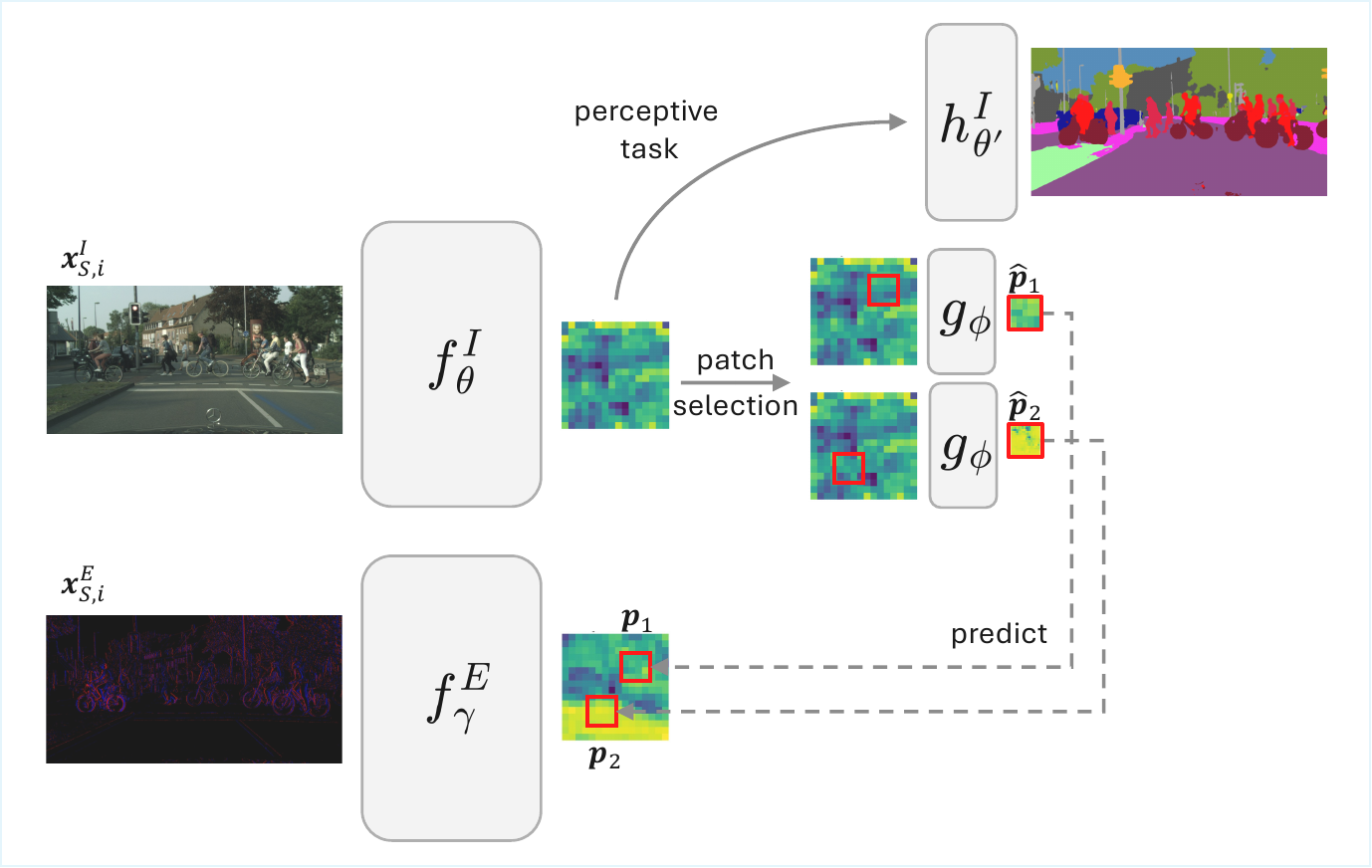}
        \caption{ } \label{fig:method_train}
    \end{subfigure}
    \begin{subfigure}{\linewidth}
        \includegraphics[width=.9\linewidth]{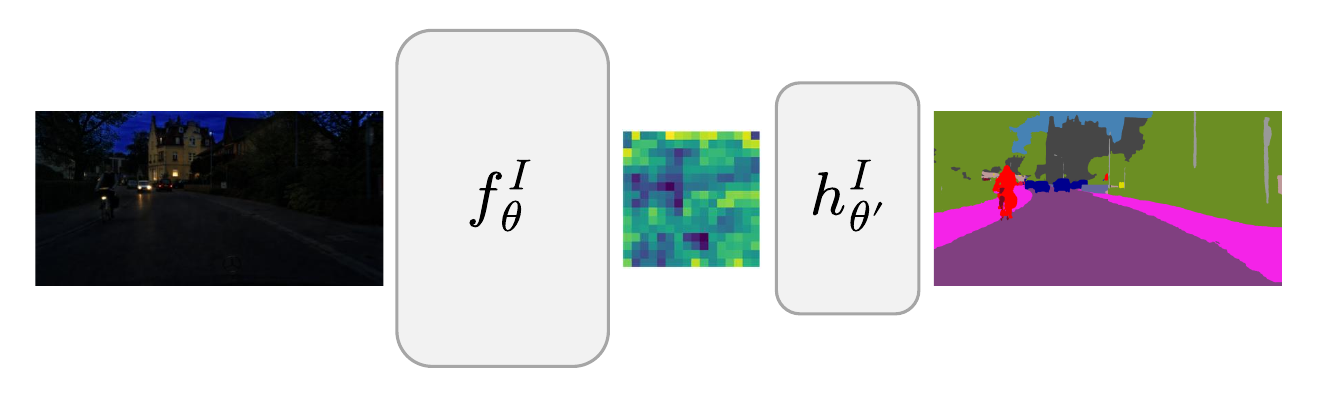}
        \caption{ } \label{fig:method_test_inference}
    \end{subfigure}
    \caption{
Overview of our \methodname{} (\method{}) framework. 
(a) During training, we employ a predictive objective which constrains the RGB encoder to learn representations that can predict the latent features ($\hat{\mathbf{p}}$) from the privileged event encoder ($\mathbf{p}$). 
(b) At inference, the event encoder and predictors are discarded, resulting in a robust, single-modality RGB model for Domain Generalization.
}
\label{fig:method}
\end{figure}

\section{Introduction}
\label{sec:intro}

Neural networks achieve state-of-the-art results across many vision tasks, but typically assume that training and test data are drawn from the same distribution. When this assumption breaks, performance degrades, a phenomenon known as domain shift~\cite{zhou2022domain}. Unsupervised domain adaptation (UDA)~\cite{li2024comprehensive, fang2024source} adapts models using unlabeled target data, yet many real deployments lack access to target data during training. The objective then becomes domain generalization~\cite{zhou2022domain,wang2022generalizing}: learn from a source domain a model that is robust to unseen shifts.
A frequent and critical instance of domain shift for computer vision models is the day-to-night transition~\cite{zhou2022domain}. Public datasets are predominantly daytime, and RGB cameras degrade in low light: visual cues attenuate or vanish, and motion blur increases. Alternative sensors can mitigate low-light issues (e.g., IR detects reflected or emitted radiation; thermal measures heat), but their appearance statistics still change across day and night and typically require additional target-domain labeling.

Event cameras, instead, exhibit properties that reduce sensitivity to day-to-night change: high dynamic range (about 140 dB versus about 60 dB for standard RGB), responses tied to local brightness changes rather than absolute intensity, and negligible motion blur due to microsecond latency. The resulting motion-centric signal is often more reliable at night than color or texture, and prior work reports improved nighttime resilience of event-based models over RGB~\cite{magrini2025fred}.
Motivated by these properties, we leverage event data as privileged information within the Learning Using Privileged Information (LUPI) paradigm~\cite{vapnik2009new,lapin2014learning,ganin2016domain} to train a robust, RGB-only model for deployment (see Fig.~\ref{fig:method_train}). In LUPI, the model has access during training to additional variables that are not available at test time (see Fig.~\ref{fig:method_train}). These privileged variables serve as a teacher or side channel that constrains the hypothesis space, guiding the model toward more robust representations than those obtainable from the standard inputs alone. Here, the privileged channel is the event stream; the deployed model processes only RGB.

The two modalities are complementary: RGB is semantically rich yet domain-dependent, whereas events are sparse yet comparatively domain-invariant. Direct cross-modal alignment is therefore counterproductive, as it encourages the RGB encoder to mimic a sparse representation and discard task-critical semantics.
We instead formulate knowledge transfer as prediction. We introduce \methodname{} (\method{}), which reframes the LUPI paradigm as a predictive problem in a shared latent space. This predictive objective preserves high-level invariance while avoiding the pitfalls of direct feature matching or pixel-level reconstruction \cite{lecun2023introduction,assran2023self,assran2025vjepa,balestriero2025lejepa}. In our setting, an event-based teacher produces event latents, and an RGB student predicts these latents in the shared space. 
This predictive objective requires the RGB representation to contain sufficient information to explain the domain-invariant event signal---for example, to validate motion cues or suppress spurious, illumination-dependent artifacts---without being forced to discard its own modality-specific features. This distills the robustness of the event stream without sacrificing the richness of the RGB stream, resulting in a standalone RGB model with significantly improved robustness and generalization to adverse domains.

The resulting standalone RGB model consistently improves robustness to day-to-night and other domain shifts, outperforming alignment-based baselines across object detection (Hard-DSEC-DET, FRED) and semantic segmentation (Cityscapes-Adverse, Dark Zurich).

In summary, our main contributions are the following:
\begin{itemize}
    \item We formulate cross-modal LUPI as a predictive knowledge transfer problem, avoiding the semantic loss induced when a dense, domain-dependent modality (RGB) is forced to match a sparse, domain-invariant one (events).
    \item We propose \methodname{} (\method{}), which adapts a predictive objective to LUPI: an RGB student predicts event latents in a shared space, transferring robustness from events without direct feature alignment or input reconstruction.
    \item Across object detection (Hard-DSEC-DET, FRED) and semantic segmentation (Cityscapes-Adverse, Dark Zurich), the resulting standalone RGB model improves robustness to day-to-night domain shifts and consistently outperforms alignment-based alternatives.
\end{itemize}

\section{Related Works}
\label{sec:related}

Robustness to domain shift is often addressed through domain adaptation, where models are trained using labeled data from a source domain together with (labeled or unlabeled) samples from the target domain \cite{ganin2015unsupervised,tzeng2017adversarial,lee2018spigan,vu2019dada}. However, collecting representative target data for every possible shift (e.g., day–night, weather, sensor, or geographic changes) is costly and often infeasible. This has motivated a large body of work on domain generalization, which aims to learn models from one or multiple source domains that perform well on unseen targets without access to target data during training \cite{wang2022generalizing,rafi2024domain,schwonberg2025domain,choe2020empirical}. Most existing approaches rely on data augmentation, feature alignment, or domain-invariant representations to mitigate distribution shifts \cite{zhou2022domain,wang2022generalizing}.
We approach robustness to domain shift through Learning Using Privileged Information (LUPI), using a complementary modality only at training time to shape the representation learned from RGB images.

LUPI is a training paradigm in which a model has access to privileged information only during training, but not at test time \cite{vapnik2009new}. This privileged data can be additional sensor modalities, richer annotations, simulation internals, or future observations, and it is exploited to improve performance on the standard input. LUPI was introduced as a new learning paradigm in \cite{vapnik2009new} and initially instantiated as SVM+ for margin-based models \cite{lapin2014learning}. Subsequent works have extended LUPI to deep neural networks \cite{lopez2015unifying, motiian2016information, sarafianos2016predicting, sharmanska2013learning}, for example by distilling teacher predictions based on privileged inputs, predicting privileged features from the standard modality, or using privileged information as auxiliary supervision or regularization. LUPI has also been explored for domain adaptation and domain generalization, where privileged supervision encourages models to focus on causal or geometry-aware cues that transfer across domains \cite{motiian2016information, lee2018spigan}.

In this work, we adopt a different perspective on LUPI by framing it as a predictive objective within a joint-embedding predictive architecture (JEPA).
Joint-embedding methods learn representations by bringing different views of the same sample close in a shared latent space, typically via contrastive or cosine-similarity losses \cite{chen2020simple, grill2020bootstrap, caron2021emerging}. JEPA-style approaches instead predict one latent representation from another, without reconstructing pixels, by masking parts of the input and modeling a conditional distribution in the representation space \cite{lecun2023introduction,assran2023self, assran2025vjepa,balestriero2025lejepa}. This predictive formulation has been shown to produce more semantic and invariant features than pure alignment or generative objectives. We adopt this JEPA view for LUPI: rather than simply aligning RGB and event encoders or reconstructing raw data, we train an RGB encoder whose latent representation is predictive of the event encoder’s latent representation. The event modality is thus used as a richer target for a JEPA-style prediction task, encouraging the RGB encoder to internalize the complementary information provided by events without requiring them at test time.

Event and RGB cameras provide complementary signals: RGB frames capture appearance, while events encode asynchronous brightness changes with high temporal resolution and dynamic range. Several works exploit this complementarity for detection and segmentation by fusing RGB and event streams via early, mid-, or late fusion \cite{zhang2025exploring,shi2025sparse,liu2025beyond,berlincioni2024feature}. These approaches improve robustness under high-speed motion, low light, or motion blur, but require both modalities at inference, limiting deployment when event sensors are unavailable. In parallel, event-based semantic segmentation (ESS) has emerged as a standalone task, predicting semantic labels from events alone~\cite{alonso2019ev,sun2022ess,kong2024openess}. Some ESS methods leverage RGB during training by transferring semantic knowledge from labeled RGB frames to unlabeled events via unsupervised domain adaptation or cross-modal alignment \cite{alonso2019ev,sun2022ess}, and recent frameworks combine image, text, and event modalities for open-vocabulary or annotation-efficient ESS \cite{kong2024openess,zhao2025eseg}. In all these cases, RGB acts as an auxiliary signal to improve event-based models, and events remain the primary inference modality.
Our work differs from prior multimodal RGB–event fusion and ESS approaches in two key aspects. First, we treat events as privileged information: they are used only during training and are not required at inference, matching scenarios where RGB cameras are ubiquitous while event sensors are available only in limited settings. Second, we cast this privileged supervision as a JEPA-style predictive objective in latent space: the RGB encoder is trained so that its embeddings predict those of an event encoder, instead of directly fusing modalities at feature or pixel level. This design exploits the complementary characteristics of RGB and events for domain generalization in segmentation and detection, without changing the test-time architecture or requiring additional sensors. 

To the best of our knowledge, no prior work jointly leverages LUPI, JEPA-style latent prediction, and RGB–event modeling for domain generalization in perceptive tasks.

\section{PEPR: Privileged Event-based Predictive Regularization}
\label{sec:method}

In this section, we present our \methodname{} (\method{}) method with which we address the domain generalization task by training a robust, single-modality RGB model that leverages privileged information from an event camera during the training phase. The goal is to distill the domain-invariant properties of the event stream into the RGB encoder, which is simultaneously optimized for a perception task, such as object detection or semantic segmentation. A high-level overview of our framework is presented in Fig.~\ref{fig:method}.

Our framework, illustrated in Fig.~\ref{fig:method_train}, consists of an RGB encoder $f_{\theta}^I$, a task-specific head $h_{\theta'}^I$, a privileged event encoder $f_{\gamma}^E$, and a predictor $g_\phi$. The RGB encoder $f_{\theta}^I$ and its corresponding head $h_{\theta'}^I$ constitute the final inference model $(f_{\theta}^I, h_{\theta'}^I)$, as shown in Fig.~\ref{fig:method_test_inference}. The event encoder $f_{\gamma}^E$ and the predictor $g_\phi$ are discarded at inference.
This design ensures that our method introduces no additional computational overhead or data dependency at test time, i.e., it processes only the target-domain RGB input $\mathbf{x}_T^I$, while being more robust to domain shifts, such as day-to-night transition. 
The model is trained end-to-end by jointly minimizing a supervised task loss, $\mathcal{L}_{\text{task}}$, and our novel predictive regularization loss, $\mathcal{L}_{\text{feat}}$. Our approach to this regularization is inspired by the principles of joint-embedding predictive architectures \cite{lecun2023introduction,assran2023self}, which operate by predicting latent representations rather than aligning them. However, our formulation is fundamentally different, as we adapt this predictive concept to a cross-modal, supervised LUPI framework.
The total loss used in \method{}, $\mathcal{L}_{\text{\method{}}}$, is defined as:
\begin{equation}\label{eq:total_loss_pepr}
\mathcal{L}_{\text{\method{}}} = \lambda_{\text{task}} \mathcal{L}_{\text{task}} + \lambda_{\text{feat}} \mathcal{L}_{\text{feat}}
\end{equation}
where $\lambda_{\text{feat}}$ and $\lambda_{\text{task}}$ balance the two objectives. 

\subsection{Task-Specific Objective}
The $\mathcal{L}_{\text{task}}$ loss is used to train the RGB representation in a discriminative, task-relevant feature space. This objective is applied only to the RGB stream. For a given source-domain (e.g., daylight) input image $\mathbf{x}_{T,i}^I$, the RGB encoder $f_{\theta}^I$ extracts features, which are then passed to the head $h_{\theta'}$ (e.g., a detection or segmentation decoder) to produce a task prediction.
$\mathcal{L}_{\text{task}}$ is the standard supervised loss (e.g., cross-entropy or a detection loss) computed between the model's prediction and the ground-truth labels. This objective is critical as it forces the encoder $f_{\theta}^I$ to learn a rich, non-trivial latent space, thereby serving as an anchor that prevents the representational collapse that can occur in joint-embedding architectures.

\subsection{Event-guided Predictive Regularization}

The loss $\mathcal{L}_{\text{feat}}$ is our core contribution. It regularizes the RGB encoder $f_{\theta}^I$ by making its latent space predictive of the event representation. The objective is designed to distill domain-invariant motion and structure cues from the event stream without enforcing a strict alignment between modalities.
Given a paired RGB–event sample $(\mathbf{x}^I, \mathbf{x}^E)$, the event encoder $f_{\gamma}^E$ produces an event feature map $f_{\gamma}^E(\mathbf{x}^E)$. From this map, we extract $M$ target patches $\{\mathbf{p}_m\}_{m=1}^M$, for example by pooling over small spatial windows. These patches are chosen to cover regions of both high and low event activity, so that the supervision reflects where motion is present and where it is absent. We denote by $\{\pi_m\}_{m=1}^M$ the corresponding spatial locations of the target patches.

The selection of target patches $\mathbf{p}_m$ is a crucial aspect of \method{}. We first compute an event-activity map (for instance, by counting events per pixel over a short temporal window) and then sample locations $\pi_m$ from a distribution that mixes regions with high activity and regions with very low or zero activity. Predicting patches from highly active regions forces the RGB encoder to discover latent correlates of motion and structure even in challenging conditions such as low light or adverse weather. Predicting patches from low-activity regions, where $\mathbf{p}_m$ is near zero, regularizes the RGB encoder against hallucinating false positives from sensor noise, since it learns that the absence of an event signal is itself a strong cue for the absence of objects and motion.
In parallel, the RGB encoder $f_{\theta}^I$ produces its own feature map $f_{\theta}^I(\mathbf{x}^I)$, which we flatten into a sequence of spatial feature vectors that serve as the context for prediction.

The predictor $g_\phi$ is implemented as a Transformer decoder layer \cite{vaswani2017attention}. Each target location $\pi_m$ is associated with a learnable positional embedding $\mathbf{e}(\pi_m)$, which serves as the query for that patch. The decoder attends over the RGB spatial features using these positional queries and produces a corresponding set of predicted patches
$\{\hat{\mathbf{p}}_m\}_{m=1}^M = g_\phi\big(f_{\theta}^I(\mathbf{x}^I), \{\mathbf{e}(\pi_m)\}_{m=1}^M\big)$.
Intuitively, each positional query asks the RGB feature map to produce the latent representation that best explains the event information at the corresponding region.
The predictive loss $\mathcal{L}_{\text{feat}}$ is then defined as the mean squared error between the predicted patches and the event patches:
\begin{equation}\label{eq:loss_feat_pred}
\mathcal{L}_{\text{feat}} = \frac{1}{M} \sum_{m=1}^M \left\| \hat{\mathbf{p}}_m - \mathbf{p}_m \right\|_2^2.
\end{equation}

This predictive formulation, rather than direct alignment, is central to \method{}. The RGB encoder is not forced to mimic the sparsity of the event representation and can retain dense semantic information while still becoming predictive of event-based cues. At the same time, the architecture is inherently robust to representation collapse that can affect joint-embedding methods. The supervised task loss $\mathcal{L}_{\text{task}}$ acts as a strong anchor by forcing the RGB encoder $f_{\theta}^I$ to learn a non-trivial latent space that solves the task, making trivial all-zero representations suboptimal. Given this constraint, the optimizer must also find a meaningful representation for the event encoder $f_{\gamma}^E$ that allows the predictive loss $\mathcal{L}_{\text{feat}}$ to be minimized. The RGB encoder learns a representation rich enough to predict the event signal, thereby retaining dense semantic information while gaining domain robustness through privileged event-based supervision.



\begin{table}[t]
    \caption{Object detection results on Hard-DSEC-DET.}
    \label{tab:dsecdethard}
    \centering
    \resizebox{\columnwidth}{!}{
    \begin{tabular}{lcccc}
    \toprule
        \textbf{Model} & \textbf{Train Mod.} & \textbf{Test Mod.} & \textbf{mAP$_{50:95}$} & \textbf{mAP$_{50}$} \\ 
        \midrule
        EA-DETR \cite{rossi2025event} & RGB+E & RGB+E & 15.3 & 31.6 \\
        DETR \cite{carion2020end} & E & E & 14.6 & 31.5 \\
        DETR \cite{carion2020end} & RGB & RGB & 20.0 & 37.6 \\ 
        \hspace{3pt} with L2 & RGB+E & RGB & 19.2 (\textit{-0.8}) & 40.1 (\textit{+2.5}) \\
        \highlight \hspace{3pt} with \method{} & RGB+E & RGB & \textbf{21.6} (\textit{+1.6}) & \textbf{42.4} (\textit{+4.8}) \\
        \bottomrule
    \end{tabular}
    }
\end{table}

\begin{table}[t]
    \caption{Object detection results on FRED Challenging.}
    \label{tab:fredchallenging}
    \centering
    \resizebox{\columnwidth}{!}{
    \begin{tabular}{lcccc}
        \toprule
        \textbf{Model} & \textbf{Train Mod.} & \textbf{Test Mod.} & \textbf{mAP$_{50:95}$} & \textbf{mAP$_{50}$} \\
        \midrule
            ER-DETR \cite{magrini2025fred} & RGB+E & RGB+E & 27.9 & 75.7 \\
            YOLO \cite{khanam2024yolov11} & E & E & 41.6 & 79.6 \\
            RT-DETR \cite{zhao2024detrs} & E & E & 35.1 & 76.9 \\
            Faster-RCNN \cite{ren2015faster} & E & E & 34.9 & 76.4 \\
            DETR \cite{carion2020end} & E & E & 19.8 & 62.2 \\
            YOLOS \cite{fang2021you} & RGB & RGB & 2.0 & 8.3 \\
            Faster-RCNN \cite{ren2015faster} & RGB & RGB & 4.8 & 16.4 \\
            YOLO \cite{khanam2024yolov11} & RGB & RGB & 6.3 & 18.1 \\
            RT-DETR \cite{zhao2024detrs} & RGB & RGB & \textbf{7.1} & 21.1 \\
            DETR \cite{carion2020end} & RGB & RGB & 4.1 & 15.8 \\
             \hspace{3pt} with L2 & RGB+E & RGB & 5.1 (\textit{+1.0}) & 18.8 (\textit{+3.0}) \\
            \highlight \hspace{3pt} with \method{}  & RGB+E & RGB & 5.7 (\textit{+1.6}) & \textbf{21.2} (\textit{+5.4}) \\
            \bottomrule
    \end{tabular}
    }
\end{table}

\begin{table*}[ht]
    \caption{Object detection results on FRED Day-to-Night.}
    \label{tab:freddaytonight}
\centering
    \resizebox{.9\textwidth}{!}{
        \begin{tabular}{lcccccccc}
            \toprule
             & & & \multicolumn{2}{c}{\textbf{Night}} & \multicolumn{2}{c}{\textbf{Pitch Black}} & \multicolumn{2}{c}{\textbf{Sunset}} \\
            \cmidrule(lr){4-5} \cmidrule(lr){6-7} \cmidrule(lr){8-9}
            \textbf{Model} & \textbf{Train Mod.} & \textbf{Test Mod.} & \textbf{mAP$_{50:95}$} & \textbf{mAP$_{50}$} & \textbf{mAP$_{50:95}$} & \textbf{mAP$_{50}$} & \textbf{mAP$_{50:95}$} & \textbf{mAP$_{50}$} \\
            \midrule
            DETR \cite{carion2020end} & RGB & RGB & 0.4 & 1.7 & 0.9 & 5.0 & 1.2 & 4.0 \\
            \hspace{3pt} with L2 (\textit{Ours})         & RGB+E & RGB & 3.4 (\textit{+3.0}) & 14.5  (\textit{+12.8}) & 0.4  (\textit{-0.5}) & 1.8  (\textit{-3.2}) & 1.4  (\textit{+0.2}) & 4.1  (\textit{+0.1}) \\
            \highlight \hspace{3pt} with \method{}(\textit{Ours})  & RGB+E & RGB & \textbf{5.1} (\textit{+4.7}) & \textbf{22.2}  (\textit{+20.5}) & \textbf{1.2} (\textit{+0.3}) & \textbf{6.1} (\textit{+1.1}) & \textbf{2.7} (\textit{+1.5}) & \textbf{9.2} (\textit{+5.2}) \\
            \bottomrule
        \end{tabular}
    }
\end{table*}

\section{Experiments}


We evaluate the effectiveness of our proposed framework on two challenging computer vision tasks: object detection and semantic segmentation, both under domain shift.
For the task of object detection, we evaluate performance using the standard COCO \cite{lin2014microsoft} evaluation metrics. The primary metric reported is the mean Average Precision (mAP) calculated over 10 Intersection over Union (IoU) thresholds, from 0.50 to 0.95 with a step of 0.05. This metric, denoted as $mAP_{50:95}$, provides a comprehensive assessment of detection and localization quality. In addition, we report the mAP at a single IoU threshold of 0.50, denoted as $mAP_{50}$, which focuses on coarse localization and is commonly used to compare with prior work and to highlight changes in overall detection robustness.
For the task of semantic segmentation, we use the standard metric of mean Intersection over Union (mIoU). The mIoU is computed by calculating the IoU (Jaccard Index) for each semantic class individually and then averaging these scores across all classes present in the dataset. This metric provides a robust measure of segmentation quality across the diverse set of classes. Implementation details are in Appendix~\ref{sec:implementation_details}.

\subsection{Datasets}
To evaluate the proposed approach, we focused on some of the most prominent multimodal datasets for object detection and semantic segmentation.


\textbf{Hard-DSEC-DET. \,}
Hard-DSEC-DET~\cite{rossi2025event}, is a 500-frame benchmark extracted from DSEC-DET \cite{Gehrig24nature}, a multimodal RGB-Event dataset for object detection in urban environments. Scenarios included in Hard-DSEC-DET exhibit dynamic lighting such as tunnel transitions.

\textbf{FRED. \,}
The Florence RGB-Event Drone dataset (FRED) \cite{magrini2025fred} is a 14 hours-long multimodal dataset for drone detection, tracking, and trajectory forecasting.
Recordings are high-resolution ($1280 \times 720$) with spatio-temporally synchronized RGB and Event with RGB at 30 FPS. The data features challenging scenarios like rain, low light, and nighttime.
FRED is organized into two distinct splits.
\textit{FRED Canonical} is a balanced 80/20 partitioning, with challenging scenarios equally distributed across both sets.
\textit{FRED Challenging} is designed to evaluate a model's generalization capabilities by introducing a significant data distribution shift between train and test.

\textbf{FRED Day-to-Night. \,}
We selected subsets of the challenging split of FRED to simulate day-to-night shifts explicitly. We defined three new splits: \textit{Night}, \textit{Pitch Black} and \textit{Sunset}.
\textit{Night} comprises sequences taken at night-time, where some illumination comes from the background, such as lights in the distance. The \textit{Pitch Black} scenario instead contains only subtle changes in illumination, as the scene is completely dark. In this scenario, the drone is clearly visible in the event domain, but almost undetectable in the RGB domain. This is an extremely challenging scenario for RGB-based models. \textit{Sunset} instead contains footage acquired at sunset, posing challenges as illumination changes, high-dynamic range and direct illumination.

\textbf{Cityscapes Adverse. \,}
Cityscape-Adverse~\cite{suryanto2025cityscape} is a benchmark used to evaluate robustness of semantic segmentation models under adverse environmental conditions. It is derived from the Cityscapes dataset \cite{cordts2016cityscapes}, it has 2,975 training and 500 validation images. Eight distinct adverse conditions are simulated with a diffusion model: rainy, foggy, spring, autumn, winter (snow), sunny, night, and dawn. This process creates a total of $2975\times8$ images for training and $500\times8$ images for validation.
We simulate event streams with ESIM \cite{Rebecq18corl}.

\textbf{Dark Zurich. \,}
Dark Zurich \cite{sakaridis2019guided} is a real-world dataset collected by driving several laps multiple times on the same day to capture corresponding images at daytime, twilight, and nighttime.
In this paper, we evaluate domain generalization by training on cityscapes and testing zero-shot on the validation split of Dark Zurich-night (2416 images).


\subsection{Baselines}
\label{sec:baselines}
We compare \method{} to state-of-the-art models for detection and segmentation. Although our approach achieves strong performance on challenging datasets such as FRED, Hard-DSEC-DET, and Dark Zurich, our goal is to show that the proposed training strategy improves robustness to domain shifts for a given architecture, rather than to design new state-of-the-art backbones. Thus, we consider two baselines built on the same RGB architecture $(f_{\theta}^I, h_{\theta'}^I)$ as \method{}. The first is RGB-only, trained using only the supervised task loss $\mathcal{L}_{\text{task}}$ on the source domain. In this setting, the model does not use event data at training or test time, and the total loss reduces to $\mathcal{L}_{\text{RGB}} = \mathcal{L}_{\text{task}}$.
The second is a feature distillation (L2) baseline that follows the learning setting of \method{}, with a privileged event encoder $f_{\gamma}^E$ available only during training. Instead of our predictive loss, we directly align the features of the two modalities on the source domain. Given a paired RGB–event sample $(\mathbf{x}^I, \mathbf{x}^E)$, we compute the corresponding feature maps $f_{\theta}^I(\mathbf{x}^I)$ and $f_{\gamma}^E(\mathbf{x}^E)$ and define the feature loss as the mean squared error between them,
$\mathcal{L}_{\text{feat}}^{\text{L2}} = \big\| f_{\theta}^I(\mathbf{x}^I) - f_{\gamma}^E(\mathbf{x}^E) \big\|_2^2$.
The total loss for this baseline mirrors Eq.~\eqref{eq:total_loss_pepr}, $\mathcal{L}_{\text{L2}} = \lambda_{\text{task}} \mathcal{L}_{\text{task}} + \lambda_{\text{feat}} \mathcal{L}_{\text{feat}}^{\text{L2}}$,
but uses direct feature alignment instead of the event-based latent prediction loss introduced in Sec.~\ref{sec:method}. As in \method{}, the event encoder $f_{\gamma}^E$ is discarded at inference, and only the RGB model $(f_{\theta}^I, h_{\theta'}^I)$ is used at test time.



\newcommand{\fredfigwidth}{0.26\textwidth}

\begin{figure*}[t] 
\centering

\begin{tabular}{ccc}

\textbf{RGB} &
\textbf{L2} &
\textbf{\method{}} \\
\includegraphics[width=\fredfigwidth, trim={10cm 11cm 10cm 4cm},clip]{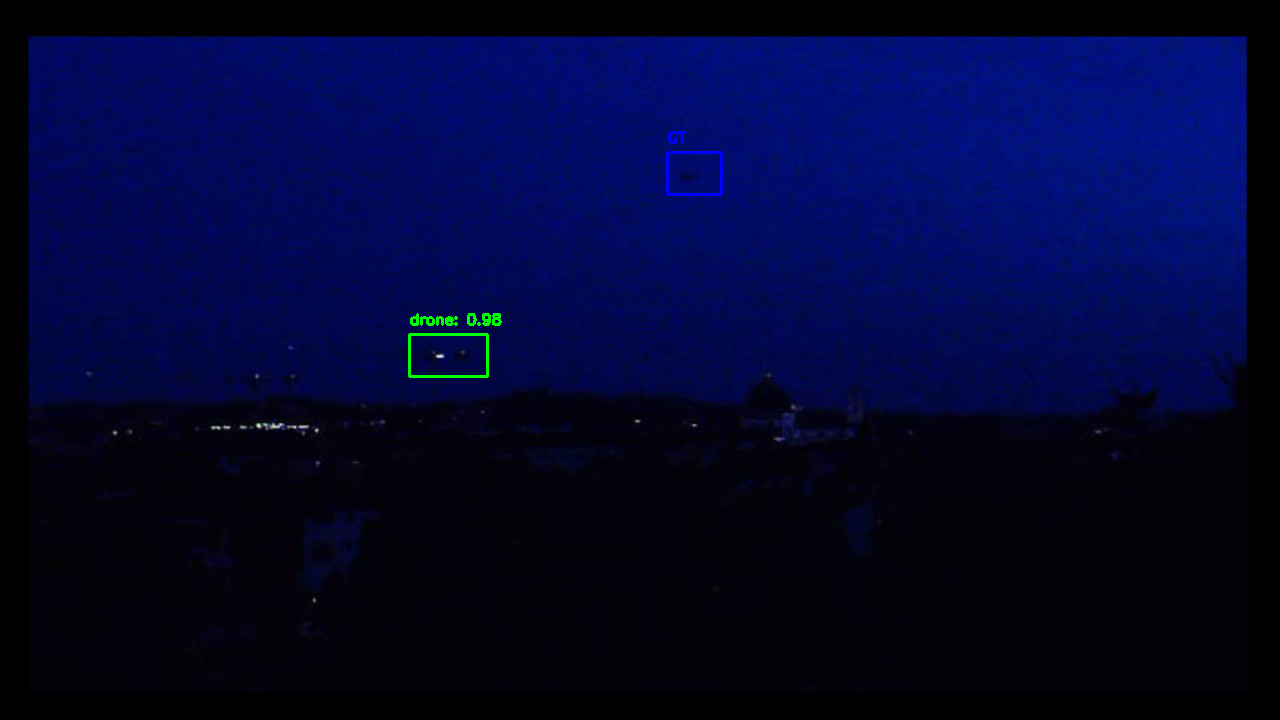} &
\includegraphics[width=\fredfigwidth, trim={10cm 11cm 10cm 4cm},clip]{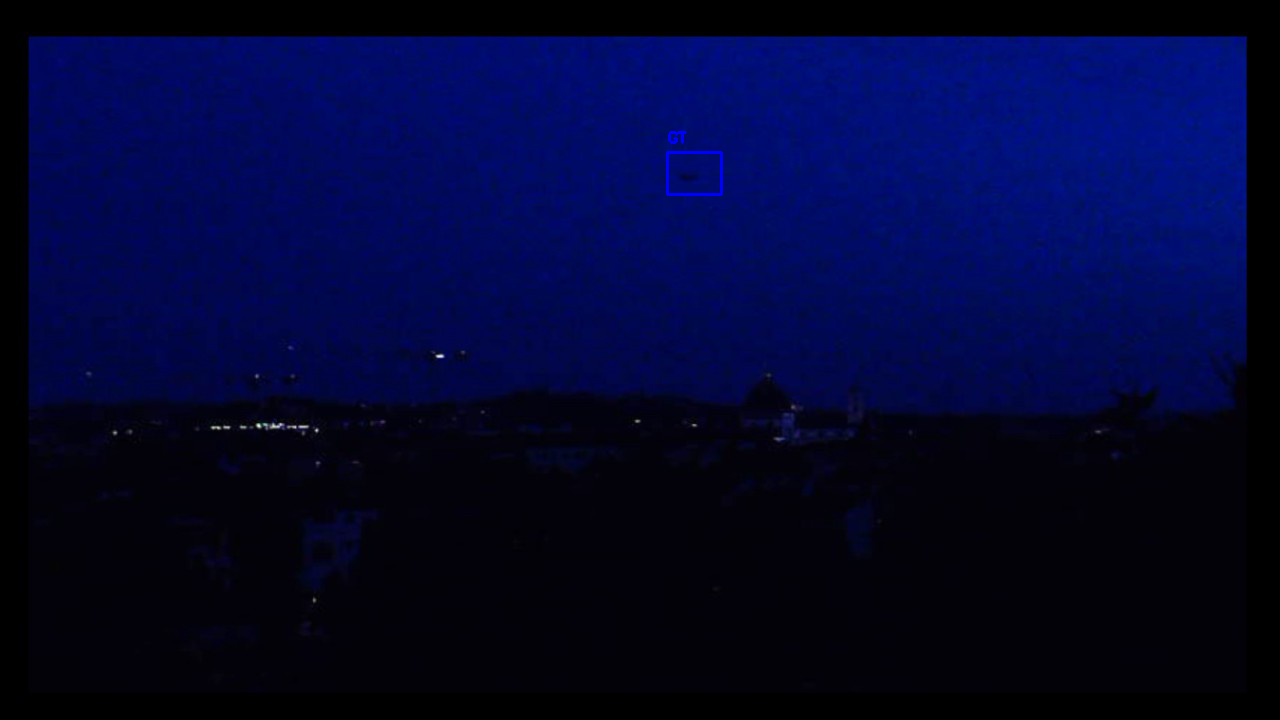} &
\includegraphics[width=\fredfigwidth, trim={10cm 11cm 10cm 4cm},clip]{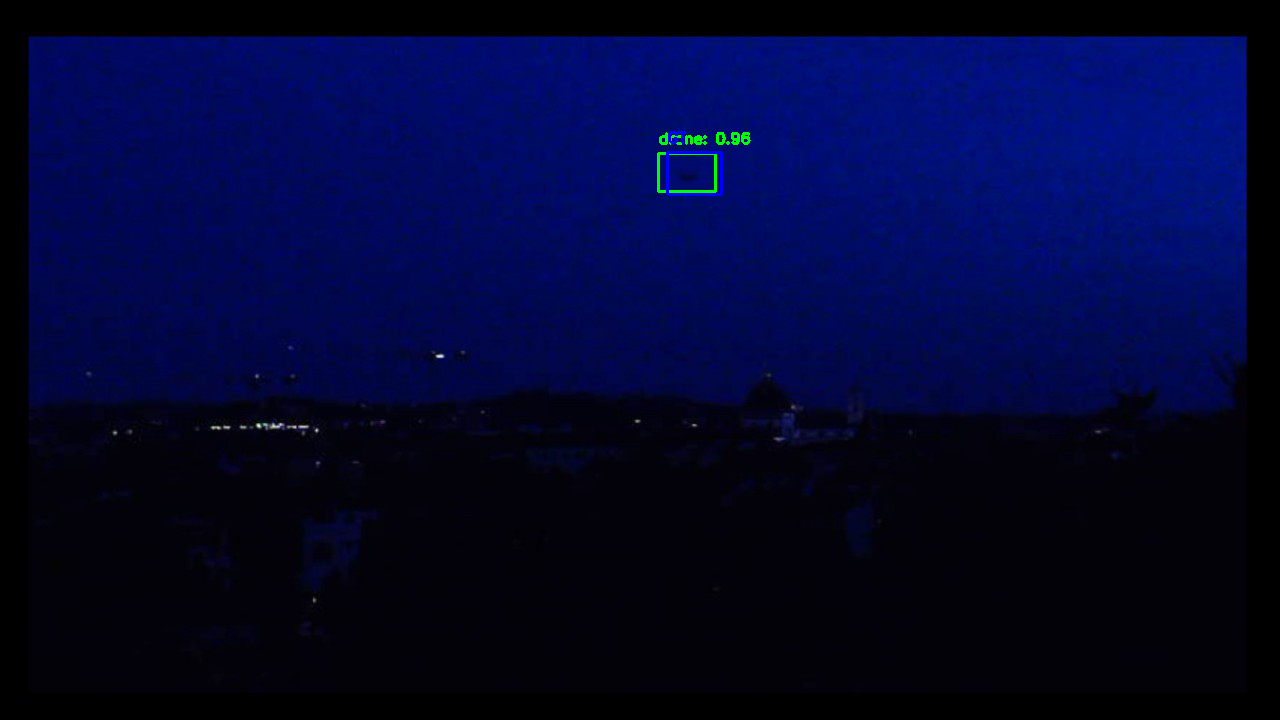} \\


\includegraphics[width=\fredfigwidth, trim={10cm 10cm 10cm 5cm},clip]{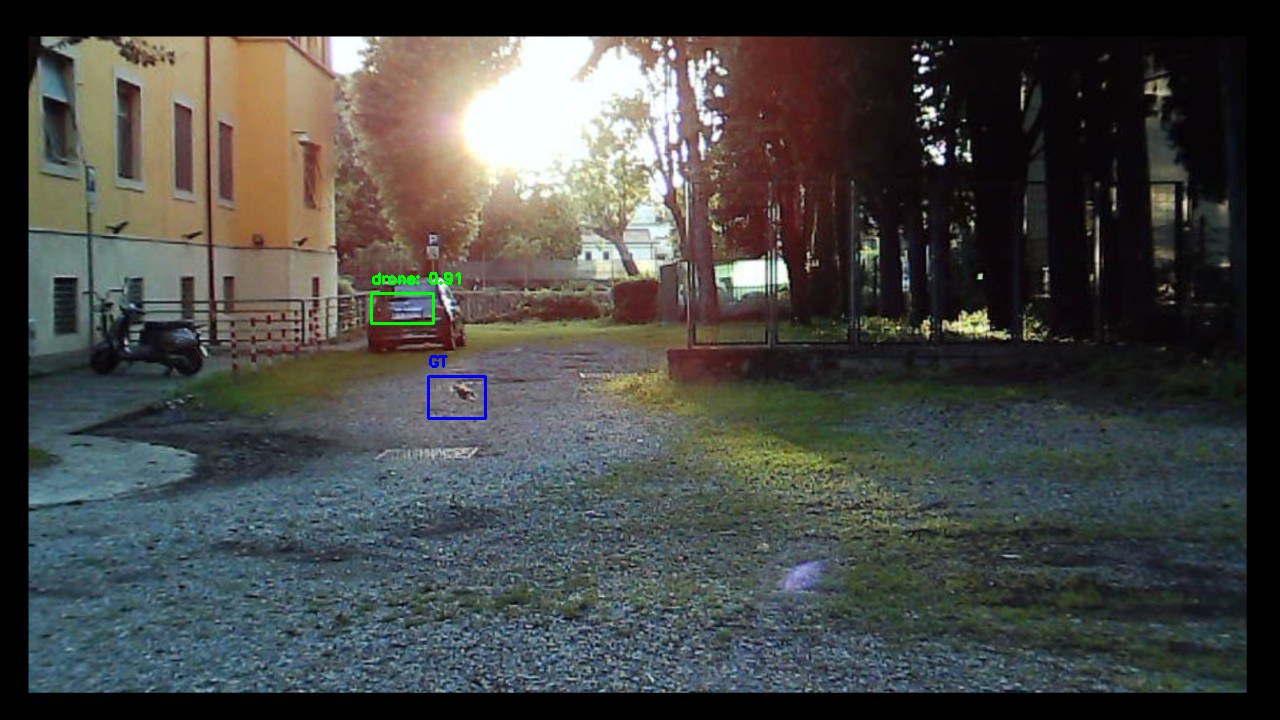} &
\includegraphics[width=\fredfigwidth, trim={10cm 10cm 10cm 5cm},clip]{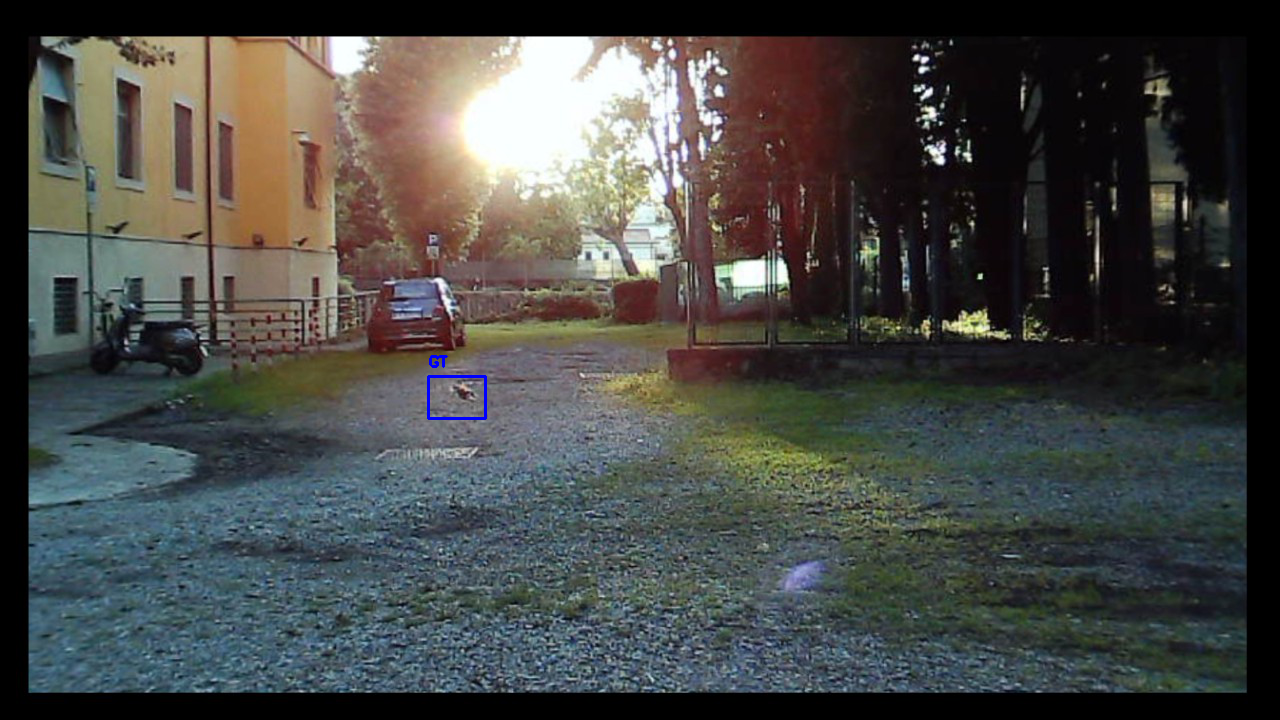} &
\includegraphics[width=\fredfigwidth, trim={10cm 10cm 10cm 5cm},clip]{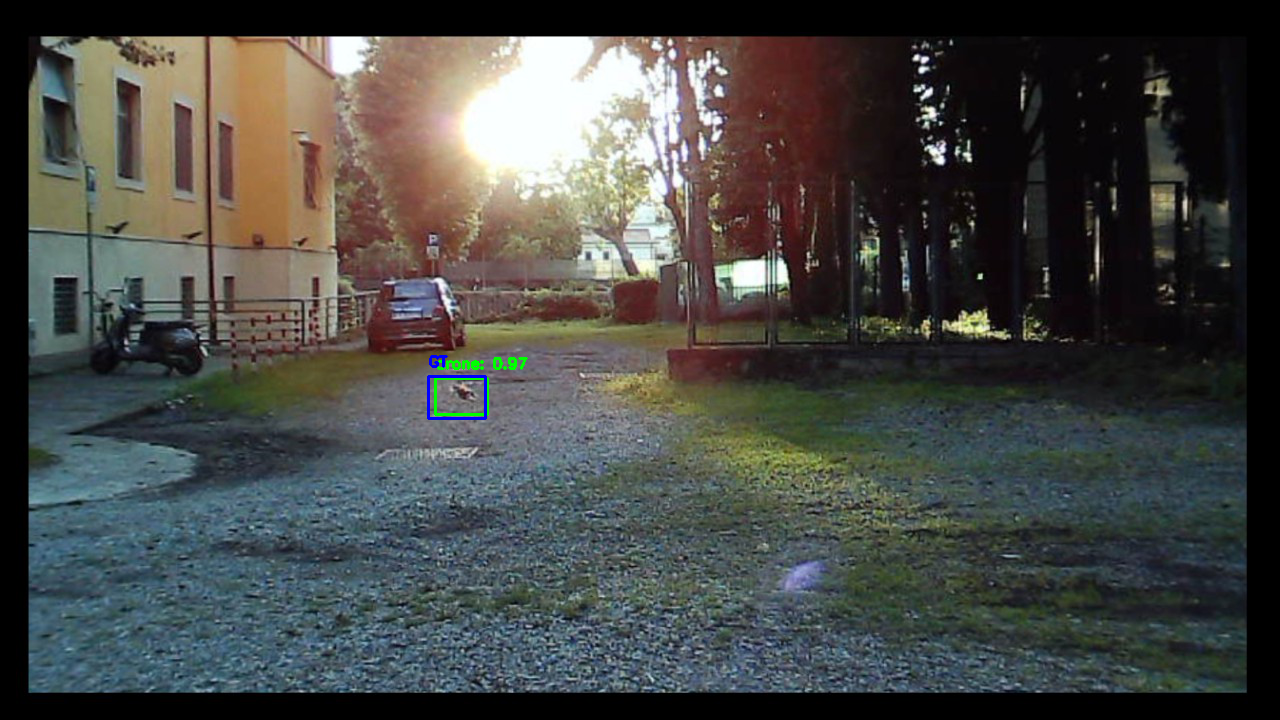} \\


\includegraphics[width=\fredfigwidth, trim={15cm 10cm 5cm 5cm},clip]{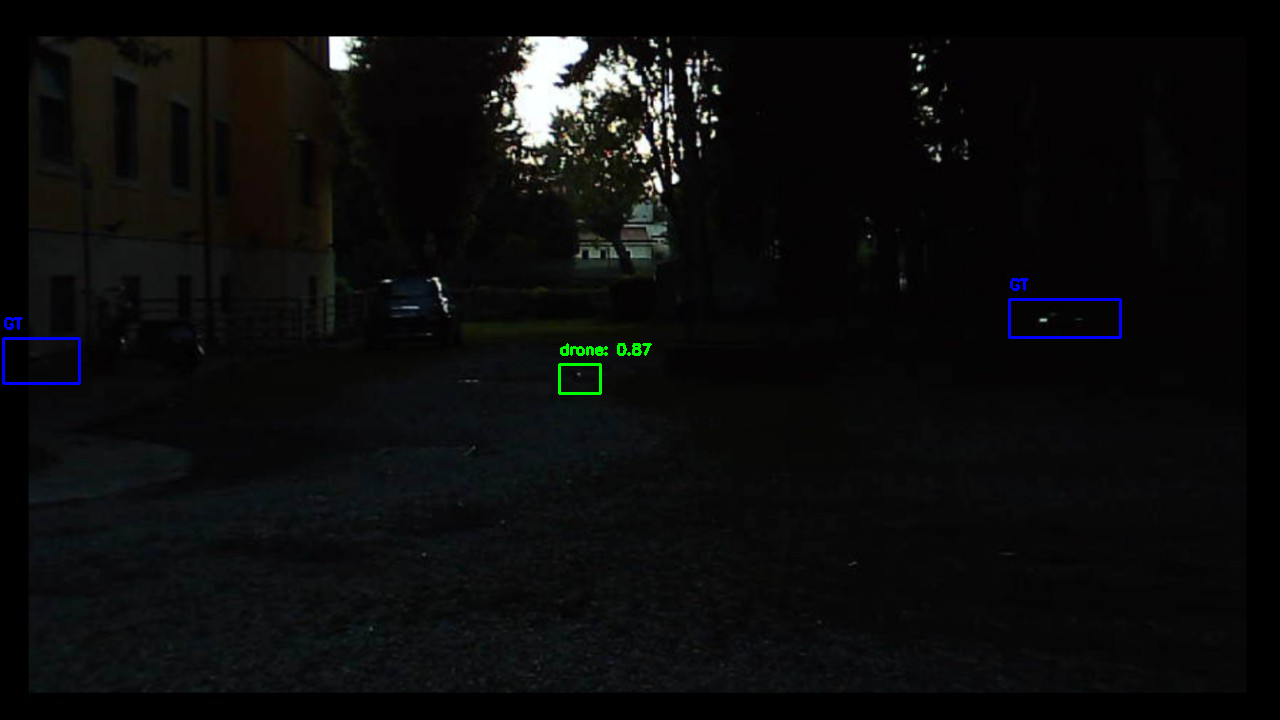} &
\includegraphics[width=\fredfigwidth, trim={15cm 10cm 5cm 5cm},clip]{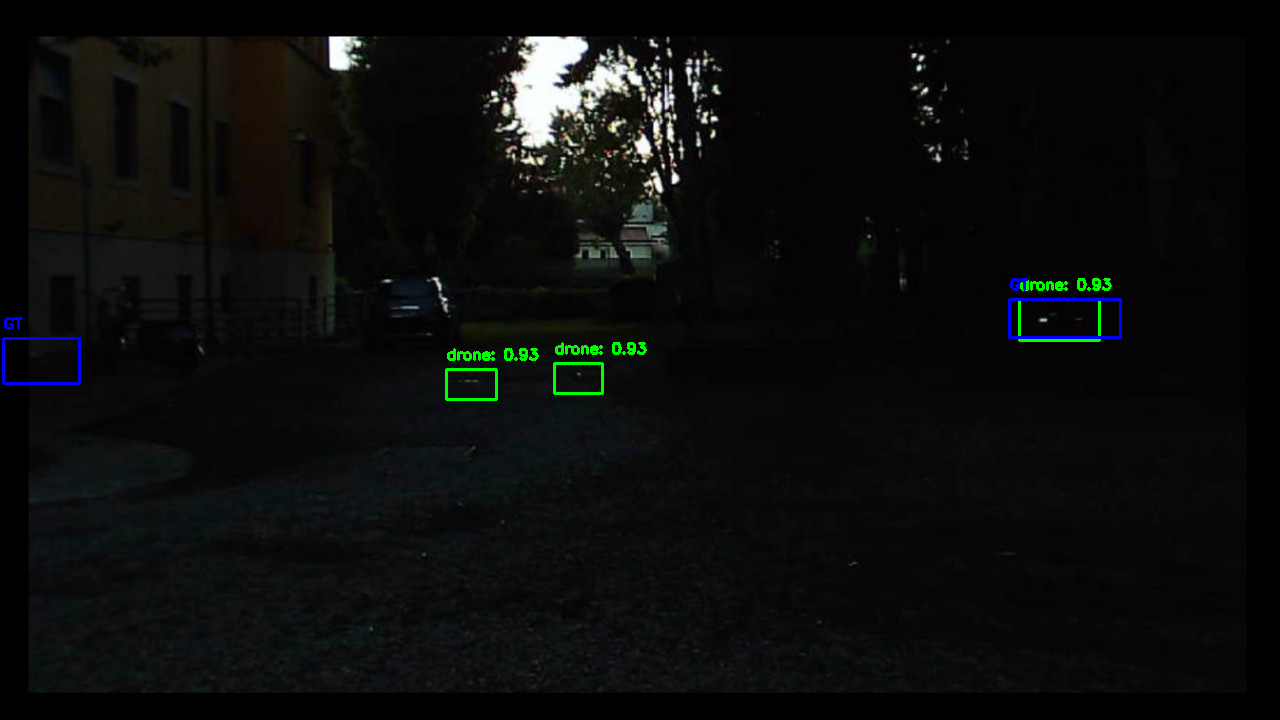} &
\includegraphics[width=\fredfigwidth, trim={15cm 10cm 5cm 5cm},clip]{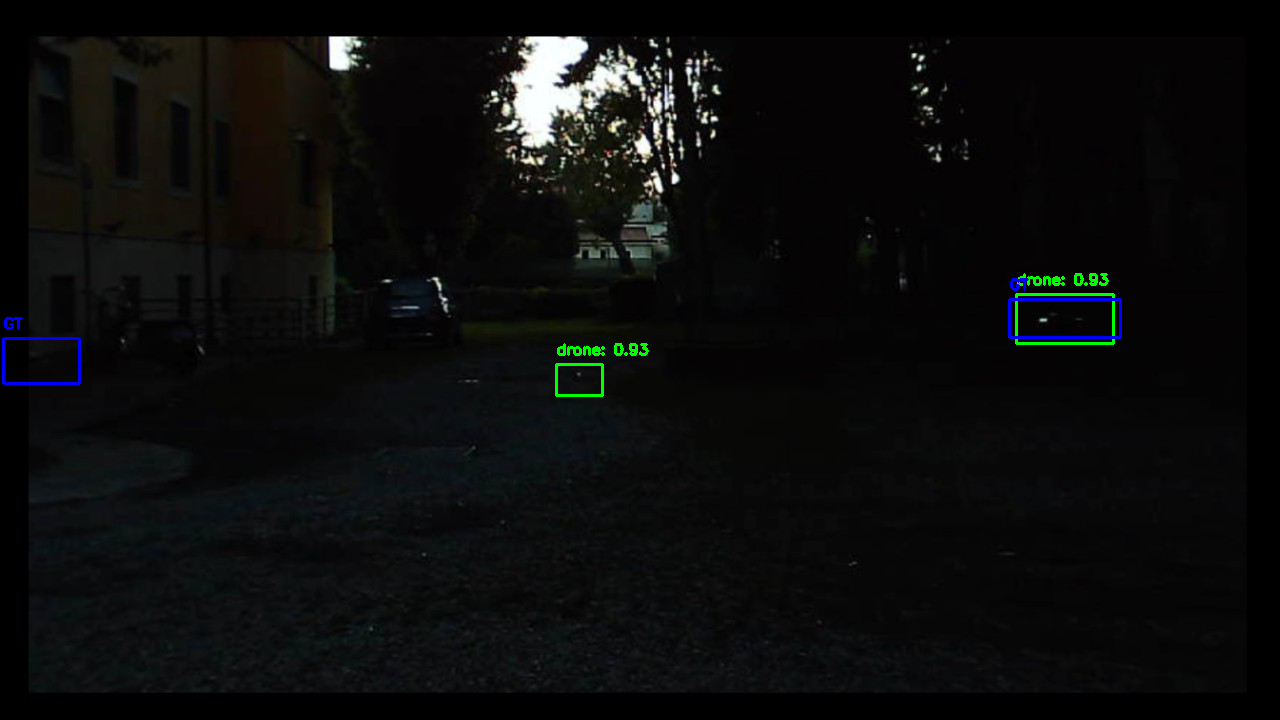} \\

\end{tabular} 

\caption{Qualitative results on the FRED Challenging dataset. \method{} manages to improve the detection rate in adverse unseen conditions, whereas the L2 feature alignment is not as effective. Ground truth boxes are shown in blue, detections in green.}
\label{fig:qualitative_fred_challenging}
\end{figure*}

\begin{table}[t]
    \caption{Object detection results on FRED Canonical.}
    \label{tab:fredcanonical}
    \resizebox{\columnwidth}{!}{
        \begin{tabular}{lcccc}
            \toprule
            \textbf{Model} & \textbf{Train Mod.} & \textbf{Test Mod.} & \textbf{mAP$_{50:95}$} & \textbf{mAP$_{50}$} \\ \midrule
            ER-DETR \cite{magrini2025fred} & RGB+E & RGB+E & 32.2 & 78.6 \\
            YOLO \cite{khanam2024yolov11} & E & E & 49.3 & 87.7 \\
            RT-DETR \cite{zhao2024detrs} & E & E & 39.0 & 82.1 \\
            Faster-RCNN \cite{ren2015faster} & E & E & 39.0 & 85.0 \\
            DETR \cite{carion2020end} & E & E & 21.8 & 68.4 \\
            YOLO \cite{khanam2024yolov11} & RGB & RGB & 13.4 & 35.2 \\
            Faster-RCNN \cite{ren2015faster} & RGB & RGB & 12.3 & 35.1 \\
            RT-DETR \cite{zhao2024detrs} & RGB & RGB & 11.8 & 34.1 \\
            DETR \cite{carion2020end} & RGB & RGB & 7.7 & 28.6 \\
            \hspace{3pt} with L2 & RGB+E & RGB & 11.4 (\textit{+3.7}) & 34.6 (\textit{+6.0}) \\
            \highlight \hspace{3pt} with \method{} & RGB+E & RGB & \textbf{11.9} (\textit{+4.2}) & \textbf{38.2} (\textit{+9.4}) \\
            \bottomrule
        \end{tabular}
    }
\end{table}

\begin{table*}[t]
    \caption{Semantic segmentations results on Cityscapes Adverse.}
    \label{tab:cityscapesadverse}
\centering
    \resizebox{.9\textwidth}{!}{
        \begin{tabular}{lHccccccccc}
            \toprule
             &  & \multicolumn{2}{c}{\textbf{Weathers}} & \multicolumn{3}{c}{\textbf{Seasons}} & \multicolumn{3}{c}{\textbf{Lightings}} & \textbf{Adverse} \\
            \cmidrule(lr){3-4} \cmidrule(lr){5-7} \cmidrule(lr){8-10}
            \textbf{Model} & \textbf{Cityscapes}  & \textbf{Rainy} & \textbf{Foggy} & \textbf{Spring} & \textbf{Autumn} & \textbf{Snow} & \textbf{Sunny} & \textbf{Night} & \textbf{Dawn} & \textbf{Avg.} \\
            \midrule
            DeepLabV3+ \cite{chen2018encoder} & 79.6 & 58.2  & 66.1  & 69.5  & 46.8  & 22.3  & 42.6  & 47.4  & 57.7  & 51.2  \\
            ICNet \cite{zhao2018icnet} & 76.3 & 68.8  & 69.3  & 70.6  & 66.7  & 27.0  & 54.7  & 61.2  & 67.5  & 61.0  \\
            DDRNet \cite{hong2021deep} & 80.0 & 65.6  & 69.3  & 71.9  & 55.1  & 30.2  & 53.5  & 58.9  & 64.8  & 59.3  \\
            SETR \cite{zheng2021rethinking} & 77.6 & 71.8  & 70.6  & 73.2  & 71.3  & 61.4  & 61.9  & 70.3  & 73.0  & 68.0  \\
            Mask2Former \cite{cheng2022masked} & 82.6 & 67.7  & 72.9  & 74.9  & 73.7  & 62.8  & 63.0  & 69.0  & 75.0  & 68.2  \\ 
            SegFormer \cite{xie2021segformer} & 76.5 & \textbf{65.9 } & 65.7  & 69.0  & 65.4  & 46.9  & 57.9  & 63.4  & 66.1  & 61.2  \\
            \hspace{3pt} with L2 & 76.2 & 63.5 (\textit{-2.4}) & 66.9 (\textit{+1.2}) & 69.1 (\textit{+0.1}) & 66.2 (\textit{+0.8}) & 46.3 (\textit{-0.6}) & 57.8 (\textit{-0.1}) & 63.2 (\textit{-0.2}) & 67.5 (\textit{+1.4}) & 62.5 \\
            \highlight \hspace{3pt} with \method{} & 76.6 & 62.8 (\textit{-3.1}) & \textbf{67.8} (\textit{+2.1}) & \textbf{70.2} (\textit{+1.2}) & \textbf{66.6} (\textit{+1.2}) & \textbf{47.1} (\textit{+0.2}) & \textbf{59.0} (\textit{+1.1}) & \textbf{63.9} (\textit{+0.5}) & \textbf{67.7} (\textit{+1.6}) & \textbf{63.1} \\
            \bottomrule
        \end{tabular}
    }
\end{table*}

\subsection{Results}
\label{sec:results}

We compare our approach \method{} against the L2 baseline for object detection and semantic segmentation. We first analyze the benefits of \method{} for domain generalization, and then show that our learning strategy can improve the learned features in both source and target domains.
In our experiments, we apply \method{} to DETR~\cite{carion2020end} for object detection and SegFormer~\cite{xie2021segformer} for semantic segmentation, due to their widespread adoption in the vision community and their simplicity of use. Importantly, \method{} is model-agnostic and can in principle be integrated into any architecture that exposes its feature extractor during training.

\subsubsection{Object Detection}
Tab.~\ref{tab:dsecdethard} presents the results on Hard-DSEC-DET~\cite{rossi2025event}. The RGB-only DETR baseline achieves 20.0 $mAP_{50:95}$. Interestingly, the L2 feature alignment baseline not only fails to provide a benefit but degrades performance to 19.2 $mAP_{50:95}$ (a -0.8 drop). This result strongly supports our central hypothesis that naive, direct alignment of a dense, domain-dependent modality (RGB) to a sparse, domain-invariant one (events) may be suboptimal and can be counterproductive. In contrast, our proposed \method{} framework effectively makes the model more robust, improving performance for both $mAP_{50:95}$ and $mAP_{50}$.

We further test generalization on FRED Challenging ~\cite{magrini2025fred}, which introduces a significant data distribution shift between the training and test sets. As shown in Tab.~\ref{tab:fredchallenging}, the domain shift is severe: the performance of the RGB-only model is extremely limited at 4.1 $mAP_{50:95}$. The L2 baseline provides a modest improvement, showing that even a simple approach like a cross-modal direct feature distillation can make the features more robust. \method{} again yields the strongest result, demonstrating a more effective knowledge transfer than direct alignment. Interestingly, DETR$_{\method{}}$ achieves the best performance across all RGB methods in terms of $mAP_{50}$. In this dataset, our approach proves to be extremely effective due to the fact that in drone detection, the signal-to-noise ratio in the event domain is very high (the drone is clearly visible, whereas the background is almost filtered out) and in the RGB domain is very low (clutter and background make it extremely challenging to spot small objects like drones). This imbalance between the two modalities is shown by the much higher mAPs reported for event-only models. The model largely benefits from our learning framework thanks to the strong guidance provided by the event data, learning to focus on subtle cues that are less evident in the target domain (e.g., at nighttime).
Qualitative results are in Fig. \ref{fig:qualitative_fred_challenging} and Appendix~\ref{sec:qualitative_obj_detection}.



To explicitly analyze performance under the critical day-to-night shift, we evaluate on the three curated splits of the FRED Day-to-Night dataset. Results are shown in Tab. \ref{tab:freddaytonight}.
In the Night split, the standard RGB model is almost unusable, scoring 1.7 $mAP_{50}$. While both L2 and \method{} provide substantial gains by leveraging the privileged event data, \method{}'s improvement is much larger: it achieves 22.2 $mAP_{50}$ (a +20.5 gain) compared to L2's 14.5 $mAP_{50}$ (a +12.8 gain).
The Pitch Black scenario, where the RGB signal is almost non-existent, represents the most extreme test. Here, the L2 alignment baseline fails catastrophically, reducing performance to 1.8 $mAP_{50}$ (a -3.2 drop). This suggests the L2 loss forces the RGB encoder into a degenerate representation in its attempt to match the sparse event signal. \method{}, however, remains robust and is the only method to extract a meaningful signal, improving the baseline. The absolute $mAP$ values, however, are still extremely low as in most of the sequence the signal is barely visible.
In the Sunset split, characterized by high dynamic range and illumination changes with respect to daylight videos, \method{} provides a +5.2 $mAP_{50}$ improvement, whereas the L2 baseline offers only a marginal +0.1 $mAP_{50}$ gain.
These results demonstrate that how the privileged information is utilized is critical. \method{}'s predictive objective is a stable and effective regularizer, successfully distilling domain-invariant cues even in extreme low-light conditions where direct feature matching breaks down.

A critical consideration concerns the effect of the proposed domain generalization strategy on in-domain performance. An overly aggressive regularizer might improve generalization at the cost of performance on the source distribution. We evaluate this by testing on the FRED Canonical split (Tab.~\ref{tab:fredcanonical}), where the test data is drawn from the same distribution as the training data.
The RGB-only DETR baseline achieves 7.7 $mAP_{50:95}$. The L2 alignment baseline improves this score to 11.4 $mAP_{50:95}$ (a +3.7 gain). Our \method{} model achieves the highest score of 11.9 $mAP_{50:95}$, a +4.2 gain over the baseline.
This result demonstrates two key points: (i) far from hindering performance, leveraging privileged event data acts as a powerful regularizer that improves performance even on the source domain, and (ii) \method{}'s predictive objective serves as a more effective regularizer than naive L2 alignment, leading to a stronger in-domain model in addition to its superior generalization capabilities.
Please refer to Appendix~\ref{sec:dsec_det} for in-domain results on the DSEC-DET dataset \cite{Gehrig24nature}.





\subsubsection{Semantic Segmentation}
We perform experiments for semantic segmentation by first training on Cityscapes, which contains only daylight images, and then evaluating zero-shot on Cityscapes Adverse and Dark Zurich.
Results for Cityscapes Adverse are shown in Tab. \ref{tab:cityscapesadverse}. The baseline RGB-only SegFormer achieves an average mIoU of 61.2\% across all adverse conditions. The direct L2 alignment baseline slightly improves to 62.5\% mIoU (+1.3), indicating some benefit from learning with privileged information. However, our PEPR framework achieves the highest score of 63.1\% mIoU (+1.9), demonstrating a superior ability to leverage the privileged data.
Notably, PEPR shows consistent gains across most conditions. The L2 baseline instead proves unstable, degrading performance in several conditions. This suggests that even when trained on synthetic events, PEPR's predictive objective provides a more stable and effective regularization against domain shifts compared to direct feature matching.
Qualitative results are shown in Fig. \ref{fig:qualitative_cityscapes} and Appendix~\ref{sec:qualitative_segmentation}.



\begin{table}[t]
\caption{Semantic segmentations results on Dark Zurich.}
\label{tab:darkzurich}
\centering
\resizebox{\columnwidth}{!}{
\begin{tabular}{lccccH}
\toprule
\textbf{Method} & \textbf{Train Mod.} & \textbf{Test Mod.} & \textbf{aAcc} & \textbf{mIoU} & \textbf{mAcc} \\
\midrule
DeepLab-v2 \cite{chen2017deeplab} & RGB & RGB & - & 12.1 & - \\
PSPNet \cite{zhao2017pyramid} & RGB & RGB& - & 12.3 & - \\
RefineNet \cite{lin2017refinenet} & RGB & RGB& - & 15.2 & - \\
SegFormer \cite{xie2021segformer} & RGB & RGB &  57.5 & 19.6 & 34.8\\
\hspace{3pt} with L2 & RGB+E & RGB & 53.2 (\textit{-4.3}) & 17.7 (\textit{-1.9}) & 34.5 \\
\highlight \hspace{3pt} with \method{} & RGB+E & RGB & \textbf{58.6} (\textit{+1.1}) & \textbf{20.3} (\textit{+0.7}) & 33.7\\
\bottomrule
\end{tabular}
}
\end{table}

We further validate our approach on the Dark Zurich-night validation set, which represents a challenging real-world day-to-night domain shift. The results in Tab. \ref{tab:darkzurich}. clearly highlight the limitations of naive feature alignment, as the L2 alignment baseline suffers a catastrophic performance degradation. This result strongly supports our hypothesis that forcing a dense RGB encoder to directly mimic a sparse event representation is counterproductive, especially under severe domain shifts where the modality characteristics diverge. In contrast, our \method{}-trained model not only remains robust but improves performance.

\subsection{Ablation Studies}

\newcommand{\cityfigwidth}{0.21\textwidth}

\begin{figure*}[t]
\centering

\begin{tabular}{cccc}

\textbf{RGB} &
\textbf{SegFormer} &
\textbf{\method{}} &
\textbf{Label} \\

\includegraphics[width=\cityfigwidth]{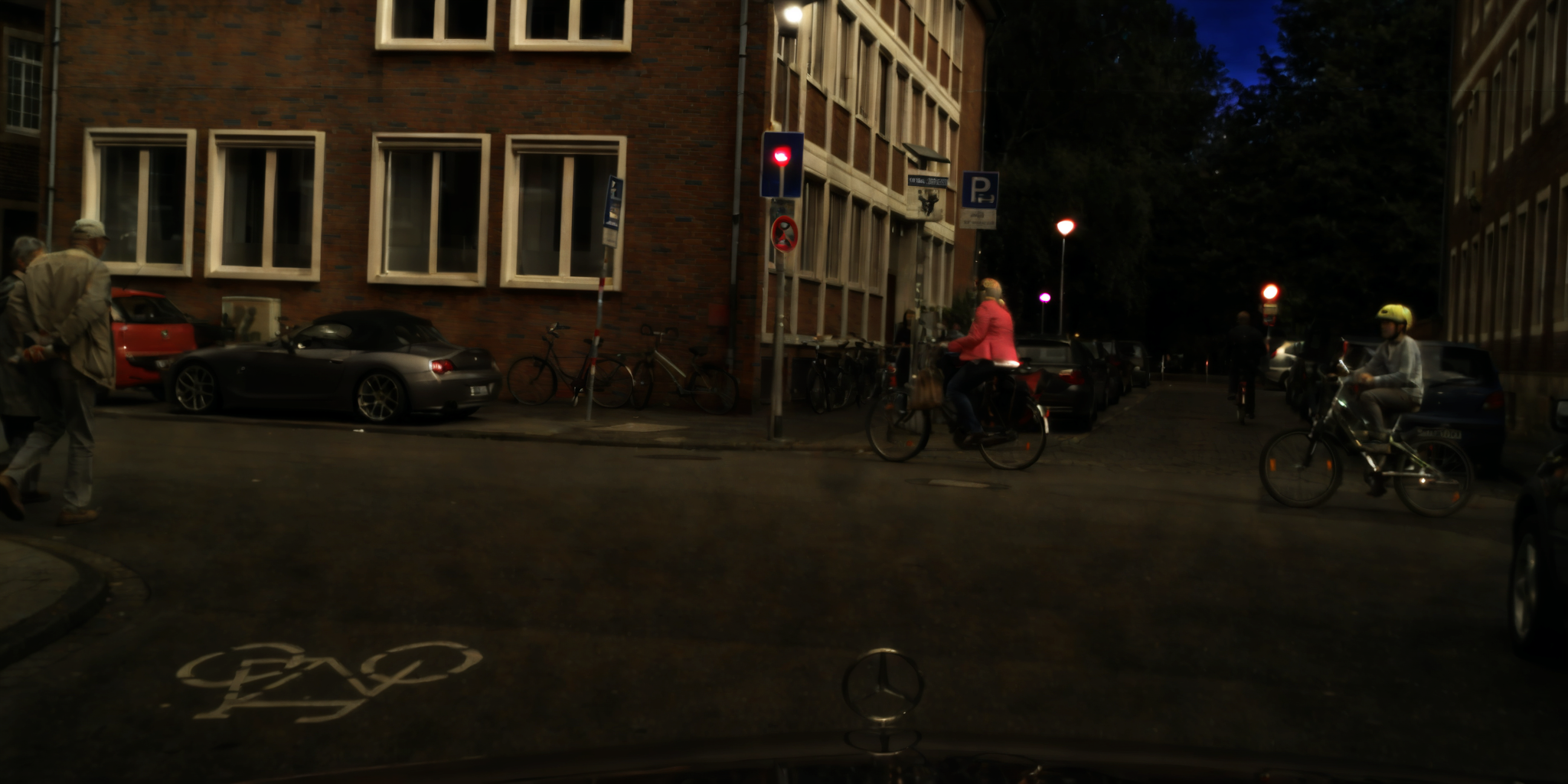} &
\includegraphics[width=\cityfigwidth]{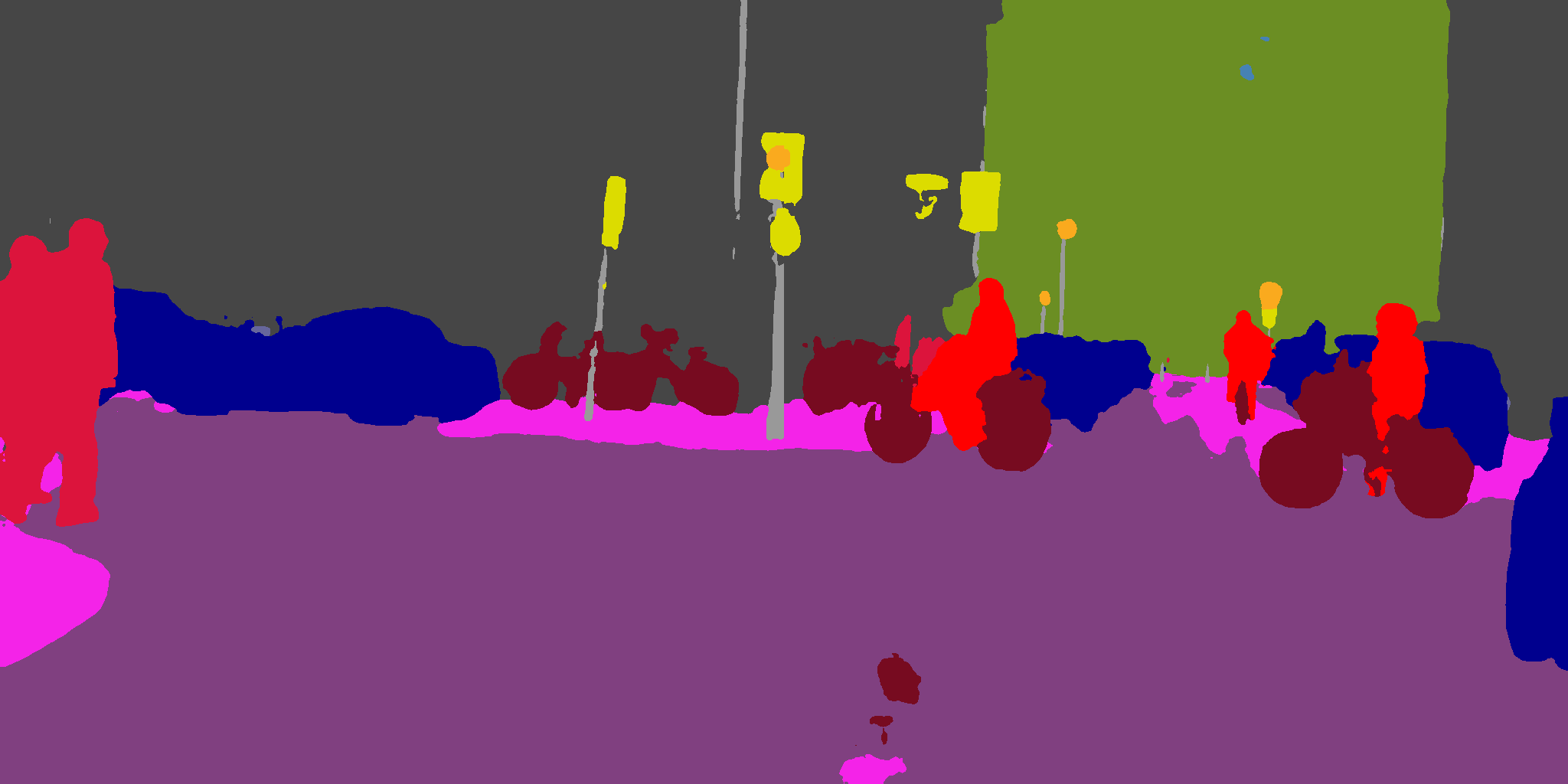} &
\includegraphics[width=\cityfigwidth]{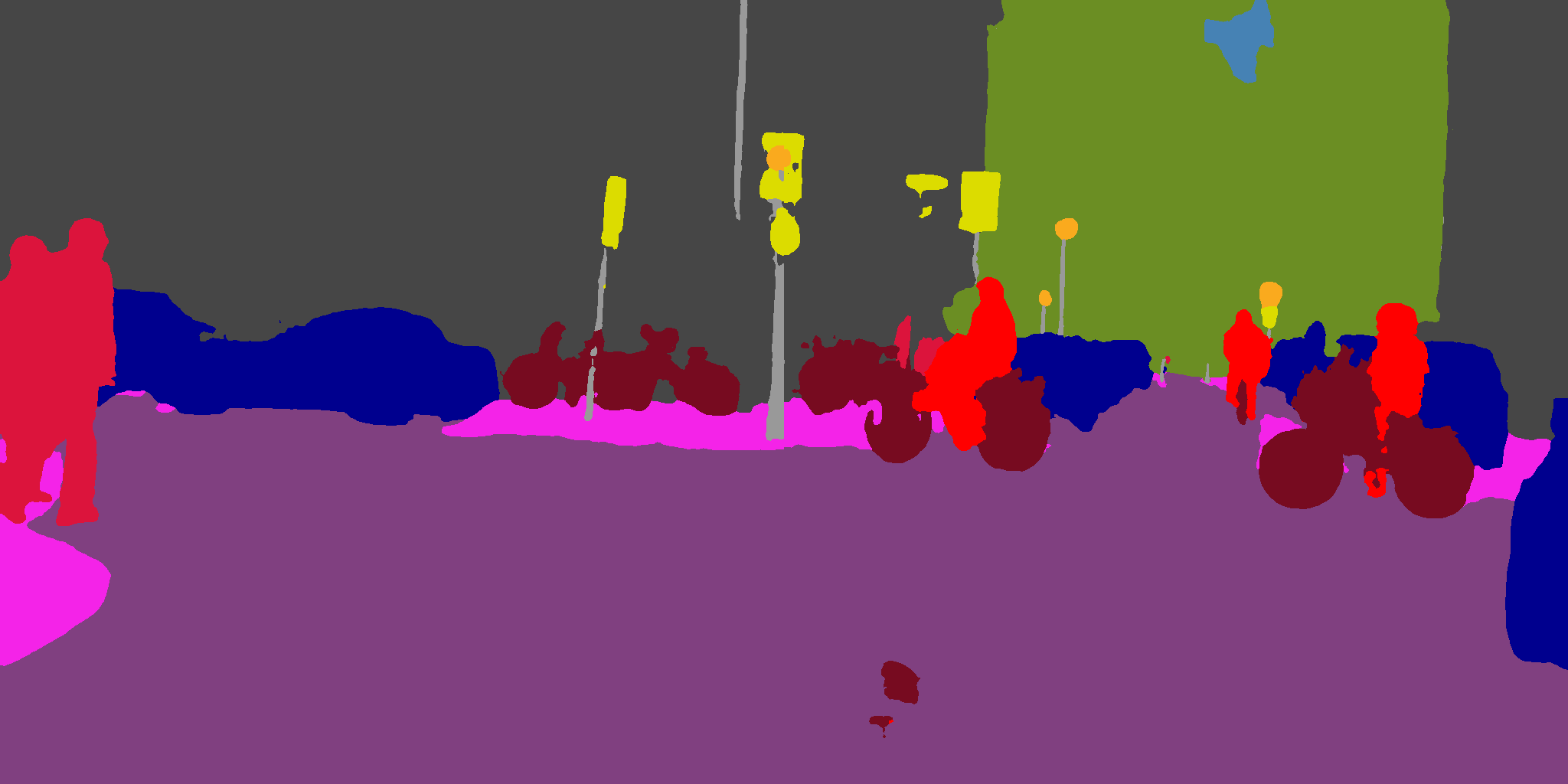} &
\includegraphics[width=\cityfigwidth]{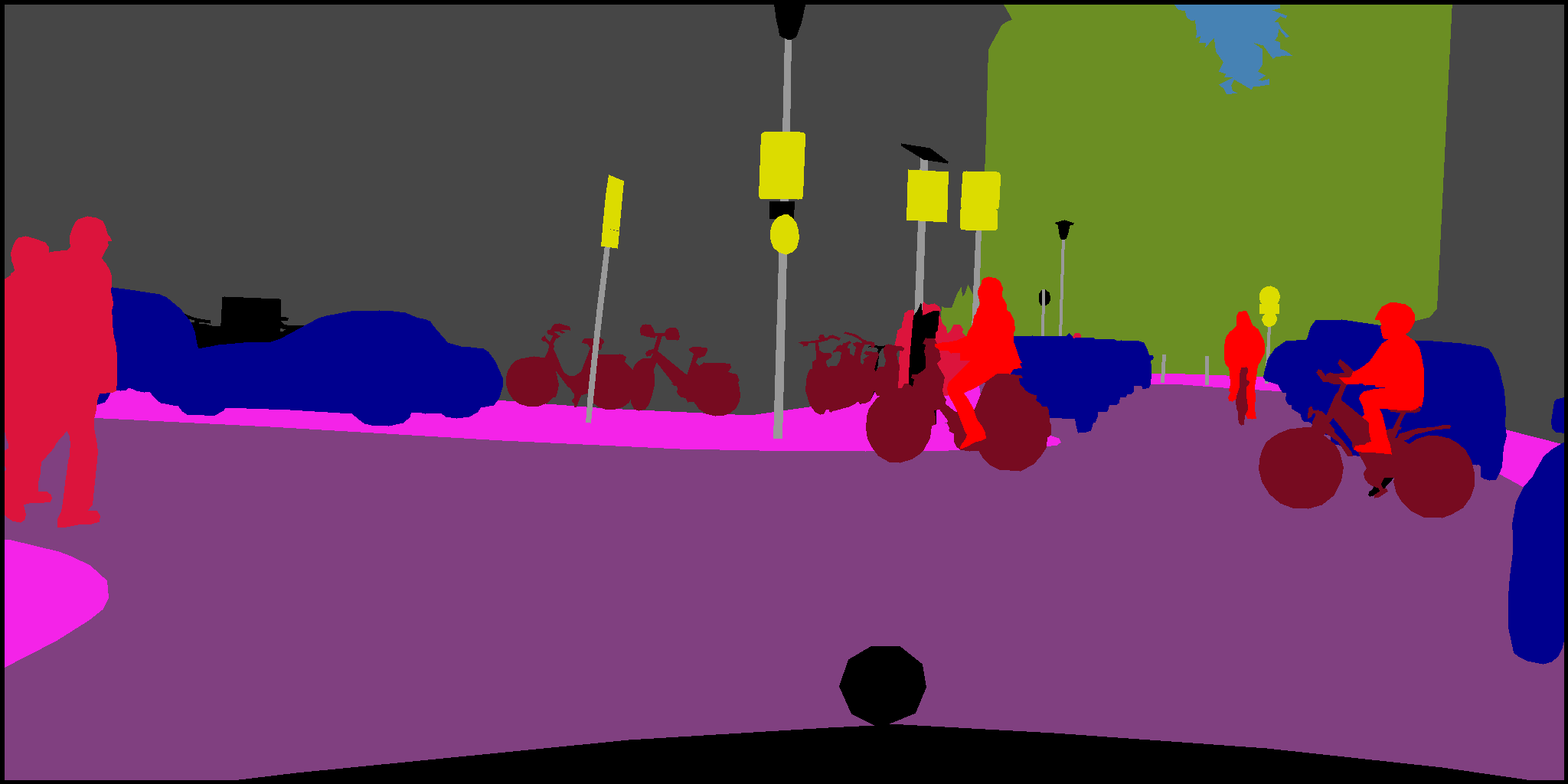} \\

\includegraphics[width=\cityfigwidth]{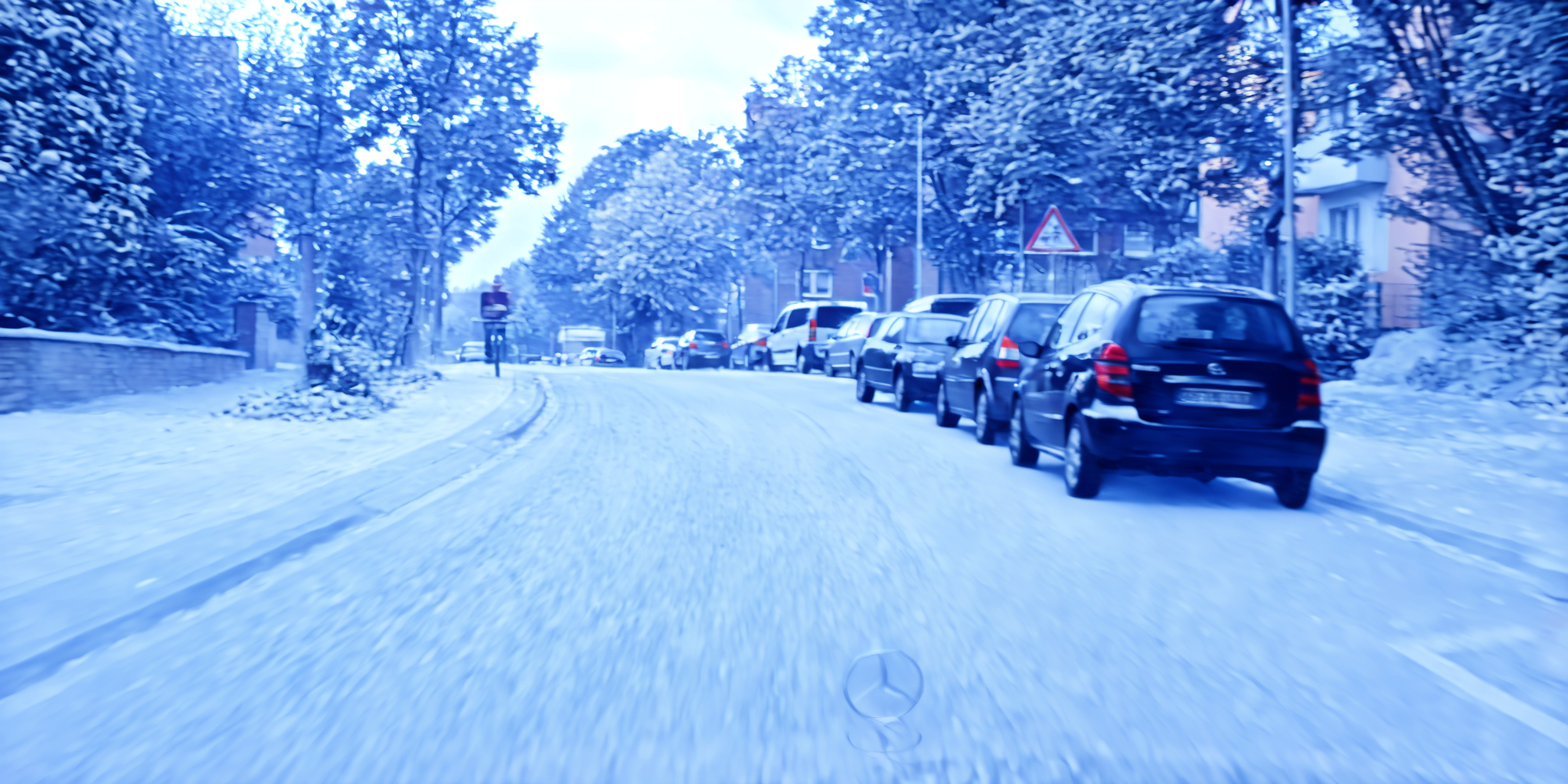} &
\includegraphics[width=\cityfigwidth]{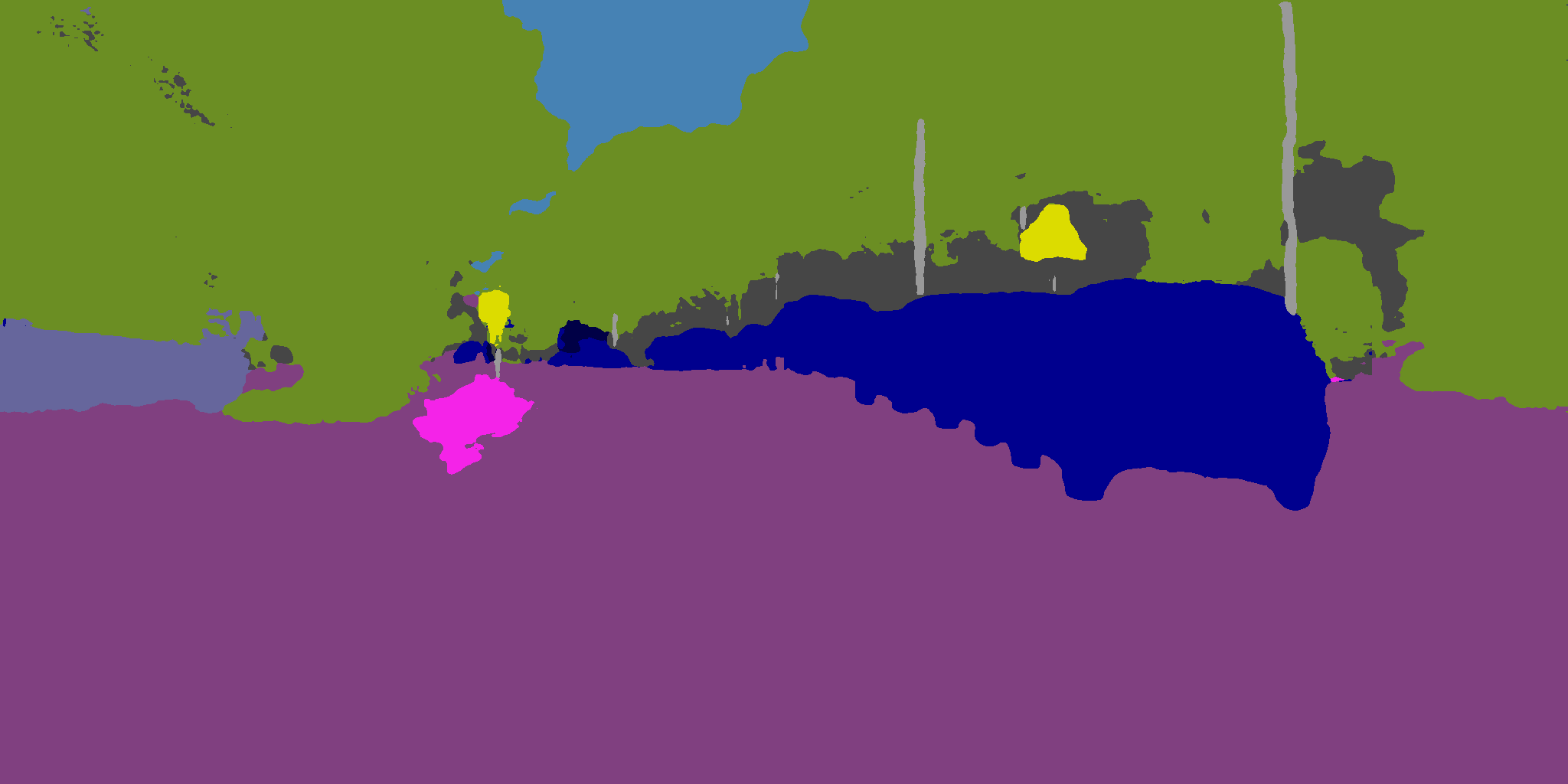} &
\includegraphics[width=\cityfigwidth]{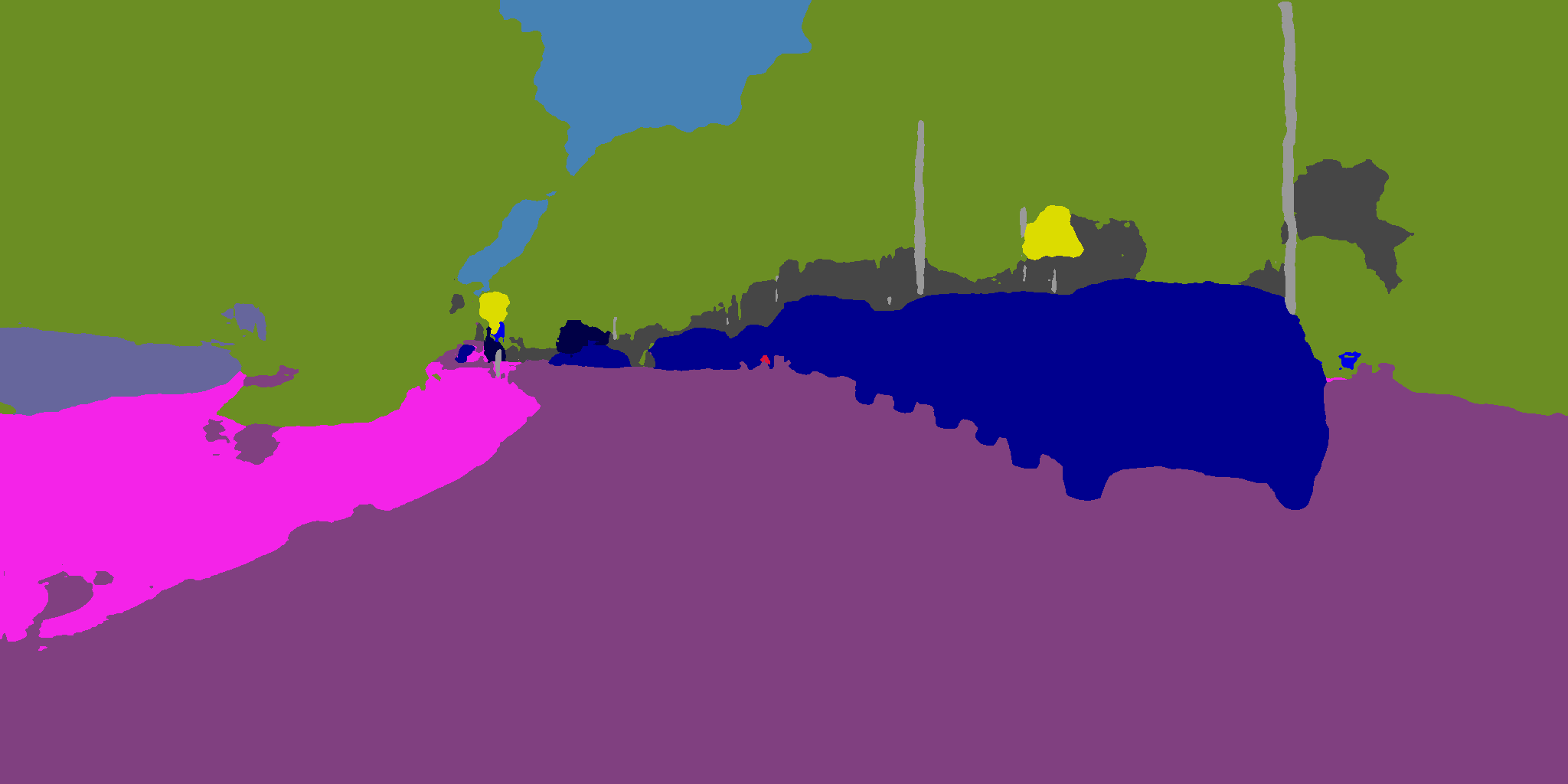} &
\includegraphics[width=\cityfigwidth]{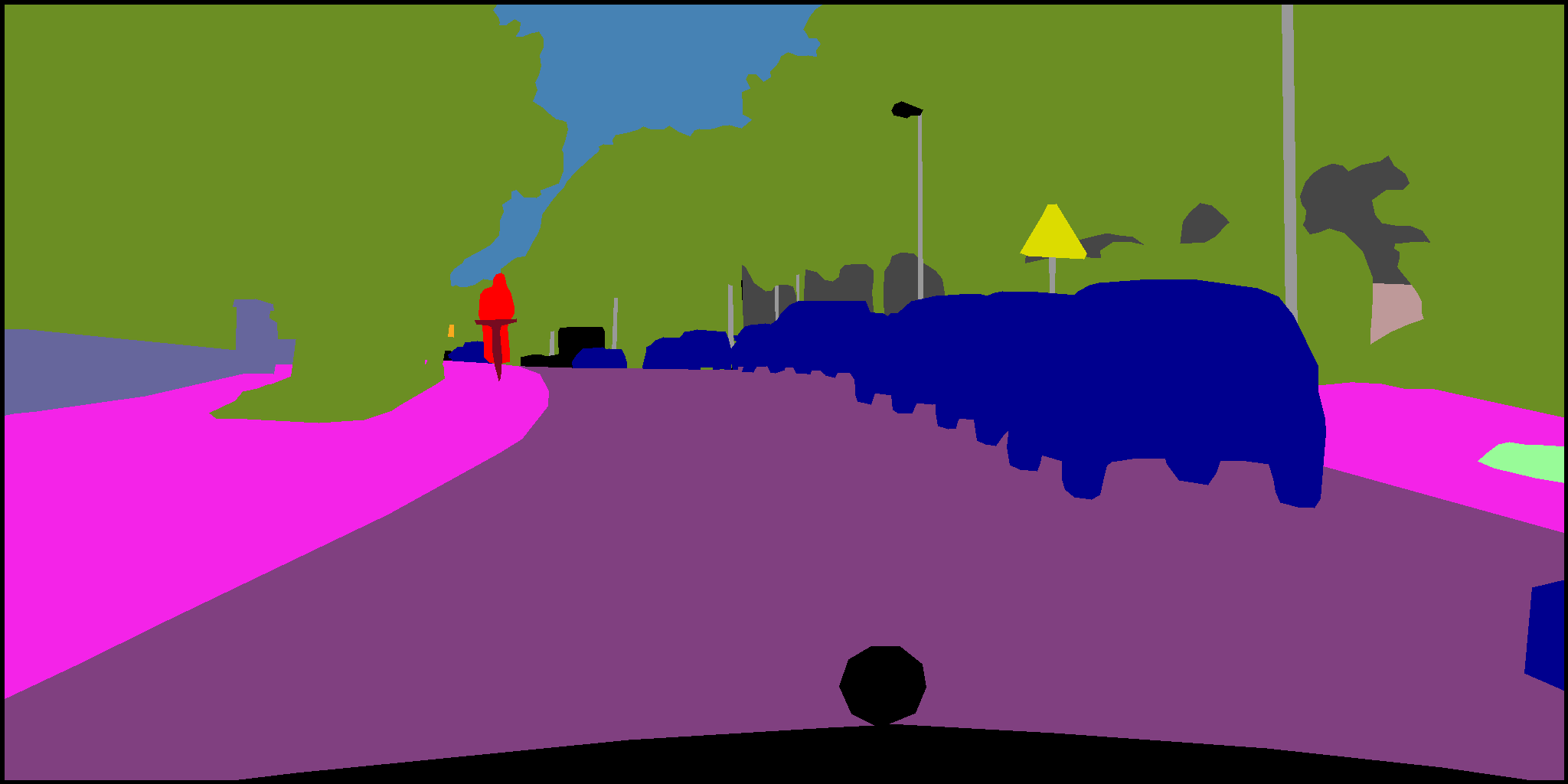} \\

\end{tabular} 

\vspace{-0.2cm}
\caption{
Qualitative results on the Cityscapes Adverse dataset. 
\method{} improves segmentation robustness, helping recover critical regions such as the sky and refining overall precision.
}
\label{fig:qualitative_cityscapes}
\end{figure*}

\begin{table*}[t]
\caption{Ablation study varying patch size and number of patches. Left: Cityscapes Adverse. Right: Hard-DSEC-DET.}
\label{tab:ablationpatch}
\centering
    \begin{subtable}{.65\linewidth}
      \centering
        \resizebox{\linewidth}{!}{
        \begin{tabular}{ccHccccccccc}
        \toprule
\textbf{Patch Size} & \textbf{Patches} & \textbf{Normal Cityscapes} & \textbf{Rainy} & \textbf{Foggy} & \textbf{Spring} & \textbf{Autumn} & \textbf{Snow} & \textbf{Sunny} & \textbf{Night} & \textbf{Dawn} & \textbf{Avg.}\\ 
\midrule
2x2 & 2 & 76.1 & \textbf{63.9} & 67.0 & 69.4 & \textbf{67.0} & 47.7 & 58.4 & 63.2 & 67.5 & 63.0 \\
2x2 & 4 & 76.0 & 62.9 & 67.1 & 69.1 & 66.8 & 46.3 & 58.5 & \textbf{63.7} & \textbf{67.7} & 62.7\\
4x4 & 2 & 75.7 & 62.5 & 67.7 & 69.2 & 66.0 & 45.4 & 57.7 & 61.0 & 66.4 & 62.0 \\
4x4 & 4 & 76.3 & 62.1 & 67.4 & 69.7 & 66.4 & 48.3 & 58.5 & 61.6 & 67.3 & 62.6\\
8x8 & 2 & \textbf{76.6} & 62.8 & \textbf{67.8} & \textbf{70.2} & 66.6 & 47.1 & \textbf{59.0} & 63.9 & 67.7 & \textbf{63.1}\\
8x8 & 4 & 75.8 & 63.2 & 67.4 & 69.4 & 66.1 & \textbf{49.0} & 58.1 & 63.5 & 67.3 & 63.0 \\
\bottomrule
\end{tabular}
}
    \end{subtable}%
    ~~~~~
    \begin{subtable}{.29\linewidth}
      \centering
                \resizebox{\linewidth}{!}{
        \begin{tabular}{cccc}
        \toprule
\textbf{Patch Size} & \textbf{Patches} & \textbf{mAP$_{50:95}$} & \textbf{mAP$_{50}$} \\
\midrule
4x4 & 2 & 20.0 & 41.0 \\
4x4 & 4 & 20.8 & 41.4 \\
8x8 & 2 & 20.2 & 41.5 \\
8x8 & 4 & 21.6 & \textbf{42.4} \\
16x16 & 2 & \textbf{22.3} & 41.4 \\
16x16 & 4 & 21.3 & 41.4 \\
\bottomrule
\end{tabular}
}

    \end{subtable} 
\end{table*}

\textbf{Analysis of Predictive Target Granularity. \,}
We perform an ablation study to investigate the influence of the latent patch configuration, specifically varying the patch size and the number of patches $M$ used for the predictive objective of Eq.~\ref{eq:loss_feat_pred}. As detailed in Tab.~\ref{tab:ablationpatch}, we evaluate configurations on both Cityscapes Adverse for semantic segmentation and Hard-DSEC-DET for object detection. For the segmentation task, which employs a SegFormer model producing $16 \times 16$ feature maps, performance remains robust across different settings. The optimal average mIoU (63.1\%) is achieved using a patch size of $8 \times 8$, with minimal sensitivity to the number of patches, yielding similar results for both $M=2$ and $M=4$. For the detection task on Hard-DSEC-DET, which utilizes a DETR model with $34 \times 25$ feature maps, we observe a different sensitivity. While the peak $mAP_{50:95}$ (22.3) is achieved with two $16 \times 16$ patches, the configuration of four $8 \times 8$ patches provides a strong $mAP_{50:95}$ of 21.6 and the highest $mAP_{50}$ (42.4). This indicates a balanced trade-off, and consequently, we adopted the $8 \times 8$ patch size with $M=2$ for segmentation and $M=4$ for detection as the preferred configurations.

\begin{table}[t]
\caption{Ablation on loss importance.}
\label{tab:ablationloss}
\centering
\resizebox{\columnwidth}{!}{
\begin{tabular}{ccHccccccccc}
\toprule
$\lambda_{\text{task}}$ & $\lambda_{\text{feat}}$  & \textbf{Normal Cityscapes} & \textbf{Rainy} & \textbf{Foggy} & \textbf{Spring} & \textbf{Autumn} & \textbf{Snow} & \textbf{Sunny} & \textbf{Night} & \textbf{Dawn} & \textbf{Avg.}\\ \midrule
1 & 0.5 & 76.4 & 62.5 & 67.6 & 70.0 & 65.6 & 47.5 & 58.6 & 62.2 & 66.2 & 62.5 \\
1 & 0.1 & 75.9 & 61.4 & 66.7 & 69.2 & 64.8 & 42.2 & 56.7 & 61.5 & 65.0 & 60.9\\
0.5 & 1 & 75.4 & \textbf{62.8} & 67.8 & 69.8 & 66.6 & 48.3 & 58.3 & 63.3 & 67.3 & 63.0\\
0.1 & 1 & 74.8 & 60.8 & 66.4 & 69.0 & 66.4 & \textbf{51.8} & 58.4 & 60.0 & 65.7 & 62.3\\
1 & 1 & \textbf{76.6} & \textbf{62.8} & \textbf{67.8} & \textbf{70.2} & \textbf{66.6} & 47.1 & \textbf{59.0} & \textbf{63.9} & \textbf{67.7} & \textbf{63.1} \\
\bottomrule
\end{tabular}
}

\end{table}

\textbf{Sensitivity to Loss Weighting. \,} 
We investigate the sensitivity of our framework to the relative weighting of the two primary objectives defined in Eq. \ref{eq:total_loss_pepr}. The supervised task loss $\mathcal{L}_{task}$ and our predictive regularization loss $\mathcal{L}_{feat}$, governed by their respective hyperparameters $\lambda_{task}$ and $\lambda_{feat}$. We conducted a parameter sweep on the Cityscapes Adverse benchmark, with the empirical results presented in Tab. \ref{tab:ablationloss}. The analysis reveals that the model's performance is relatively robust to moderate variations in these weights. However, an imbalance where the task-specific objective is excessively down-weighted results in a performance drop to 62.3\% mIoU. A more significant degradation to 60.9\% mIoU is observed when the predictive regularization is minimized ($\lambda_{feat}=0.1$, $\lambda_{task}=1$), pointing out the critical contribution of the privileged information. Our evaluation identifies an optimal balance at $\lambda_{task}=1$ and $\lambda_{feat}=1$. This configuration was adopted throughout the paper.


\section{Conclusions}
\label{sec:conclusions}

We addressed domain generalization by leveraging event cameras as privileged information at training while keeping the final model purely RGB at test time. We introduced \methodname{} (\method{}), which reframes LUPI as a predictive problem in a shared latent space: instead of directly aligning RGB and event features, the RGB encoder predicts event-based latent representations, distilling domain-invariant cues without sacrificing semantic richness.
Experiments on object detection (Hard-DSEC-DET, FRED) and semantic segmentation (Cityscapes Adverse, Dark Zurich) show that \method{} consistently improves under severe domain shifts, particularly day-to-night transitions, and outperforms a direct alignment baselines, being also a beneficial regularizer on the source domain.
Results demonstrate that predictive cross-modal regularization is a more effective way to exploit privileged information than direct feature matching.
A discussion of limitations and future work is provided in Appendix~\ref{sec:limitations}.

\section*{Acknowledgment}
This work was partially supported by: EU Horizon project ELLIOT (No. 101214398); FIS project GUIDANCE (No. FIS2023-03251); RAINFALL: Recognition Algorithms for Interpreting Neuromorphic-based Facial Actions with Low Latency (piano per lo sviluppo della ricerca PSR 2025, University of Siena).


{
    \small
    \bibliographystyle{ieeenat_fullname}
    \bibliography{main}
}

\clearpage
\appendix
\maketitlesupplementary

\section*{Overview}
In the supplementary material, we provide additional technical details, extended experiments, and qualitative results to further support the findings presented in the main paper. Specifically, we include:
\begin{itemize}
    \item \textbf{Sec. \ref{sec:implementation_details}}: a detailed description of the training setup, architectural components, and implementation choices.
    \item \textbf{Sec. \ref{sec:qualitative_obj_detection}}: additional qualitative results for detection are discussed here, highlighting the proposed method qualitites.
    \item \textbf{Sec. \ref{sec:dsec_det}}: in-domain evaluation on DSEC-DET, to demonstrate the effect of \method{} on data drawn from the same distribution as the training data.
    \item \textbf{Sec. \ref{sec:qualitative_segmentation}}: discussion regarding the qualitative results obtained under diverse adverse domains for the Cityscapes Adverse dataset. 
    \item \textbf{Sec. \ref{sec:limitations}}: discussion of limitations and future works.
\end{itemize}

\section{Implementation Details}
\label{sec:implementation_details}

In this section, we provide a detailed description of the experimental setup, architectural choices, and hyperparameter configurations adopted to evaluate \method{} on the tasks of Object Detection and Semantic Segmentation. 
Both for the detection and the segmentation task, the predictor $g_\phi$ is a transformer decoder with a depth of 4 and 8 attention heads. We attach the source code as supplementary material.

\paragraph{Object Detection.} For the detection task, we employ the DETR architecture \cite{carion2020end}. We utilize the implementation provided by the \texttt{transformers} Python library, initializing both the RGB and Event encoders with weights pretrained on the COCO dataset \cite{lin2014microsoft}. 
Event data are converted into Time Surface representations~\cite{zhou2021event} and then processed by the model. 
The two encoders and the RGB decoder are then fine-tuned end-to-end during training. The loss of Eq. \ref{eq:loss_feat_pred} is computed with the outputs of the encoders for the two modalities. At test time, only the RGB branch is used, discarding the event part.
The model is optimized using AdamW \cite{loshchilov2017decoupled} with a learning rate of $1 \times 10^{-5}$ and a weight decay of $1 \times 10^{-4}$. Training is conducted for 20 epochs.

\paragraph{Semantic Segmentation.} For the segmentation task, we adopt SegFormer \cite{xie2021segformer} as our backbone, utilizing the implementation provided by MMSegmentation \cite{mmseg2020}. 
In this case, the predictive loss of Eq. \ref{eq:loss_feat_pred} is computed using features from the last block of the encoder of dimension 64.
To generate the privileged information for the Cityscapes dataset, we simulate event streams using the ESIM simulator~\cite{Rebecq18corl} in conjunction with FILM frame interpolation~\cite{reda2022film}, following the video-to-event pipeline established in \cite{hu2021v2e}.
These simulated events, as well as the real events for DSEC, are converted into Time Surface representations~\cite{zhou2021event}. 
The model is trained using the AdamW optimizer with a learning rate of $6 \times 10^{-5}$, 
a weight decay of $0.01$, and betas set to~$(0.9, 0.999)$. 
We employ a \texttt{poly} learning rate scheduler with a power of $1.0$, a warmup ratio of $1 \times 10^{-6}$, and $1500$ warmup iterations. Following the original SegFormer protocol, we use separate learning rate multipliers for the decoder head. The model is trained for a maximum of 40 epochs with an early stopping patience of 7 epochs.


\section{Qualitative Results Object Detection}
\label{sec:qualitative_obj_detection}

\begin{figure*}[t] 
\centering
\begin{tabular}{ccc}
\textbf{RGB} & \textbf{L2} & \textbf{\method{}} \\

\includegraphics[width=\fredfigwidth, trim={1.5cm 1.5cm 1.5cm 1.5cm},clip]{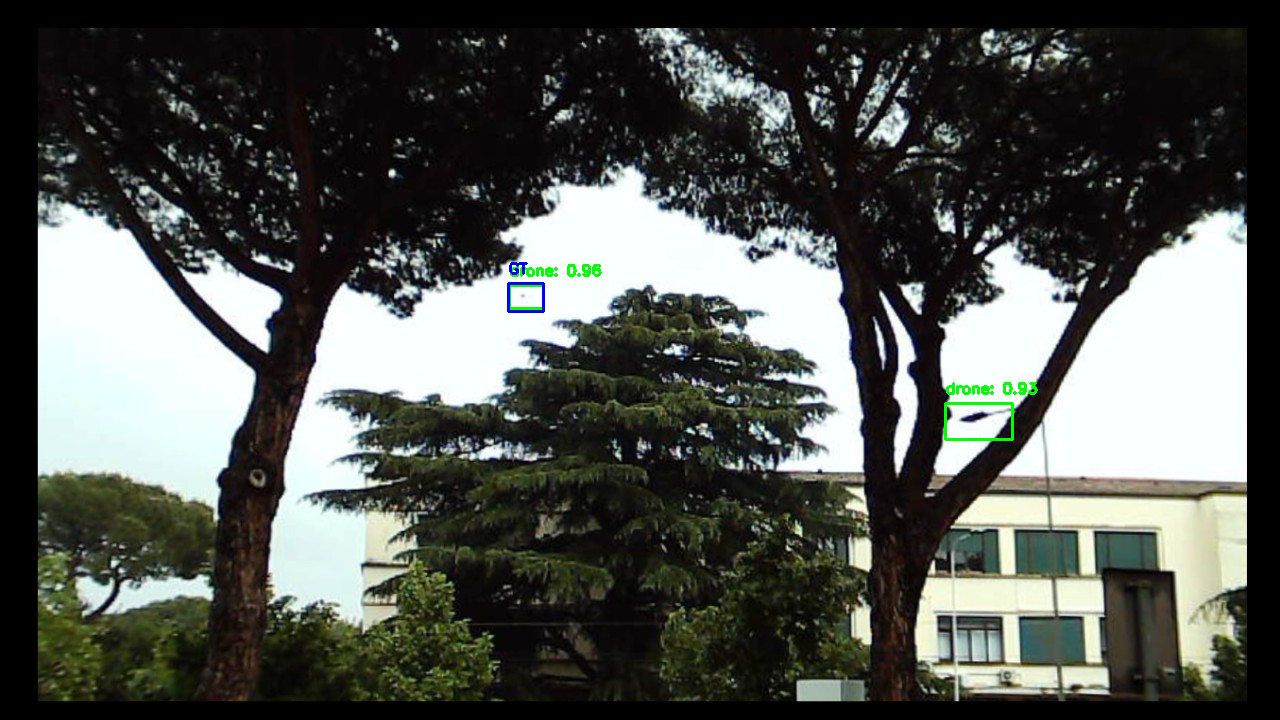} &
\includegraphics[width=\fredfigwidth, trim={1.5cm 1.5cm 1.5cm 1.5cm},clip]{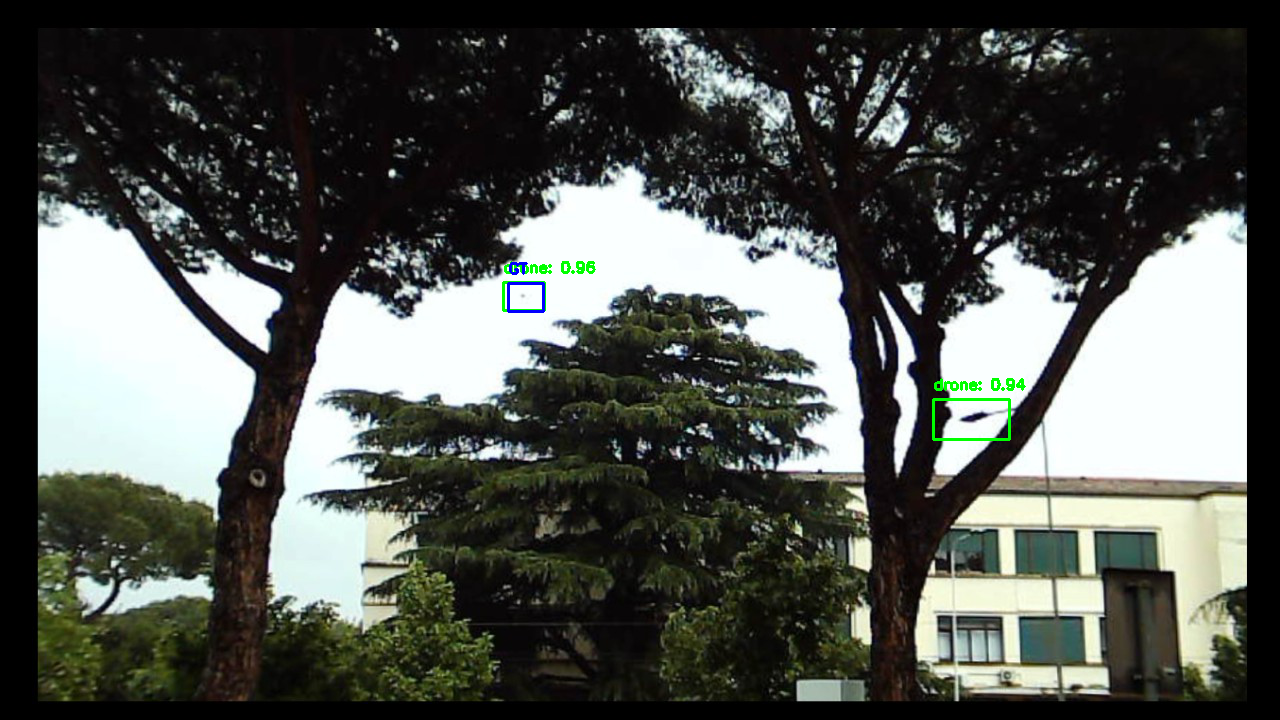} &
\includegraphics[width=\fredfigwidth, trim={1.5cm 1.5cm 1.5cm 1.5cm},clip]{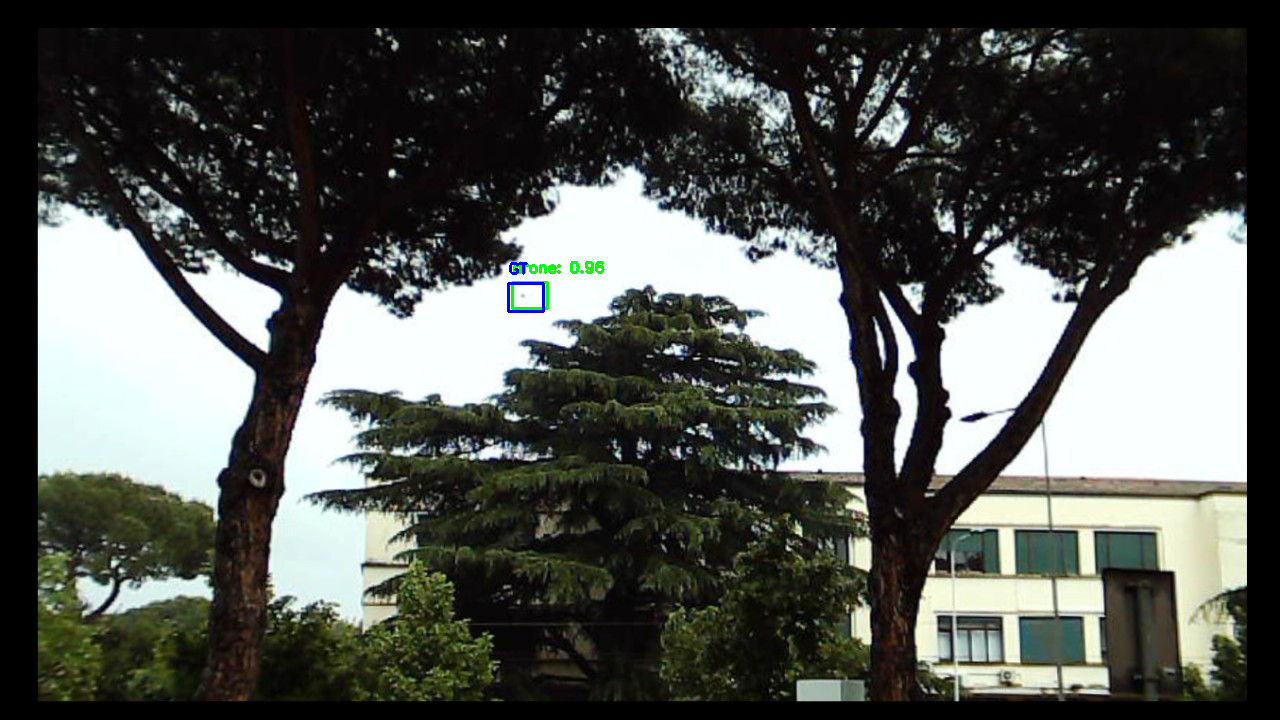} \\
        
\includegraphics[width=\fredfigwidth, trim={1.5cm 1.5cm 1.5cm 1.5cm},clip]{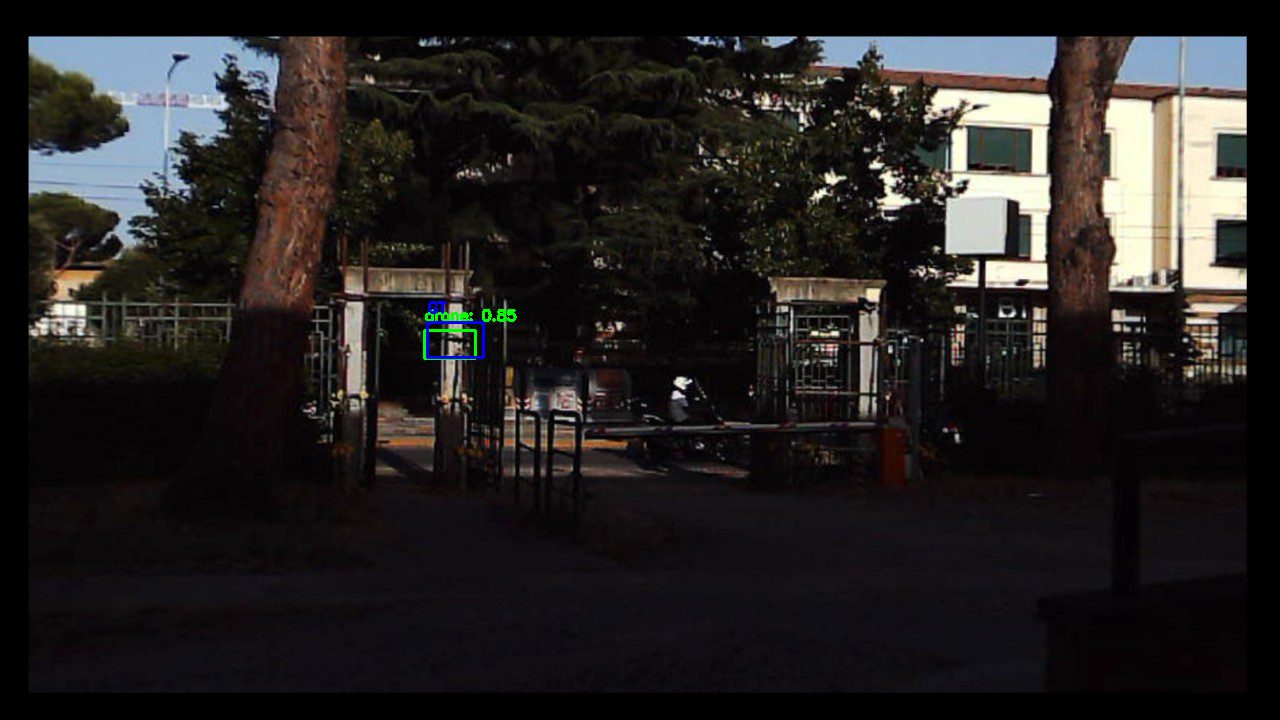} &
\includegraphics[width=\fredfigwidth, trim={1.5cm 1.5cm 1.5cm 1.5cm},clip]{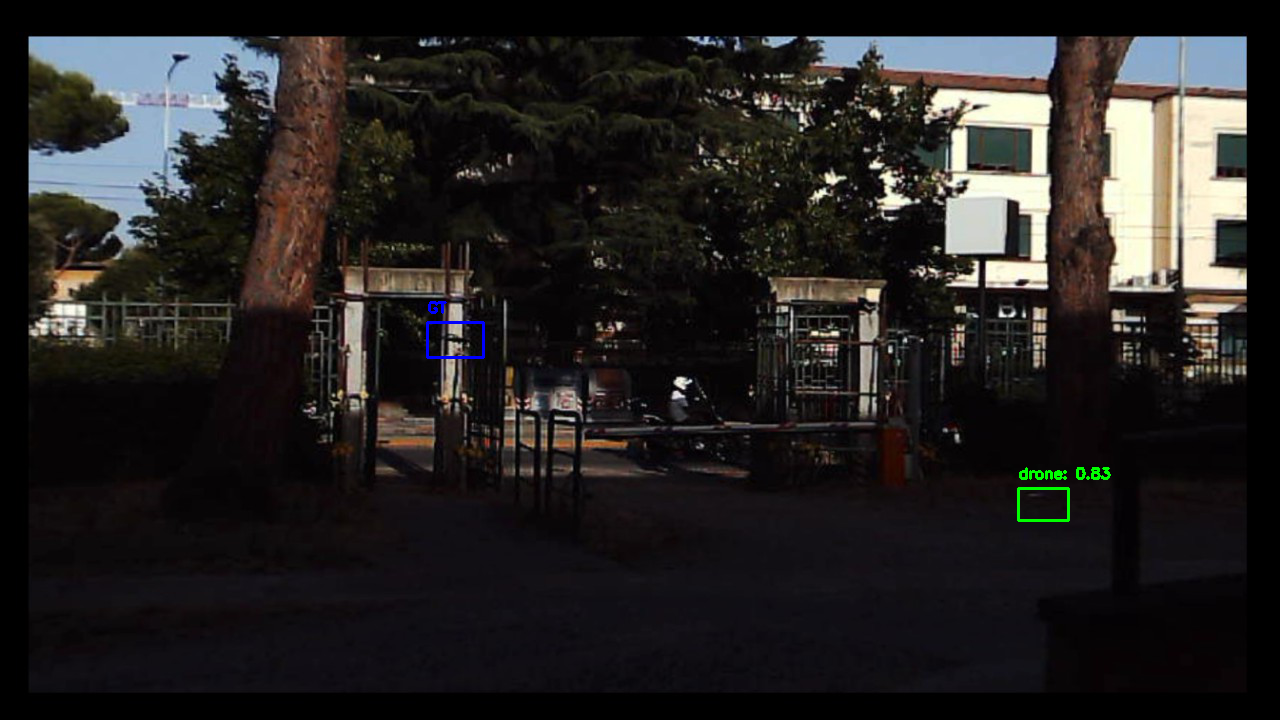} &
\includegraphics[width=\fredfigwidth, trim={1.5cm 1.5cm 1.5cm 1.5cm},clip]{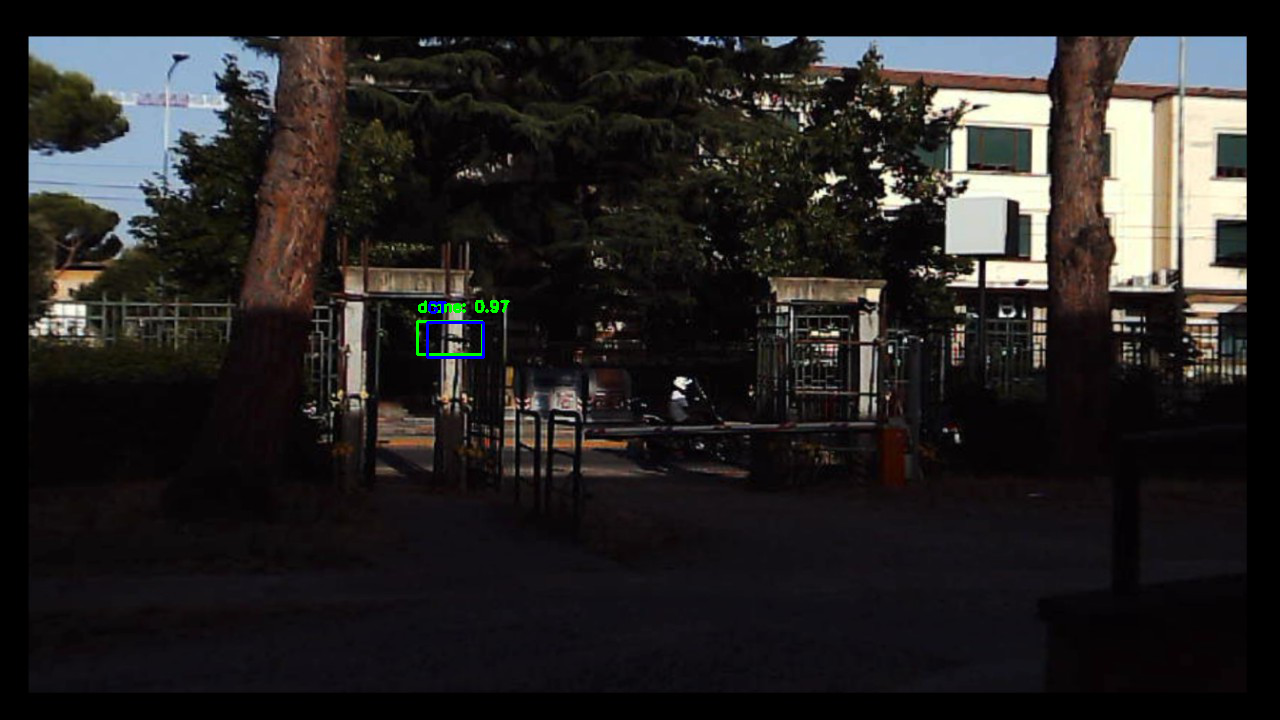} \\

\includegraphics[width=\fredfigwidth, trim={1.5cm 1.5cm 1.5cm 1.5cm},clip]{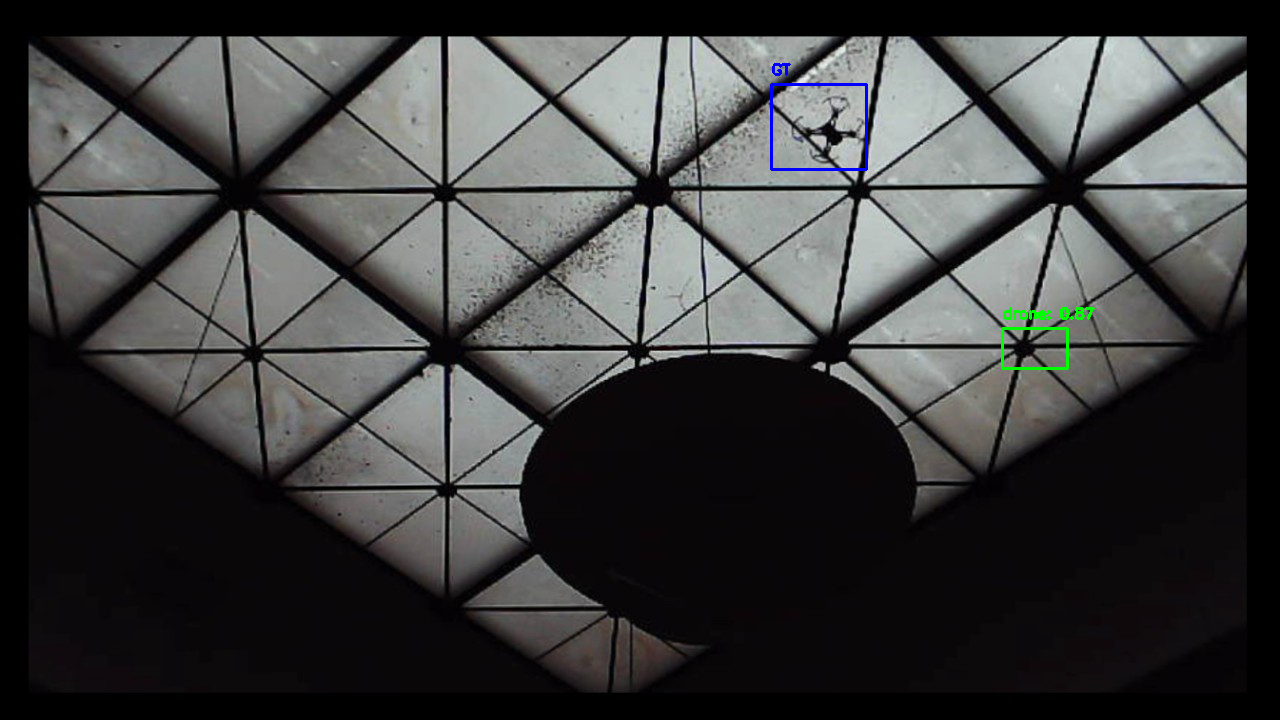} &
\includegraphics[width=\fredfigwidth, trim={1.5cm 1.5cm 1.5cm 1.5cm},clip]{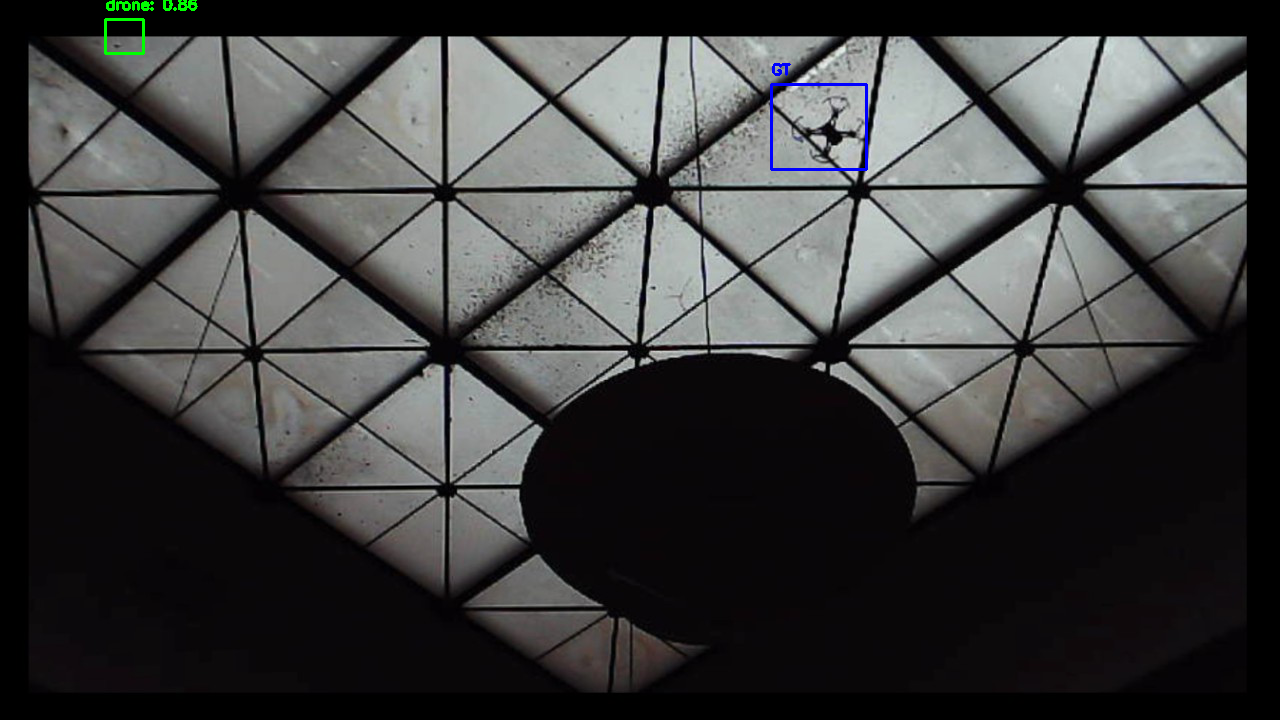} &
\includegraphics[width=\fredfigwidth, trim={1.5cm 1.5cm 1.5cm 1.5cm},clip]{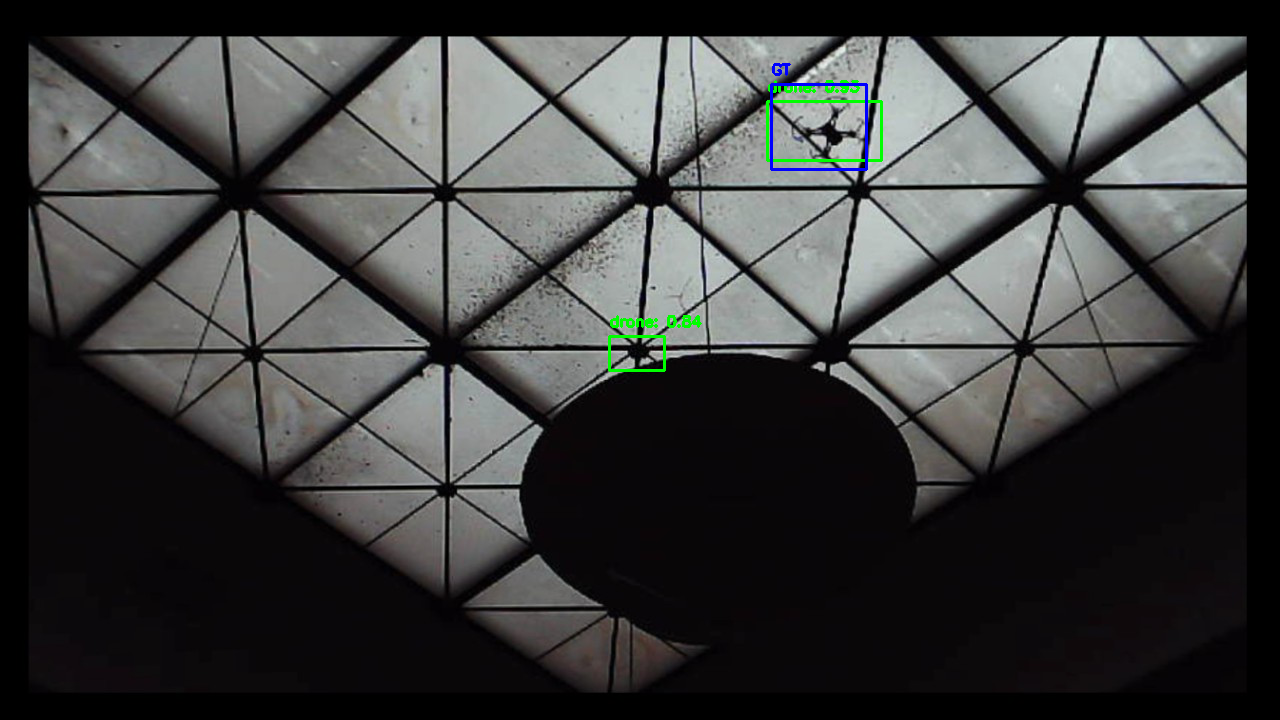} \\

\includegraphics[width=\fredfigwidth, trim={1.5cm 1.5cm 1.5cm 1.5cm},clip]{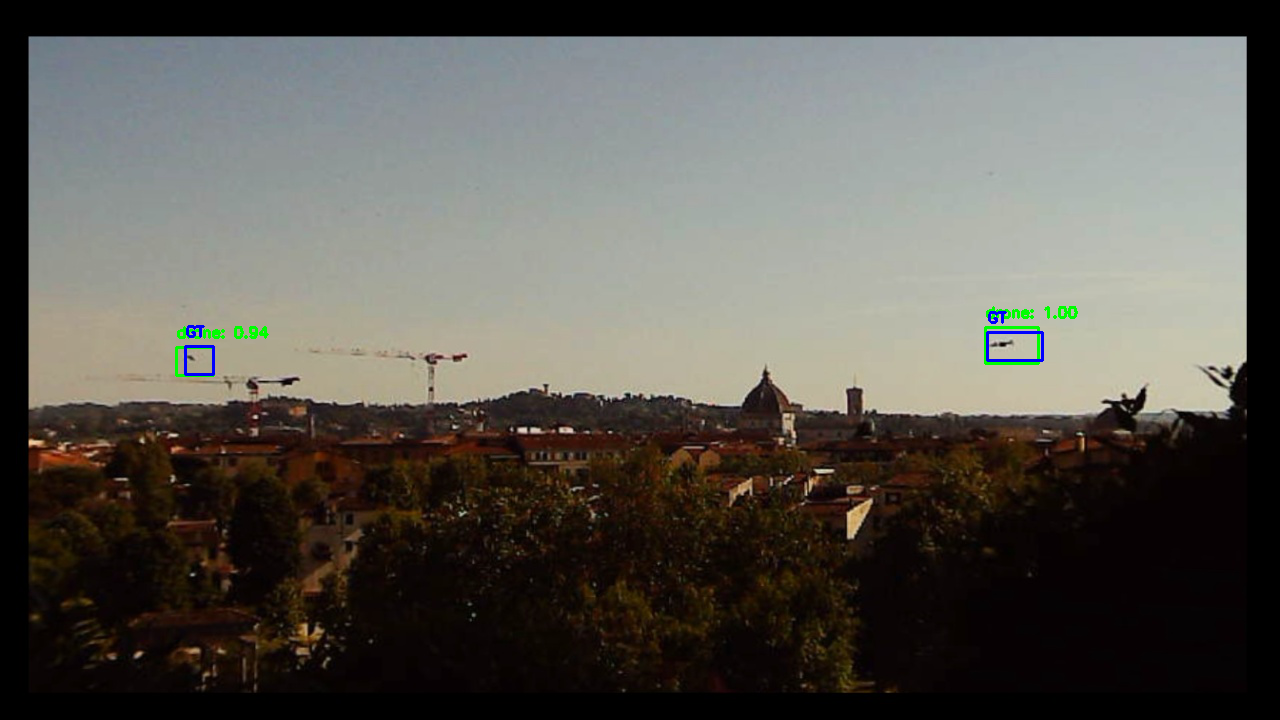} &
\includegraphics[width=\fredfigwidth, trim={1.5cm 1.5cm 1.5cm 1.5cm},clip]{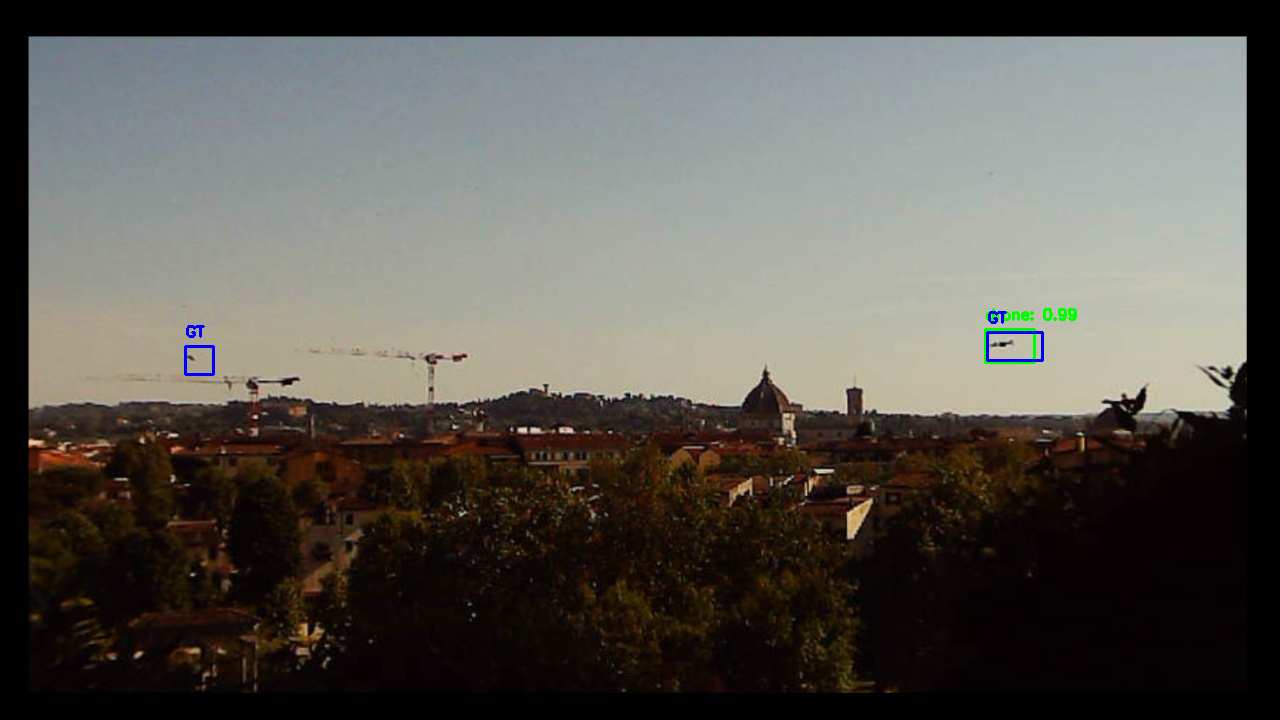} &
\includegraphics[width=\fredfigwidth, trim={1.5cm 1.5cm 1.5cm 1.5cm},clip]{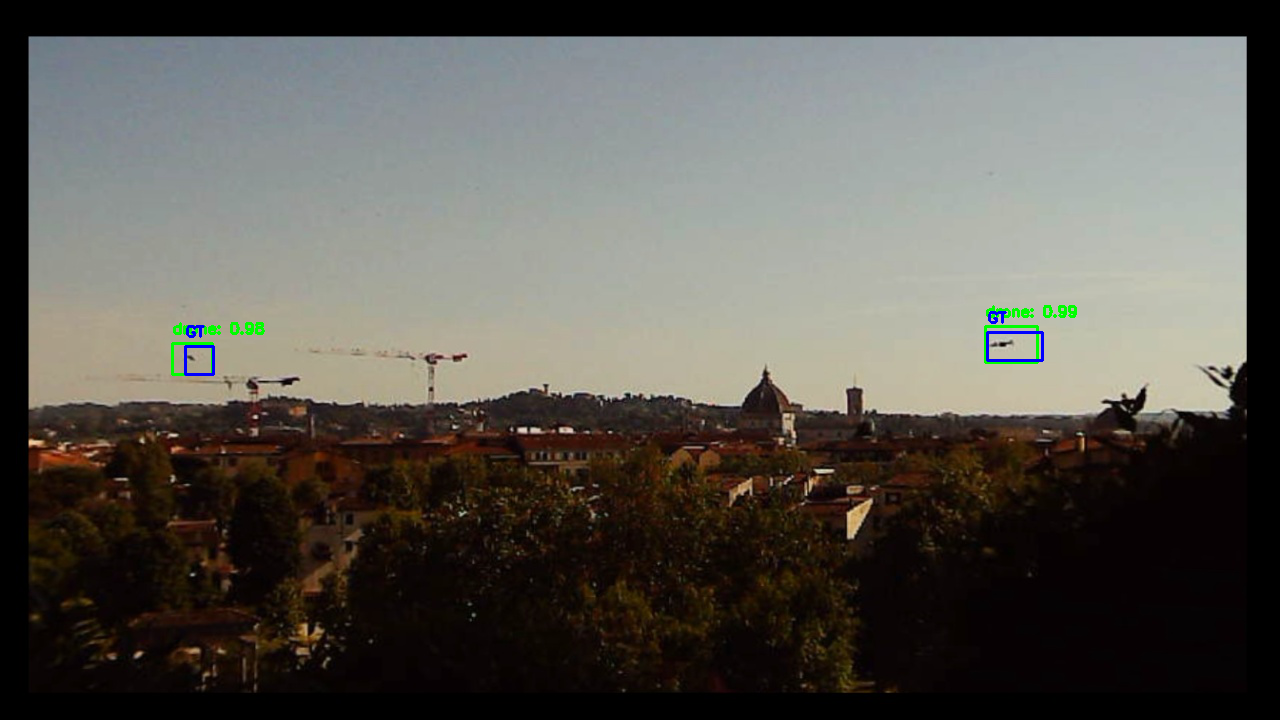} \\

\includegraphics[width=\fredfigwidth, trim={1.5cm 1.5cm 1.5cm 1.5cm},clip]{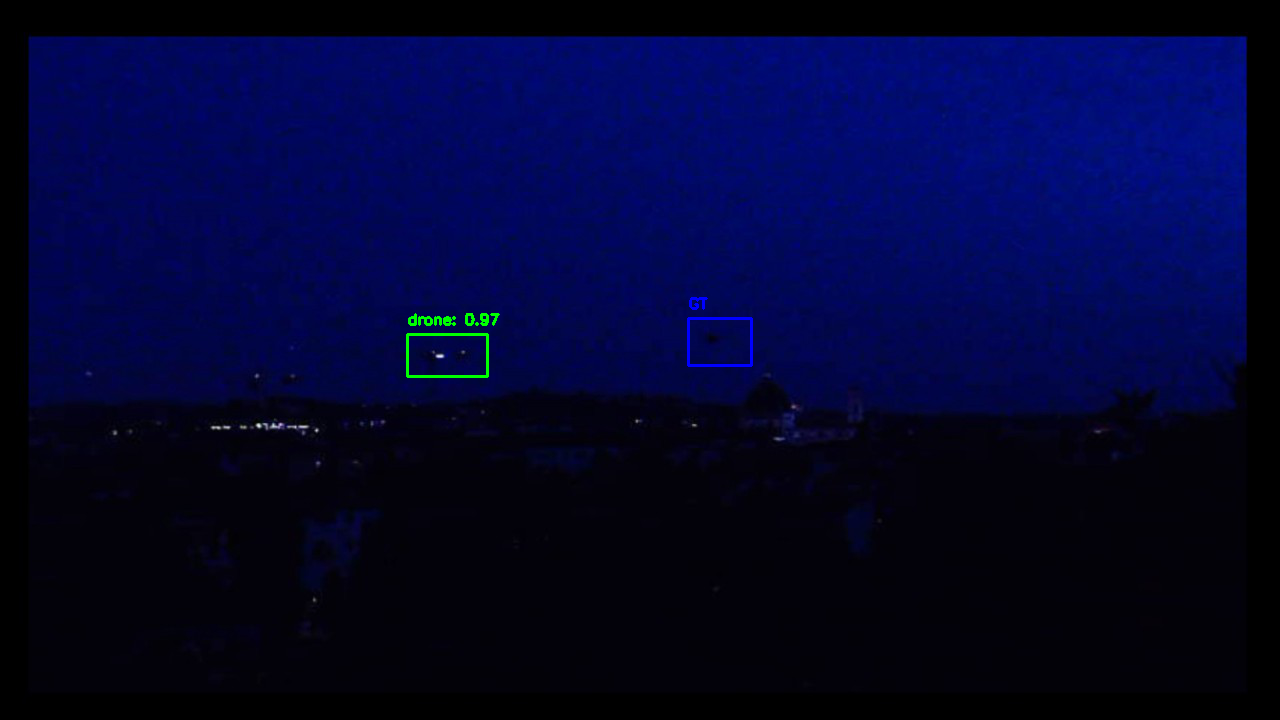} &
\includegraphics[width=\fredfigwidth, trim={1.5cm 1.5cm 1.5cm 1.5cm},clip]{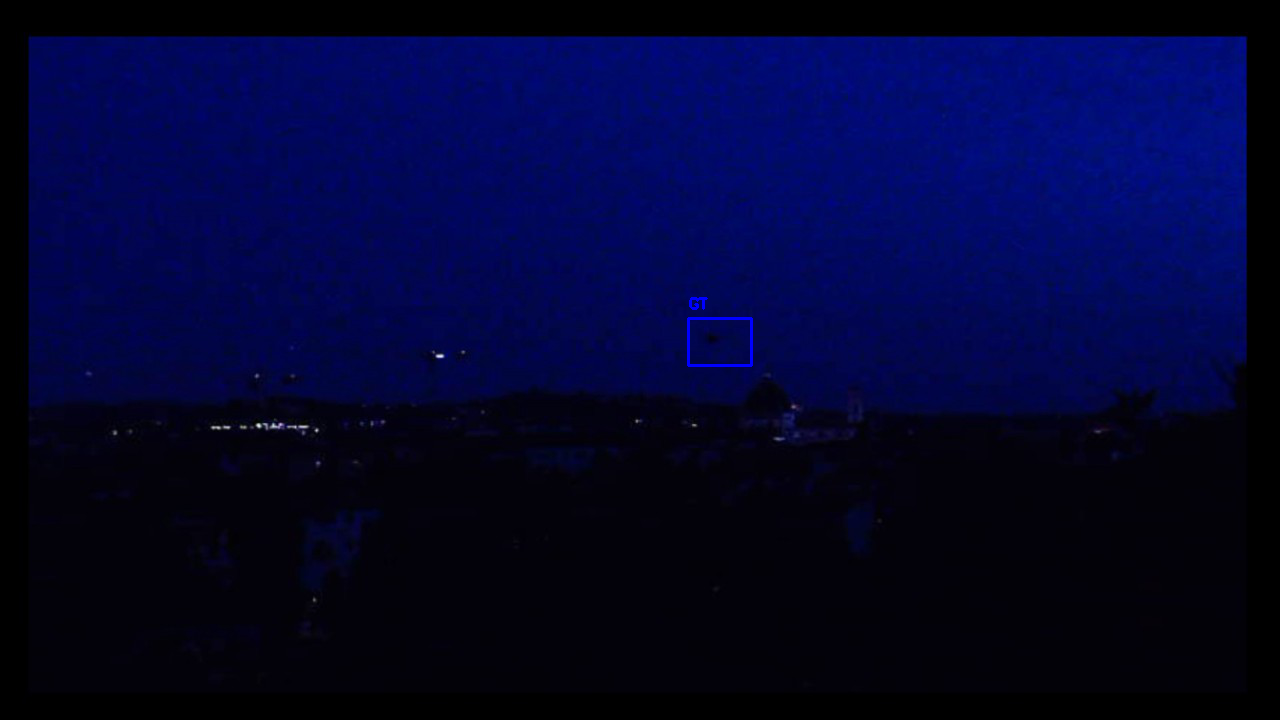} &
\includegraphics[width=\fredfigwidth, trim={1.5cm 1.5cm 1.5cm 1.5cm},clip]{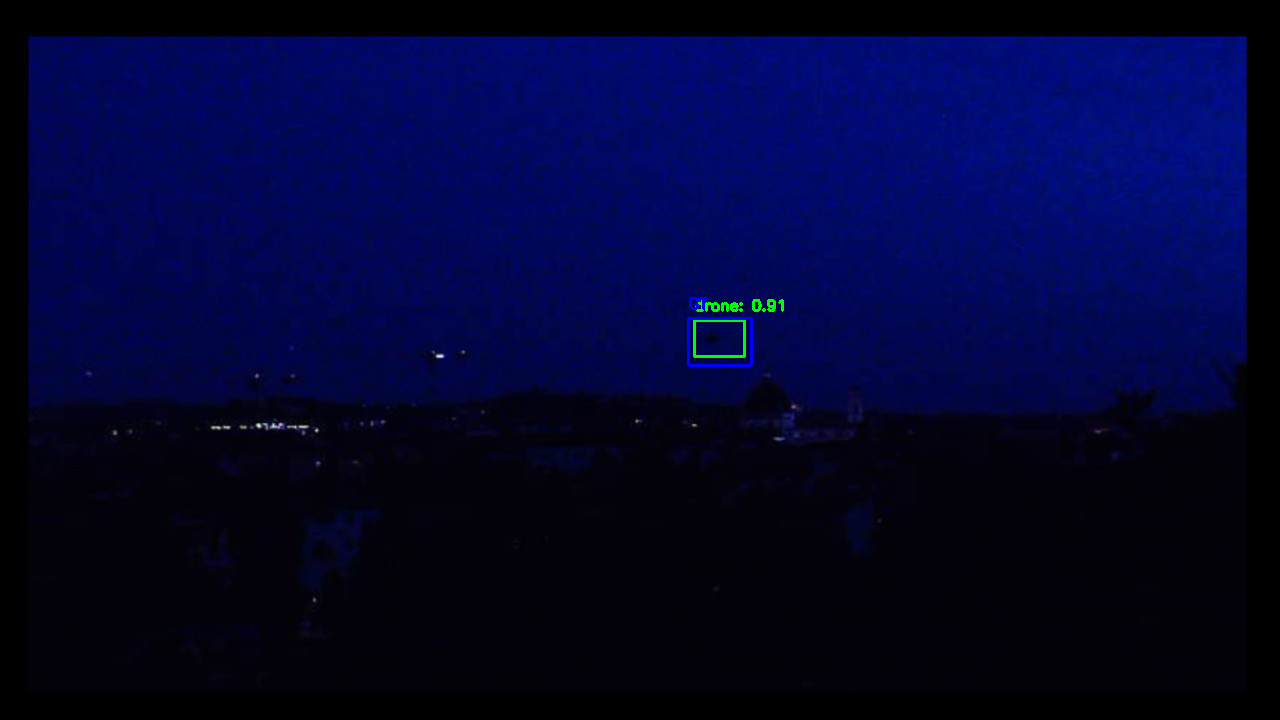} \\

\end{tabular}

\caption{Qualitative results on FRED Challenging. Detections are shown in green, ground truth in blue. }
\label{fig:supplementary_fred}
\end{figure*}

\begin{figure*}[t] 
\centering
\begin{tabular}{ccc}
\textbf{RGB} & \textbf{L2} & \textbf{\method{}} \\

\includegraphics[width=\fredfigwidth, trim={1.5cm 1.5cm 1.5cm 1.5cm},clip]{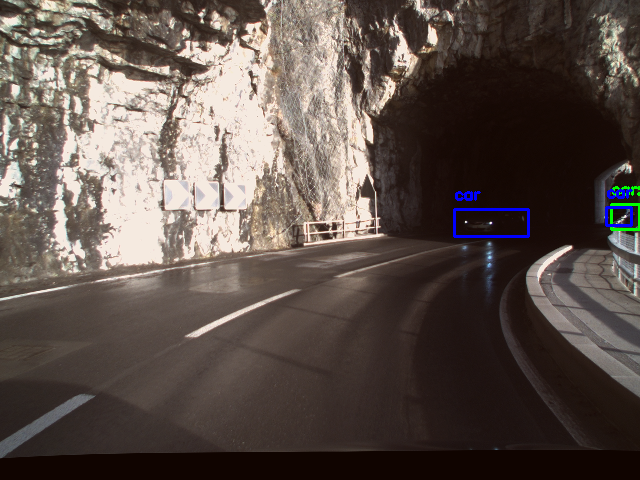} &
\includegraphics[width=\fredfigwidth, trim={1.5cm 1.5cm 1.5cm 1.5cm},clip]{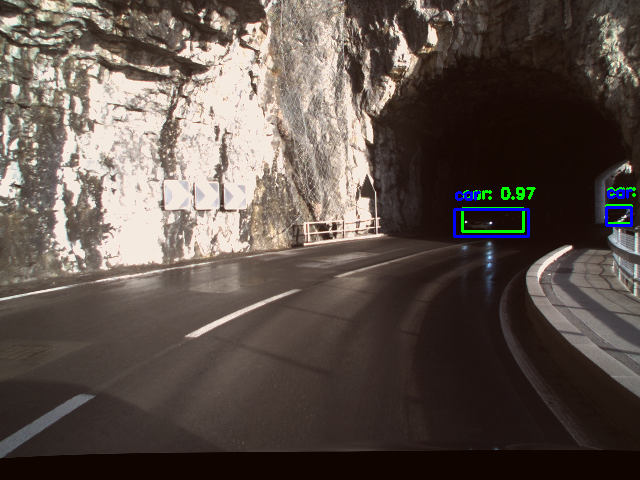} &
\includegraphics[width=\fredfigwidth, trim={1.5cm 1.5cm 1.5cm 1.5cm},clip]{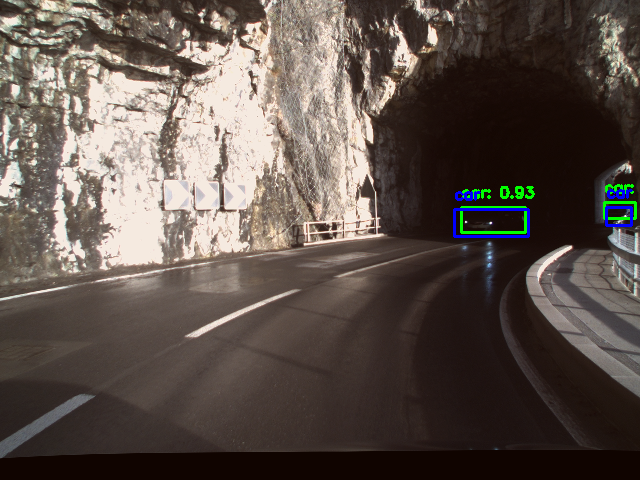} \\
        
\includegraphics[width=\fredfigwidth, trim={1.5cm 1.5cm 1.5cm 1.5cm},clip]{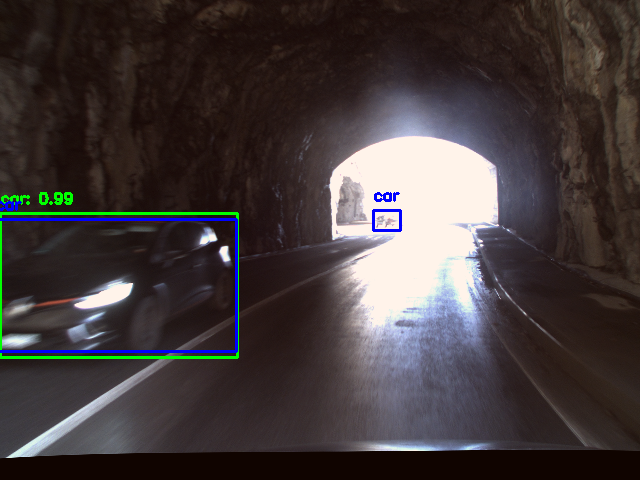} &
\includegraphics[width=\fredfigwidth, trim={1.5cm 1.5cm 1.5cm 1.5cm},clip]{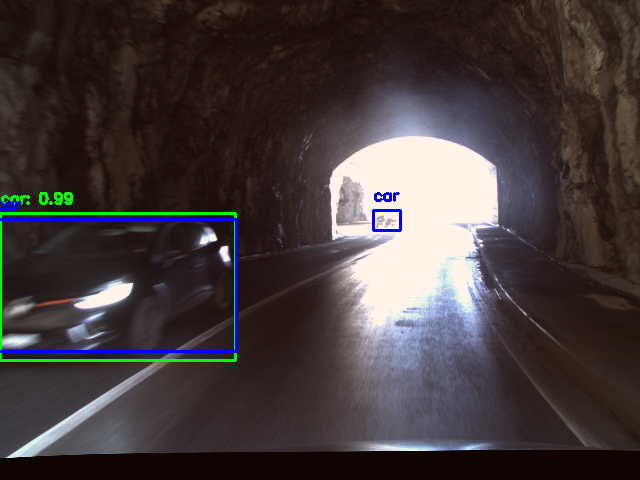} &
\includegraphics[width=\fredfigwidth, trim={1.5cm 1.5cm 1.5cm 1.5cm},clip]{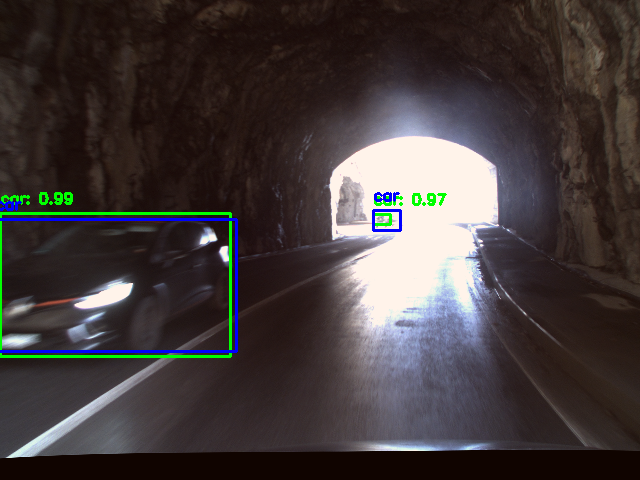} \\

\includegraphics[width=\fredfigwidth, trim={1.5cm 1.5cm 1.5cm 1.5cm},clip]{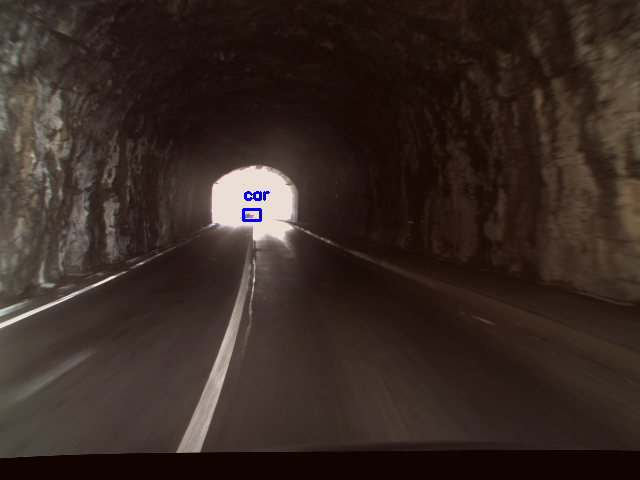} &
\includegraphics[width=\fredfigwidth, trim={1.5cm 1.5cm 1.5cm 1.5cm},clip]{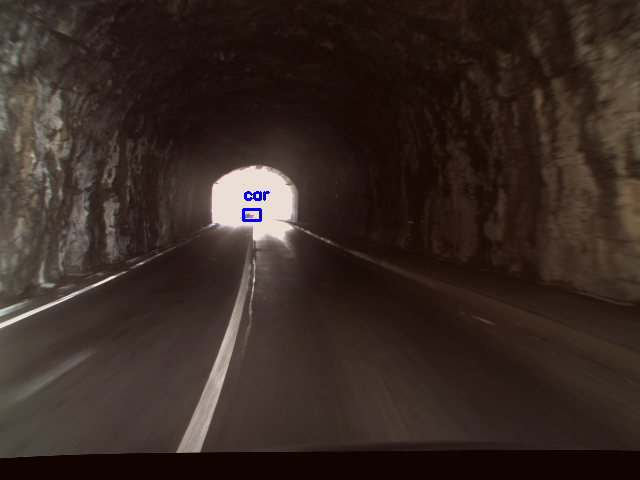} &
\includegraphics[width=\fredfigwidth, trim={1.5cm 1.5cm 1.5cm 1.5cm},clip]{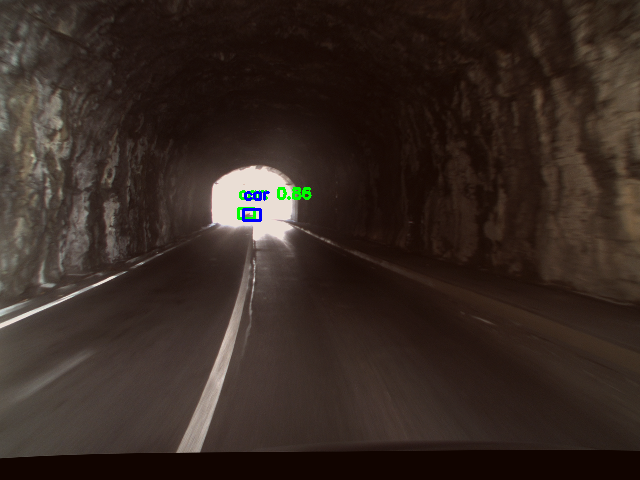} \\

\end{tabular}

\caption{Qualitative results on Hard-DSEC-DET. Detections are shown in green, ground truth in blue.}
\label{fig:supplementary_dsechard}
\end{figure*}

We report additional qualitative results. In Fig. \ref{fig:supplementary_fred} we compare outputs of the standard RGB DETR model compared with the L2 baseline and \method{}. Detections are shown in green, while ground truth boxes are represented in blue.
Similarly, in Fig. \ref{fig:supplementary_dsechard}, we show outputs for Hard-DSEC-DET.
For both datasets, it can be seen how \method{} strengthens the model into detecting difficult objects, in challenging conditions. The L2 baseline also often provides an improvement, yet still falls behind in the most difficult scenarios.

\begin{table}[b]
    \resizebox{\columnwidth}{!}{
        \begin{tabular}{lcccc}
            \toprule
            \textbf{Model} & \textbf{Train Mod.} & \textbf{Test Mod.} & \textbf{mAP$_{50:95}$} & \textbf{mAP$_{50}$} \\
            \midrule
            EA-DETR  \cite{rossi2025event} & RGB+E & RGB+E & 14.7 & 27.2 \\
            DETR \cite{carion2020end} & E & E & 12.0 & 25.8 \\
            DAGr \cite{Gehrig24nature} & E & E & 14.0 & - \\
            Faster-RCNN \cite{ren2015faster} & RGB & RGB & 18.2 & 35.4 \\
            RetinaNet  \cite{lin2017focal} & RGB & RGB & 16.6 & 30.5 \\
            CenterNet \cite{zhou2019objects} & RGB & RGB & 10.4 & 35.1 \\
            YOLOv7-E6E \cite{wang2022yolov7} & RGB & RGB & 18.2 & 31.5 \\
            YOLOv5-L \cite{yolov5} & RGB & RGB & 20.9 & 33.2 \\
            DETR \cite{carion2020end} & RGB & RGB & 27.7 & 50.6 \\
            DETR$_{L2}$ (\textit{Ours}) & RGB+E & RGB & 29.1 (\textit{+1.4}) & 50.3 (\textit{-0.3}) \\
            DETR$_{\method{}}$ (\textit{Ours}) & RGB+E & RGB & 28.3 (\textit{+0.6}) & \textbf{52.2} (\textit{+1.6})\\
            \bottomrule
        \end{tabular}
    }
    \caption{Object detection results on DSEC-DET in-domain split.}
    \label{tab:indomain_dsec}
\end{table}

\section{DSEC-DET Results}
\label{sec:dsec_det}

Alongside the hard test split, DSEC-DET also contains an in-domain split, where train and test scenarios are balanced domain-wise, resulting less challenging for RGB-based models. In fact, the difficulty in this split is not explicitly linked to light intensity or weather of the scene.
Nonetheless, as a control experiment, we report the results obtained by \method{} also on this in-domain test split.
Results are presented in Tab. \ref{tab:indomain_dsec}.
Interestingly, as observed for FRED Canonical in Tab. \ref{tab:fredcanonical} of the main paper, adopting event data as privileged information also improves robustness when domain shifts are not present, with \method{} improving on the L2 baseline.

\section{Qualitative Results Semantic Segmentation}
\label{sec:qualitative_segmentation}
In Fig. \ref{fig:supplementary_segmentation} we show segmentation results for Cityscapes Adverse. We highlighted details of interest with yellow circles, where the improvements obtained by \method{} are most noticeable. Interestingly, \method{} is capable of recovering errors in challenging conditions, such as sky at dawn, wet road and snowy vegetation.
\newcommand{\circledincludegraphics}[5]{%
\begin{tikzpicture}
    \node[anchor=south west, inner sep=0] (img) at (0, 0) 
        {\includegraphics[width=#1]{#2}};

    \begin{scope}[x=(img.south east), y=(img.north west)]
        
        \draw[yellow, ultra thick] (#3,#4) circle (#5); 
        
    \end{scope}
\end{tikzpicture}
}

\begin{figure*}[t]
\centering

\begin{tabular}{cccc}

\textbf{RGB} &
\textbf{SegFormer} &
\textbf{\method{}} &
\textbf{Label} \\

\circledincludegraphics{\cityfigwidth}{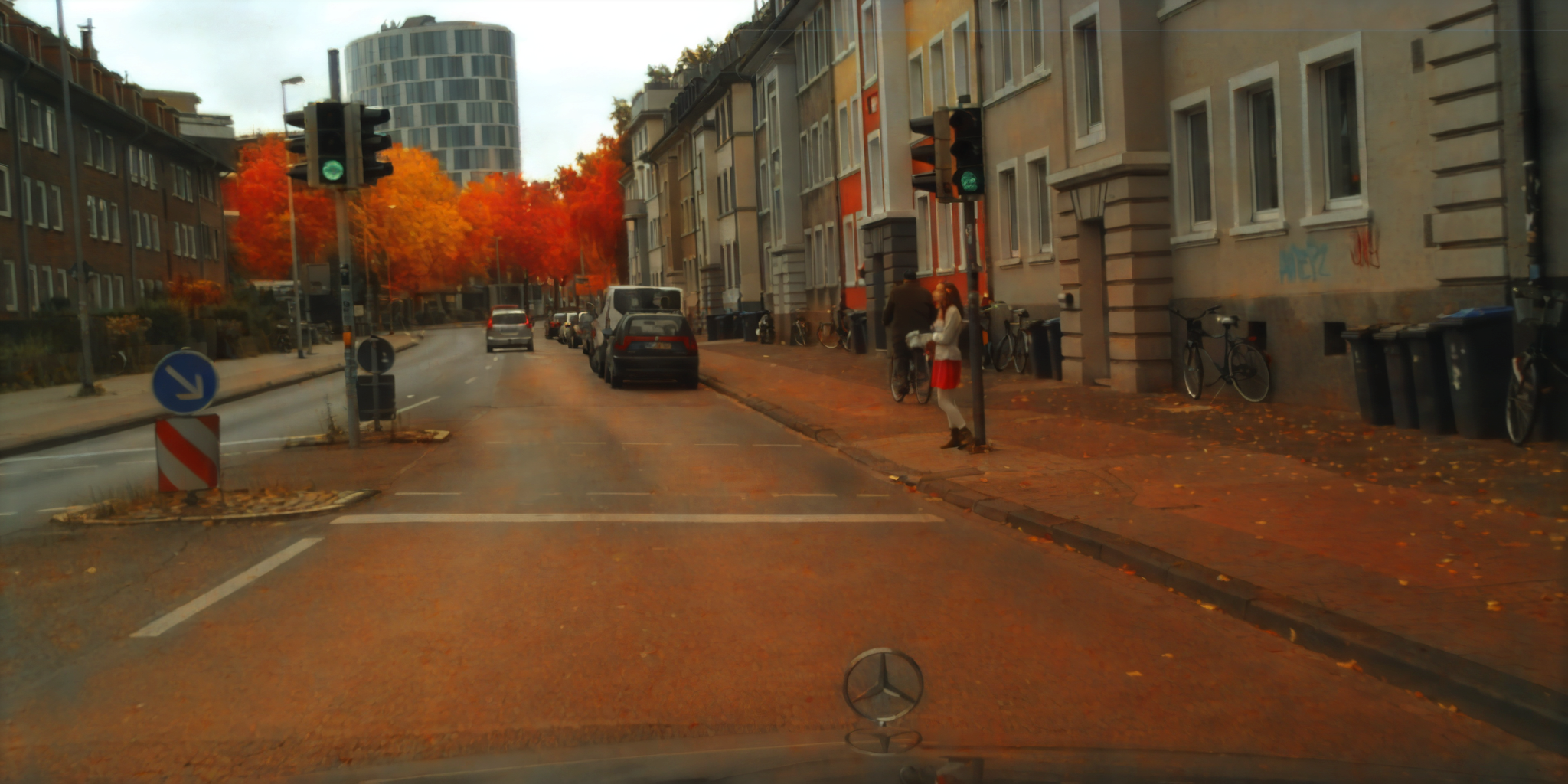}{0.85}{0.35}{0.12} &
\circledincludegraphics{\cityfigwidth}{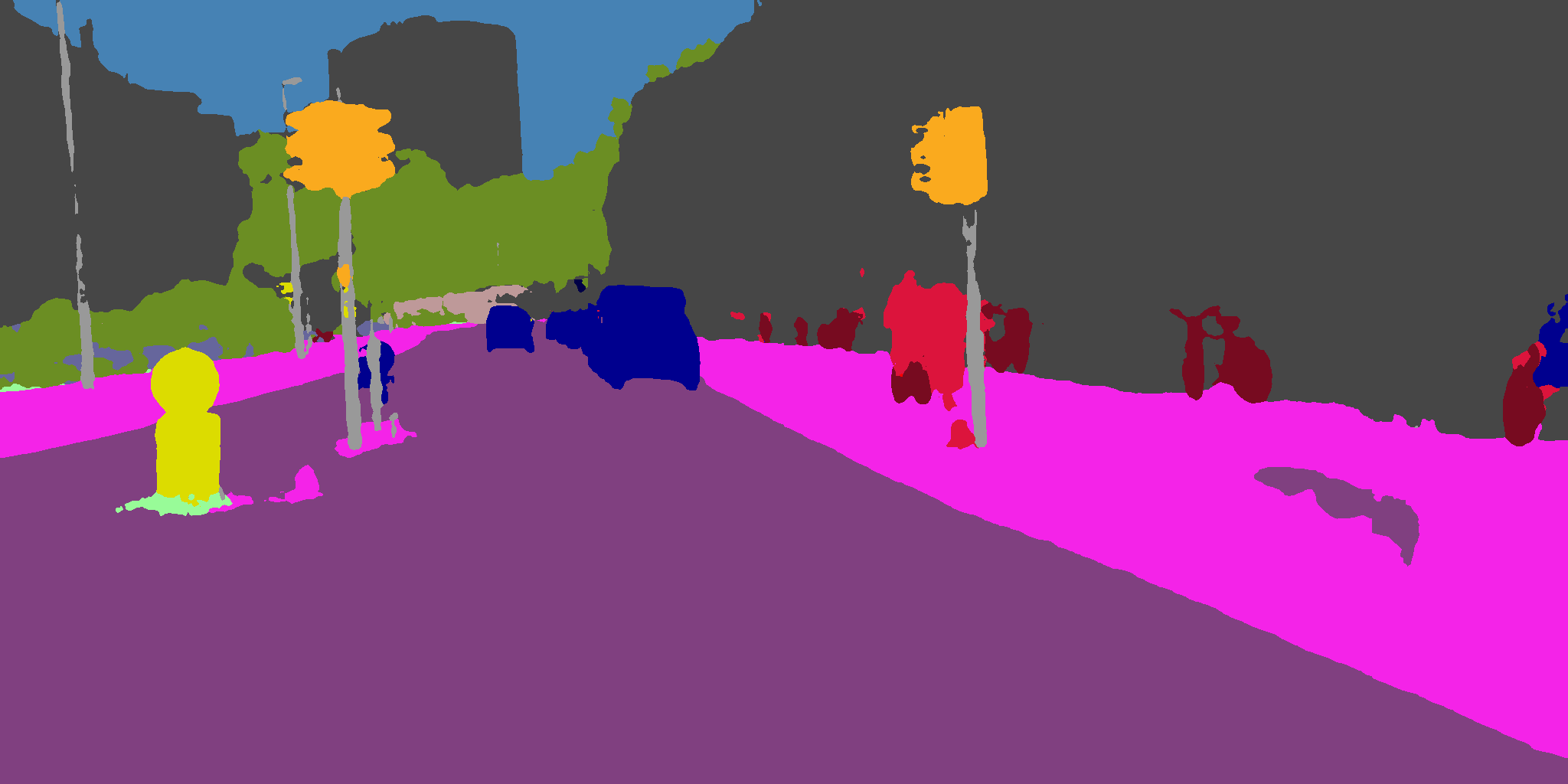}{0.85}{0.35}{0.12} &
\circledincludegraphics{\cityfigwidth}{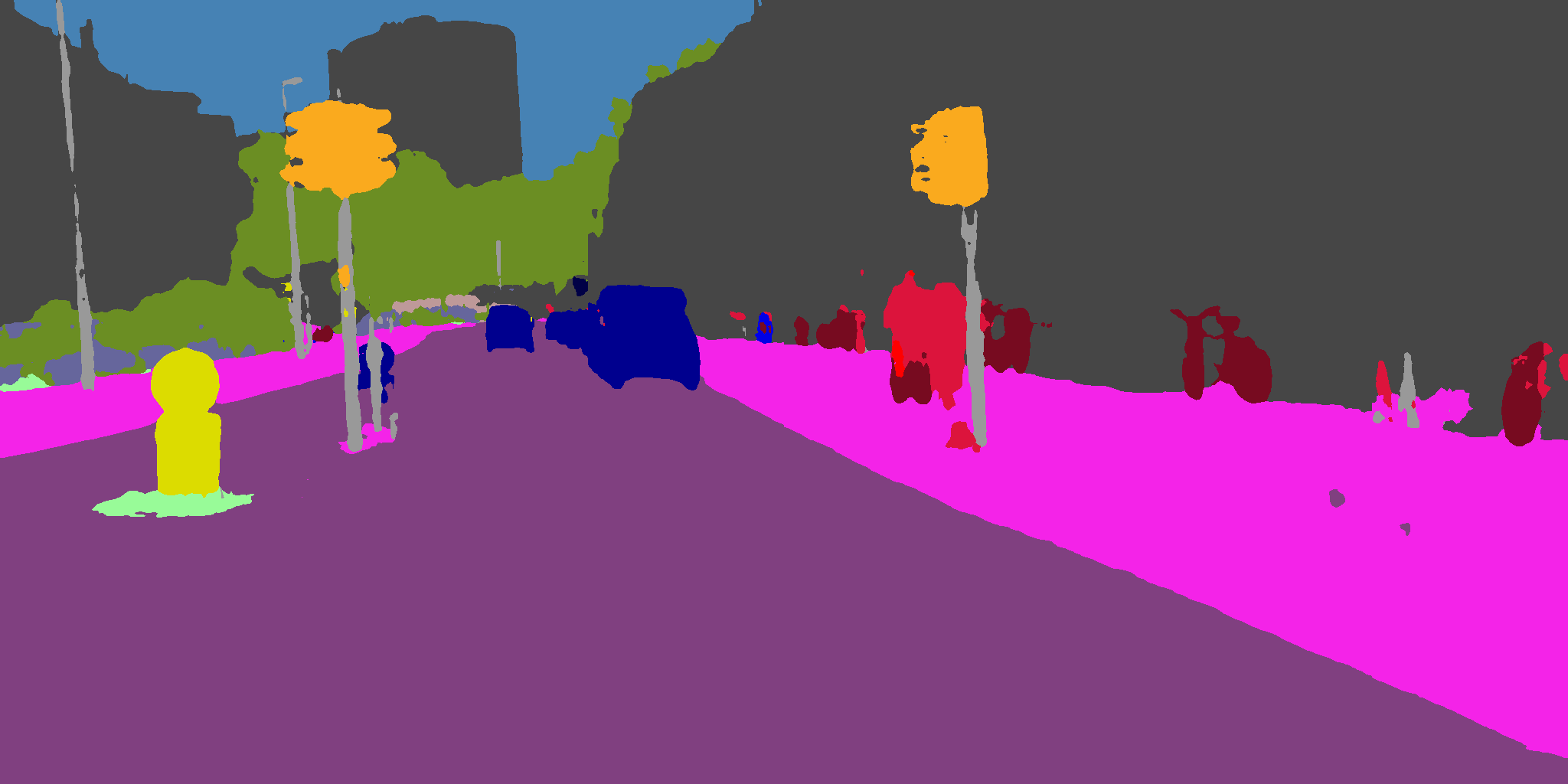}{0.85}{0.35}{0.12} &
\circledincludegraphics{\cityfigwidth}{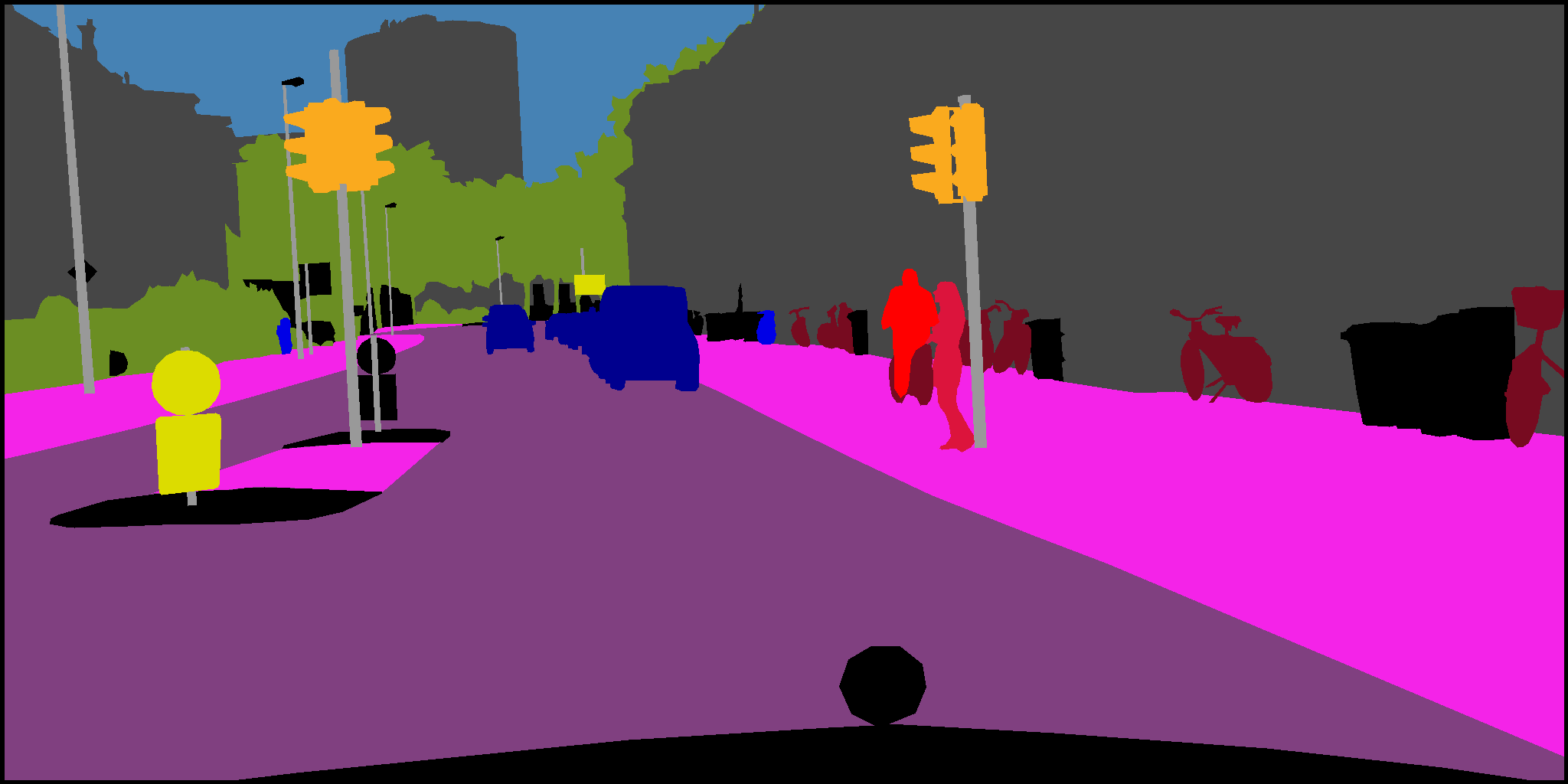}{0.85}{0.35}{0.12} \\

\circledincludegraphics{\cityfigwidth}{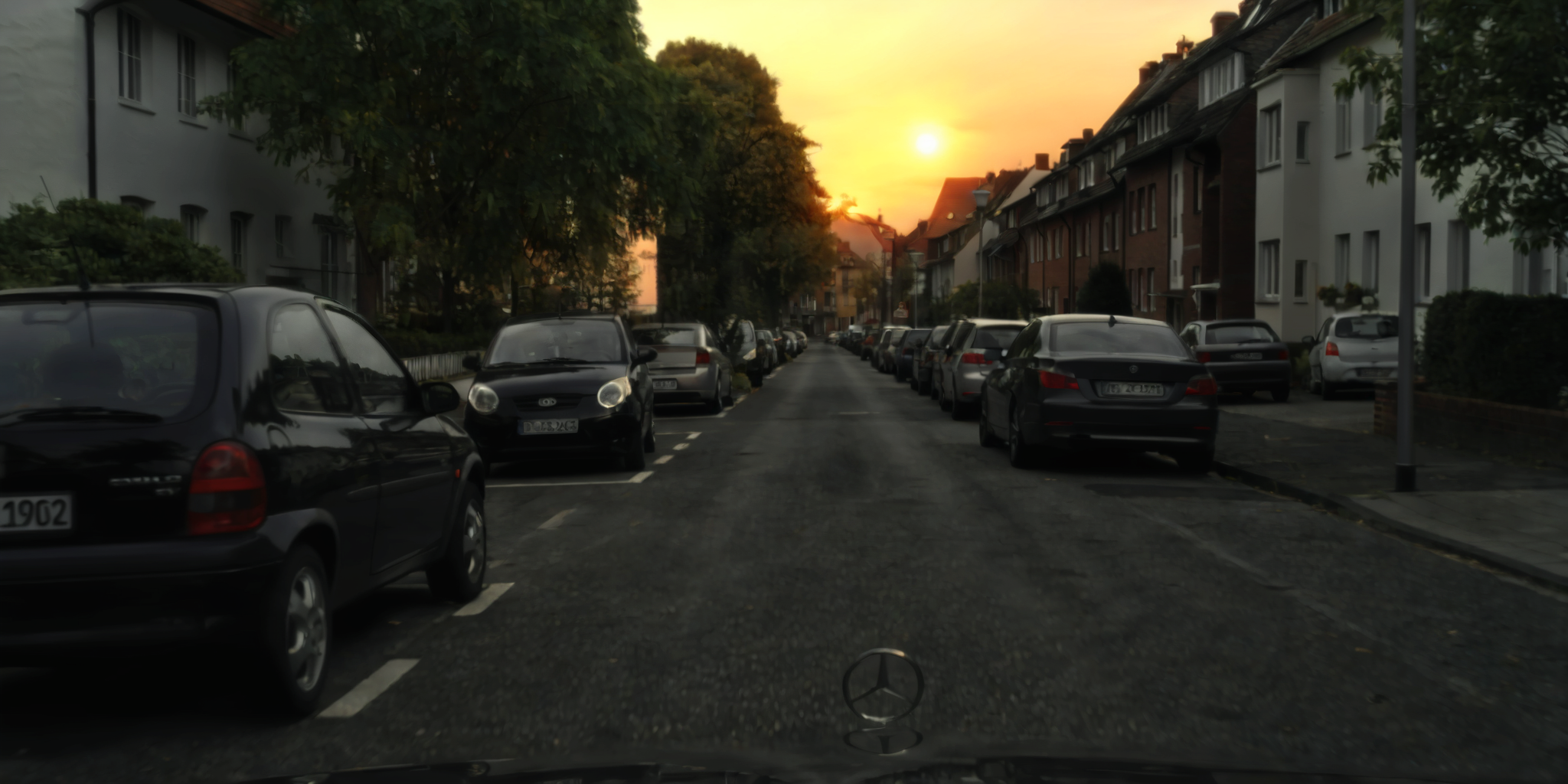}{0.6}{0.82}{0.18} &
\circledincludegraphics{\cityfigwidth}{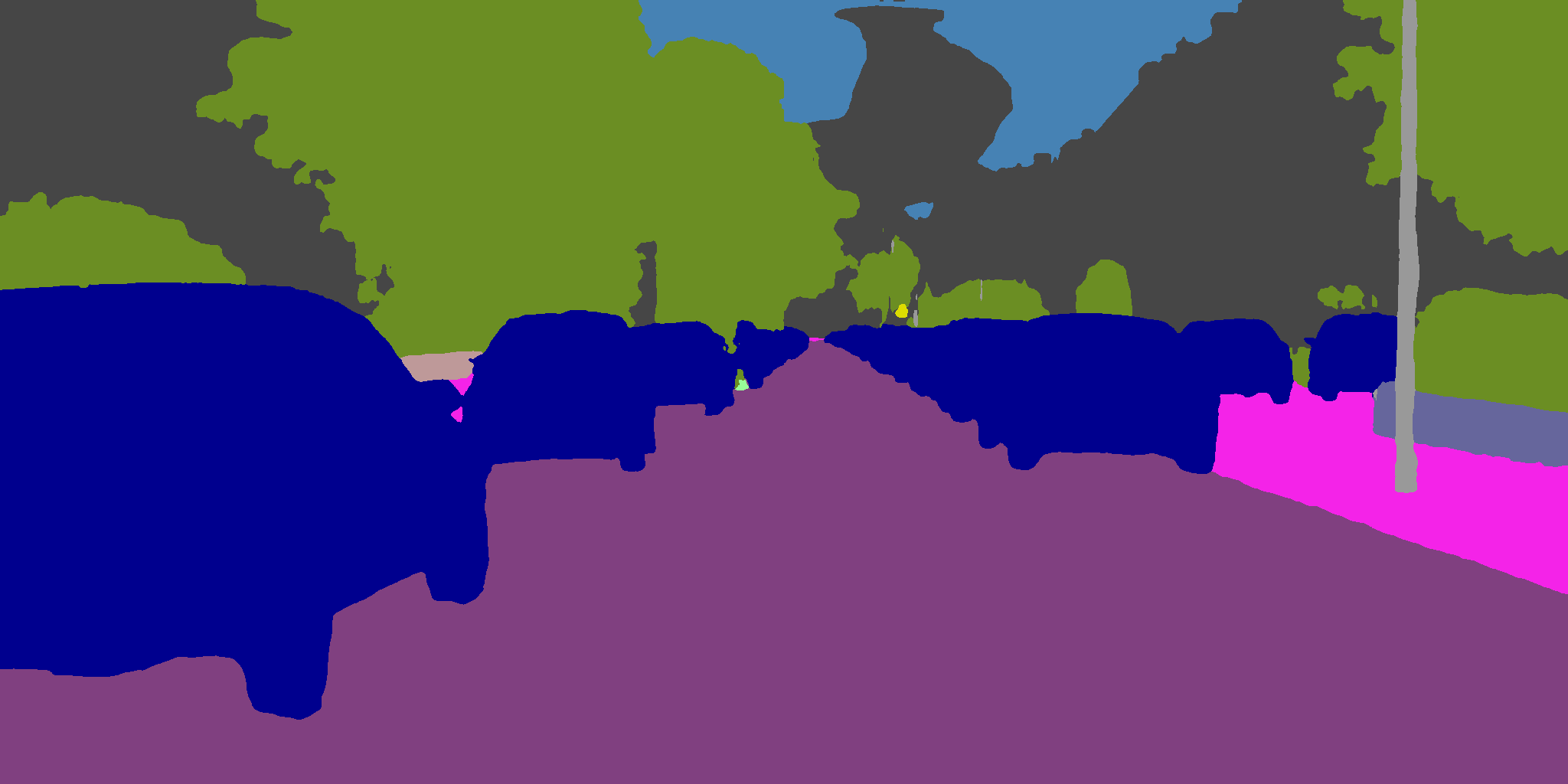}{0.6}{0.82}{0.18} &
\circledincludegraphics{\cityfigwidth}{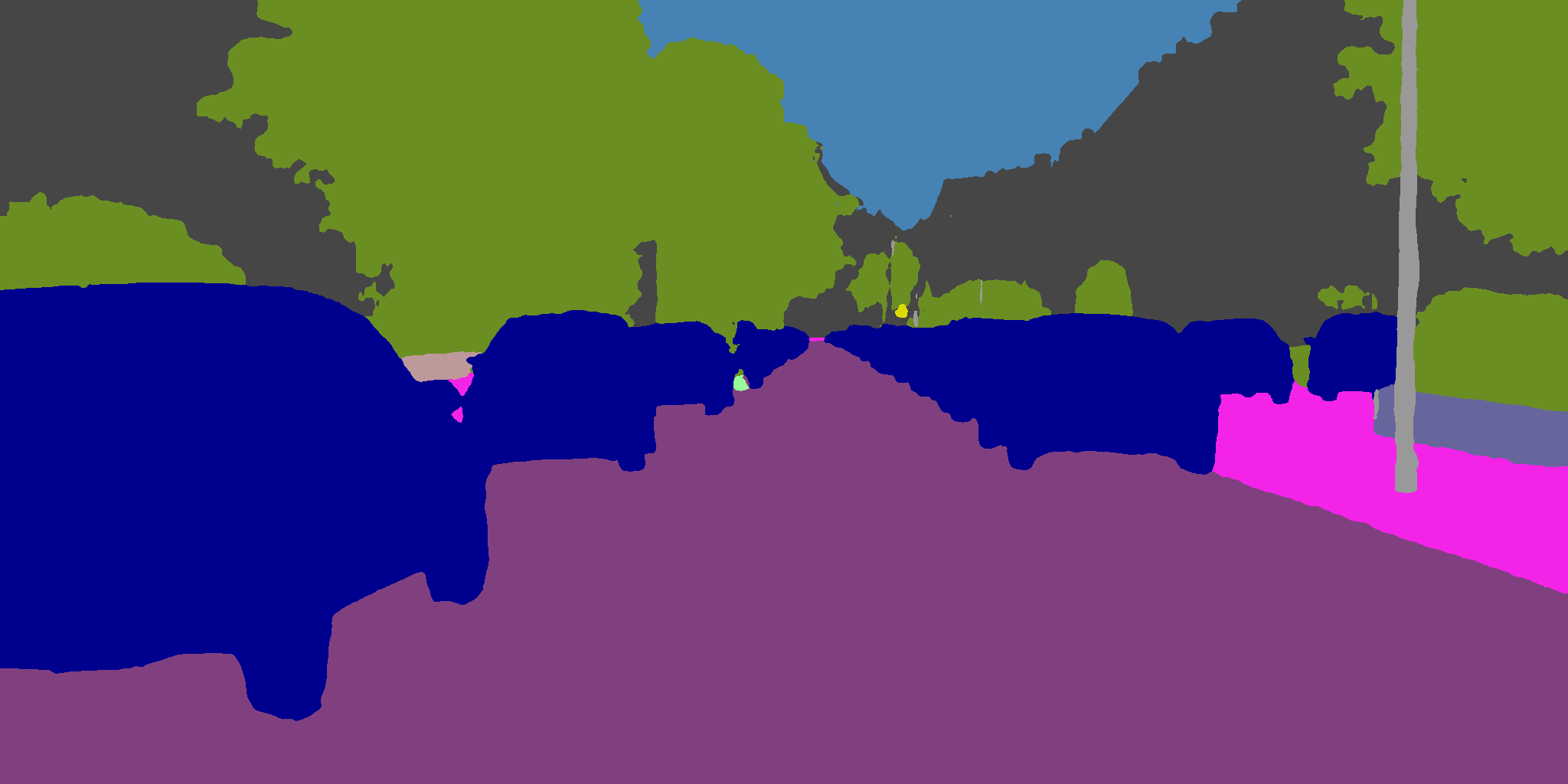}{0.6}{0.82}{0.18} &
\circledincludegraphics{\cityfigwidth}{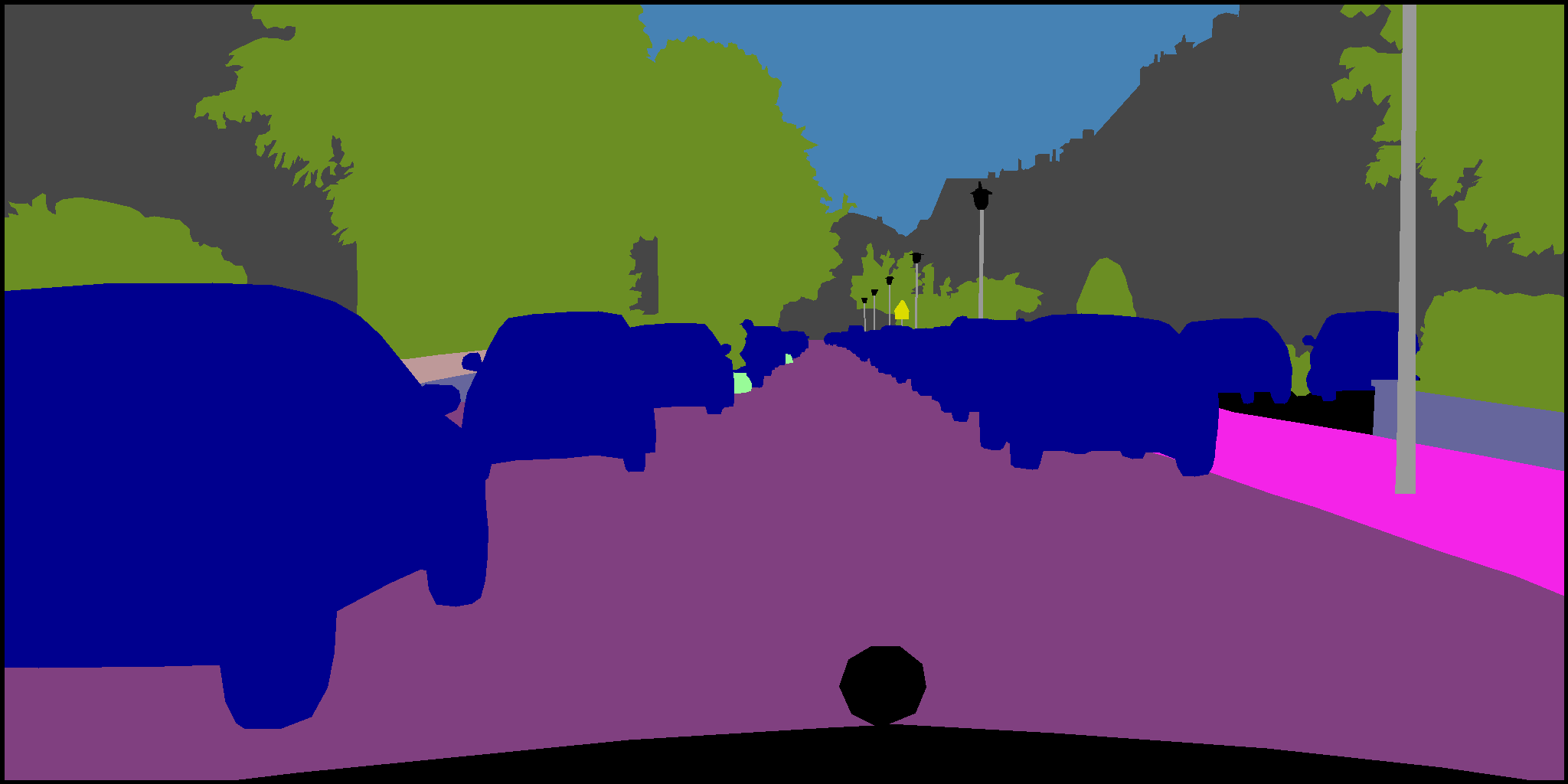}{0.6}{0.82}{0.18} \\

\circledincludegraphics{\cityfigwidth}{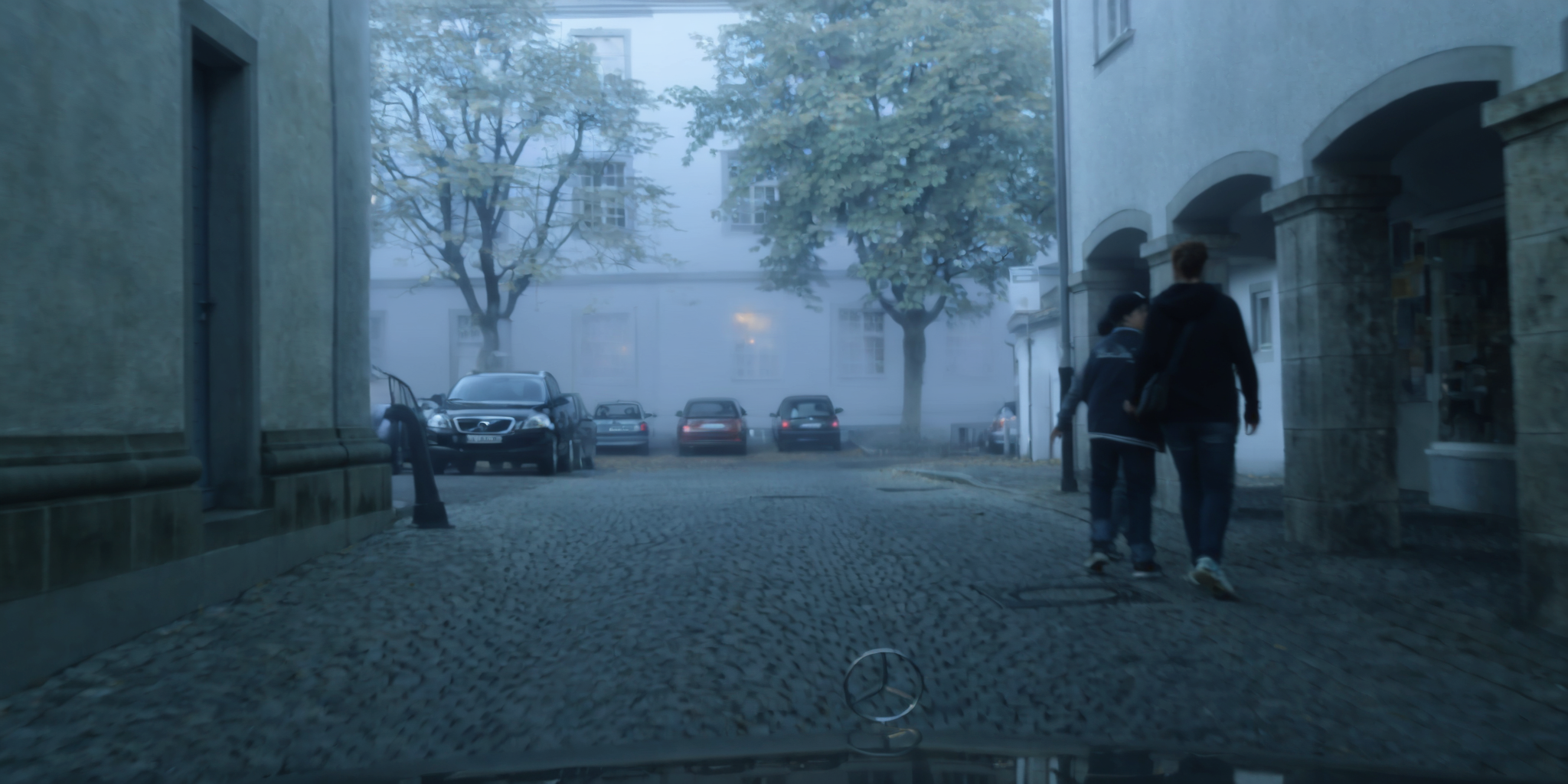}{0.2}{0.2}{0.2} &
\circledincludegraphics{\cityfigwidth}{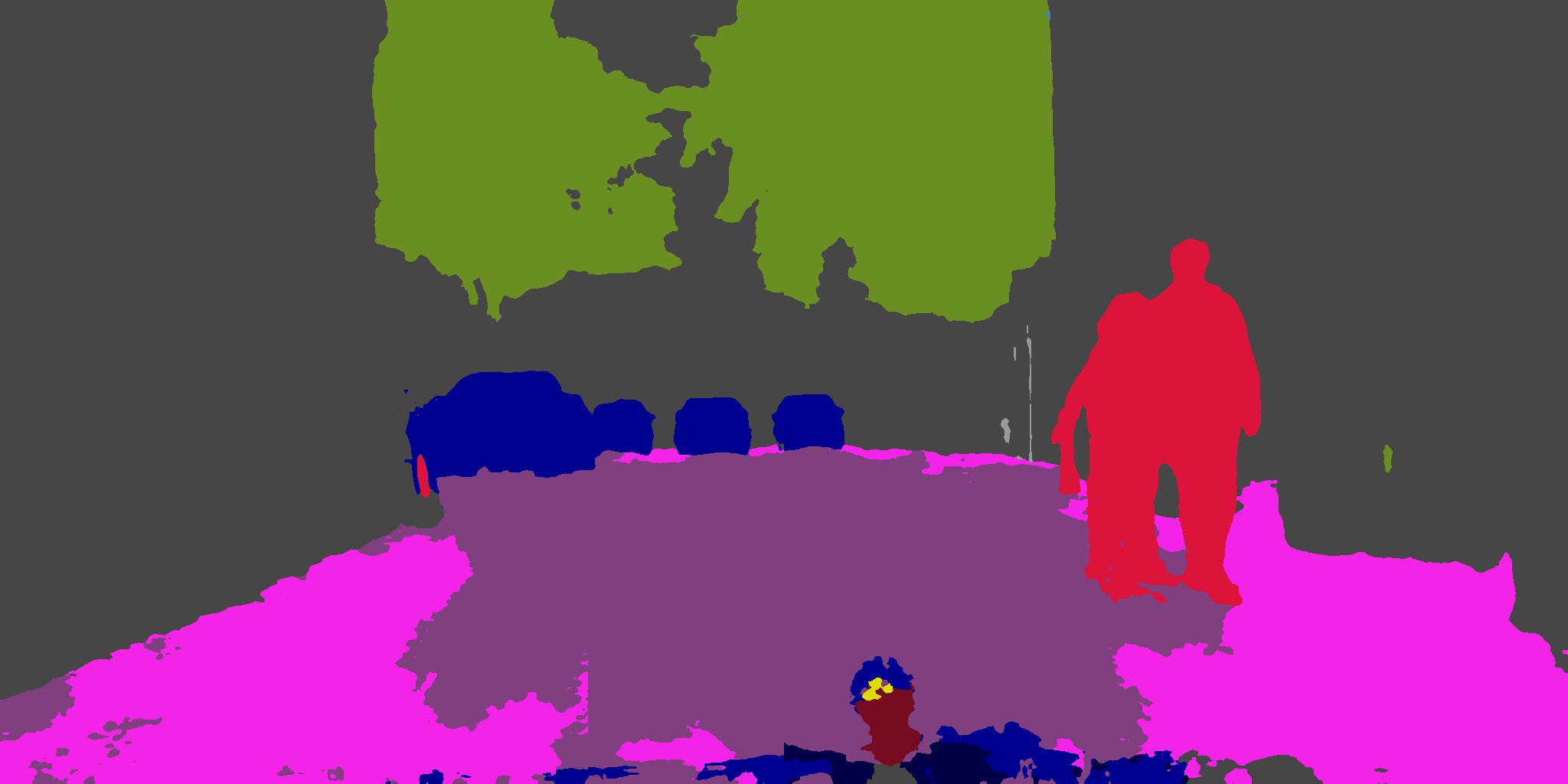}{0.2}{0.2}{0.2} &
\circledincludegraphics{\cityfigwidth}{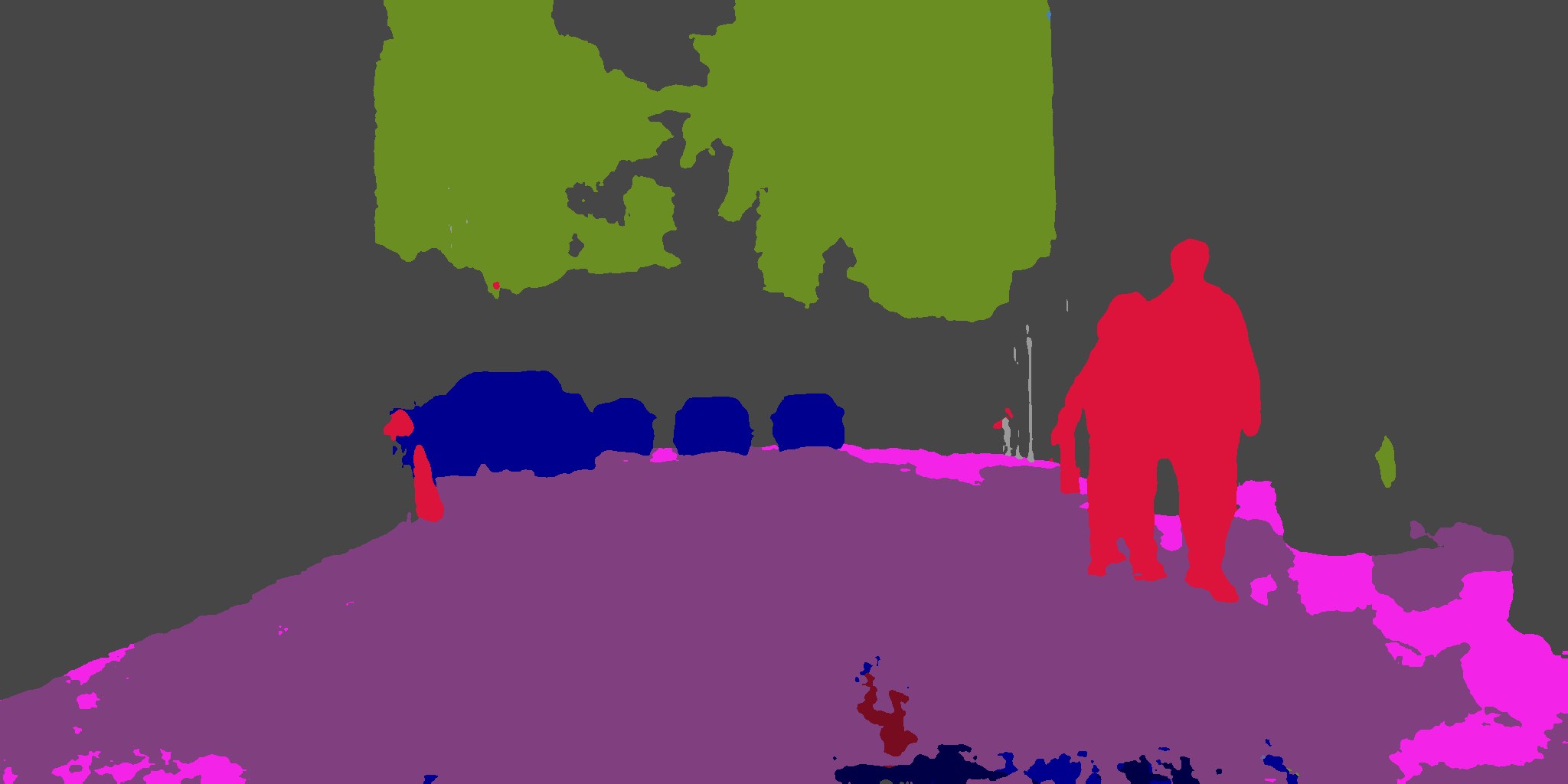}{0.2}{0.2}{0.2} &
\circledincludegraphics{\cityfigwidth}{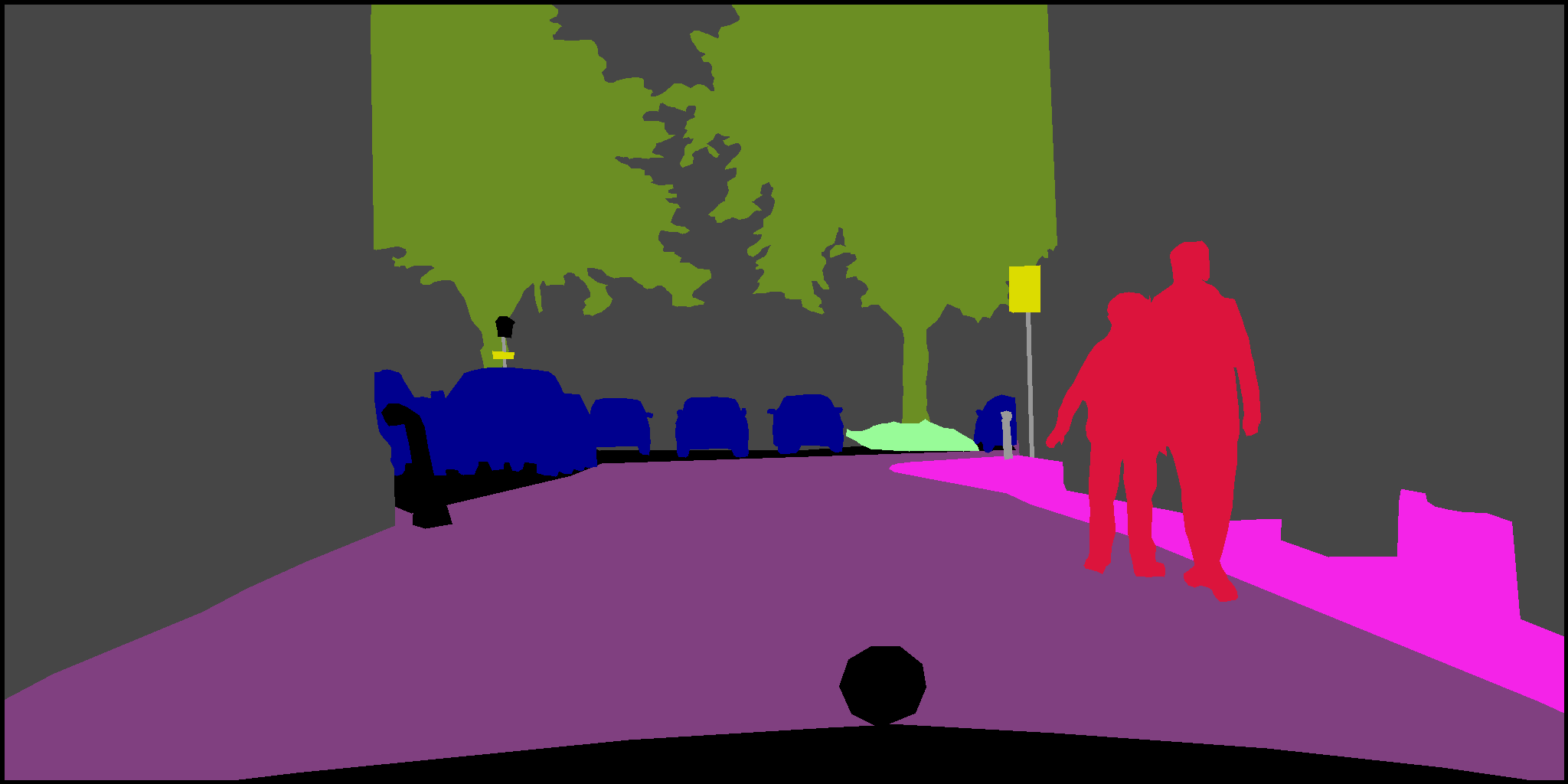}{0.2}{0.2}{0.2} \\

\circledincludegraphics{\cityfigwidth}{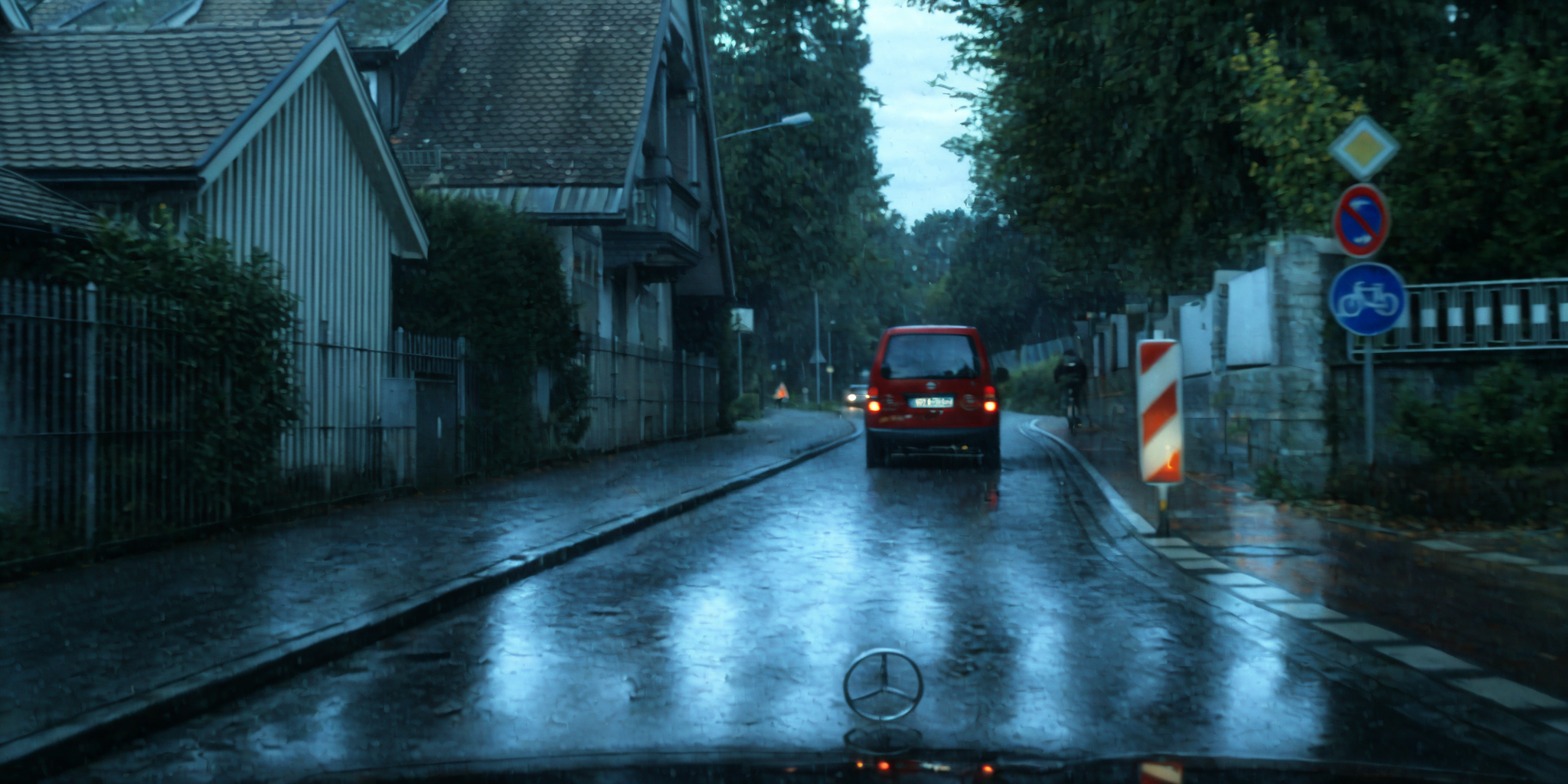}{0.55}{0.2}{0.22} &
\circledincludegraphics{\cityfigwidth}{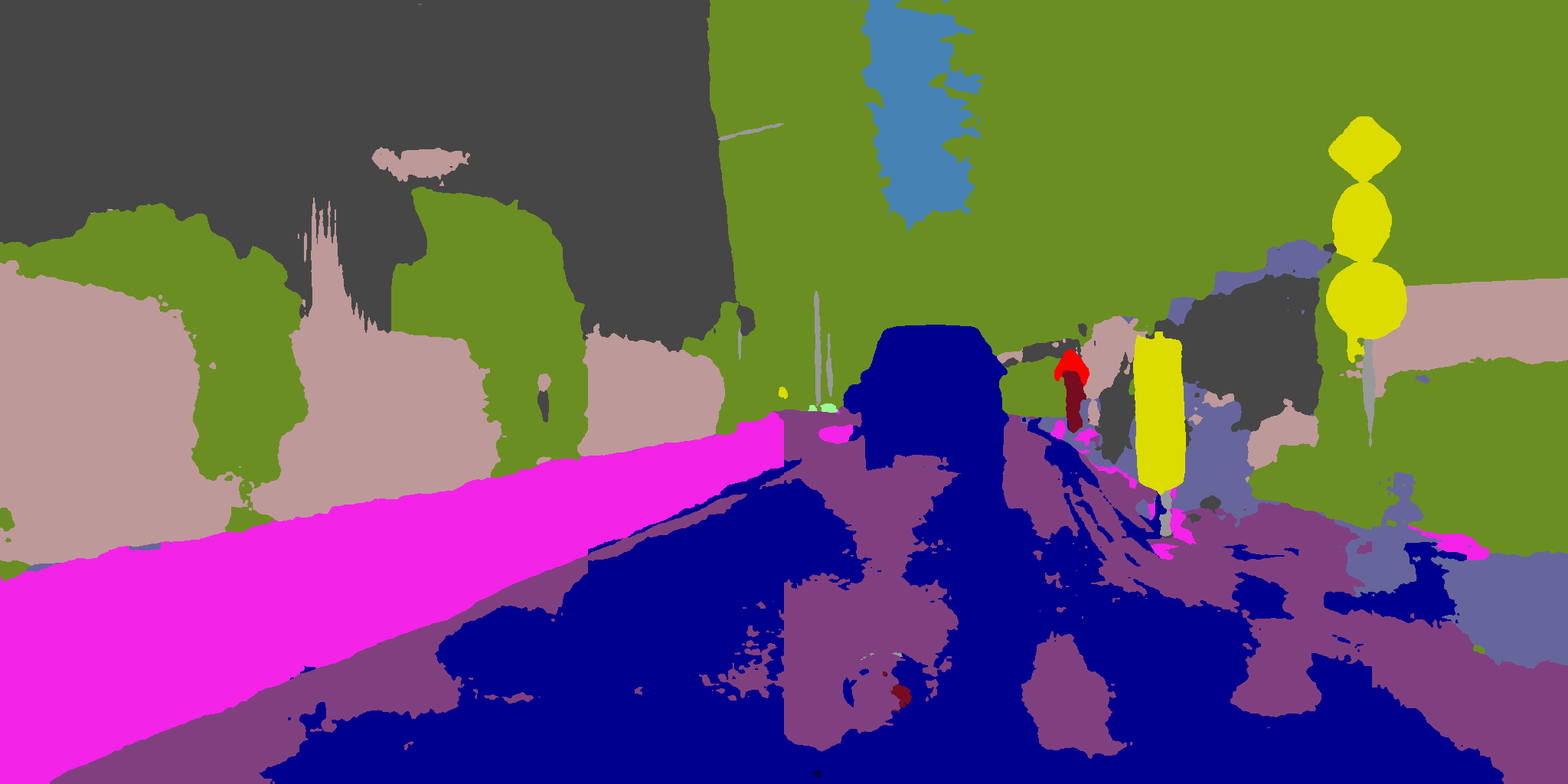}{0.55}{0.2}{0.22} &
\circledincludegraphics{\cityfigwidth}{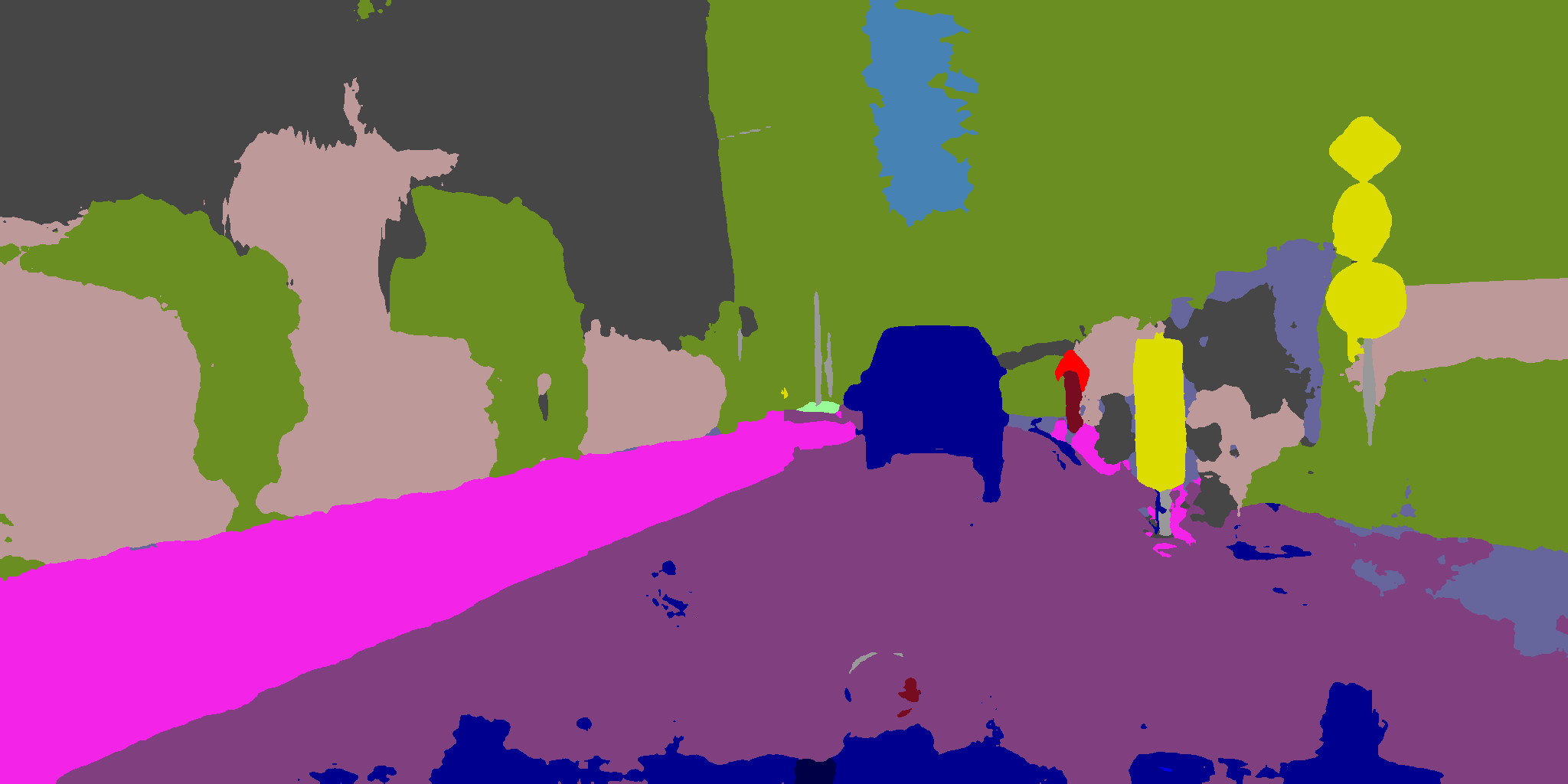}{0.55}{0.2}{0.22} &
\circledincludegraphics{\cityfigwidth}{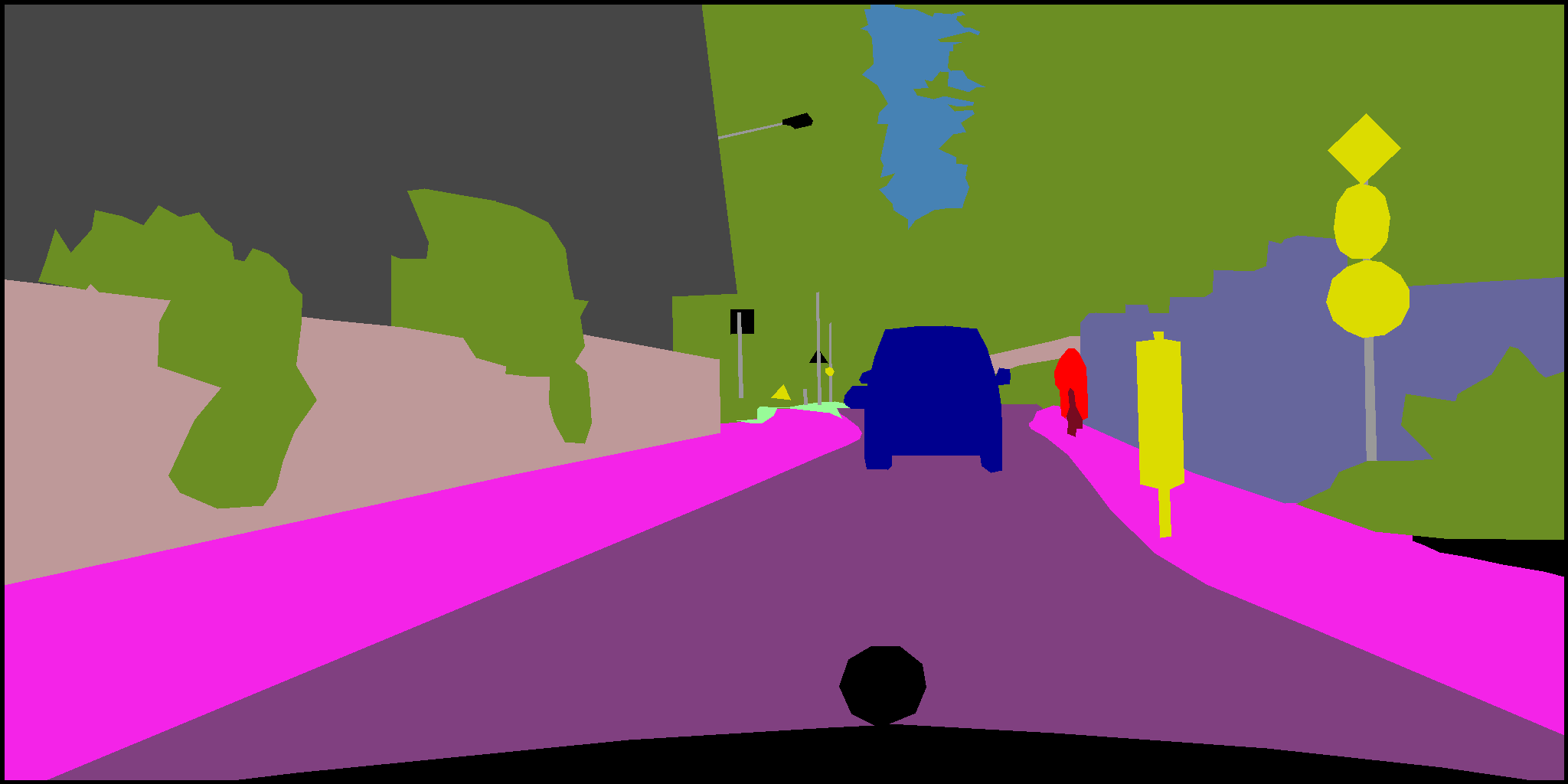}{0.55}{0.2}{0.22} \\
\circledincludegraphics{\cityfigwidth}{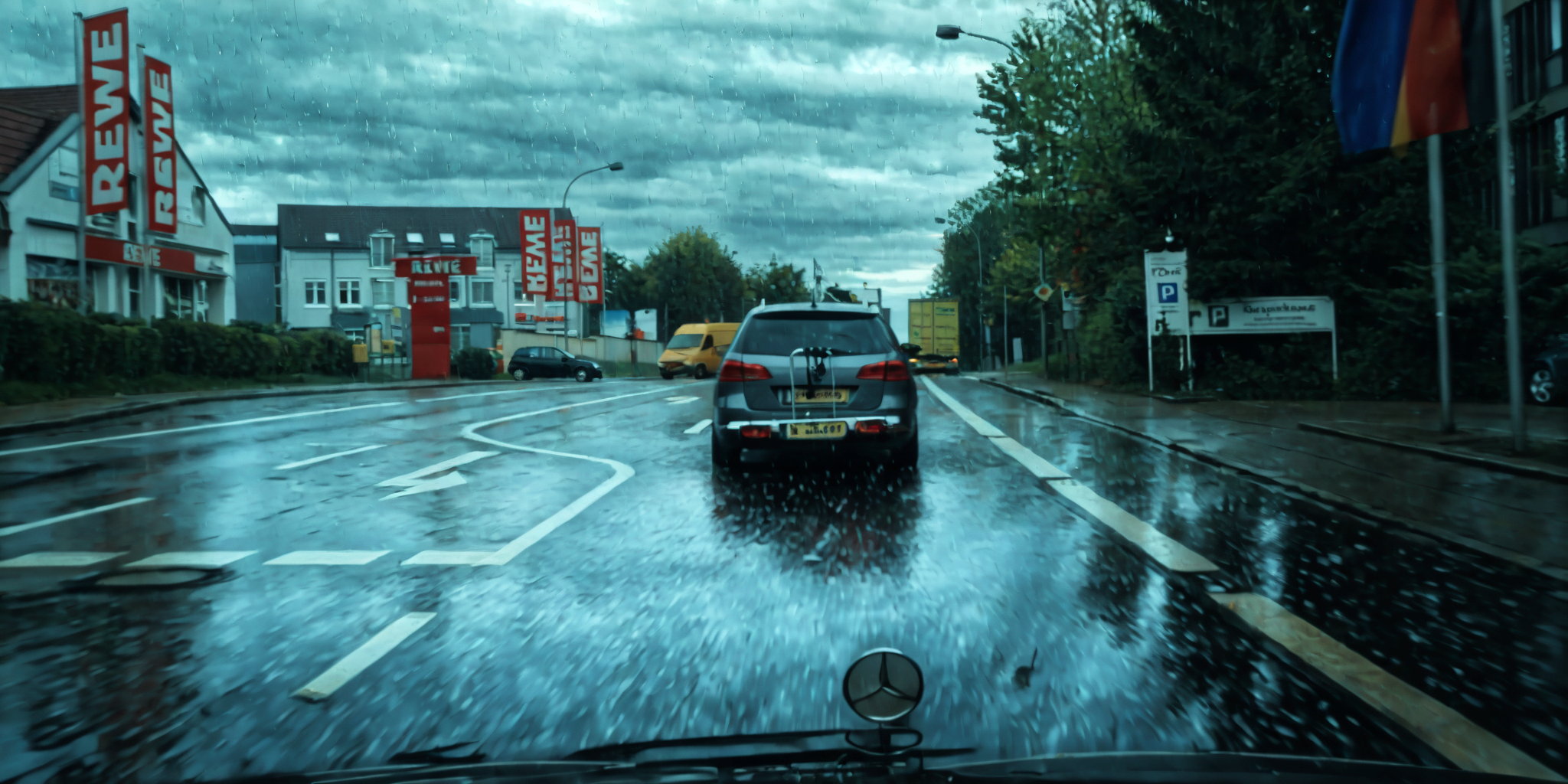}{0.65}{0.3}{0.25} &
\circledincludegraphics{\cityfigwidth}{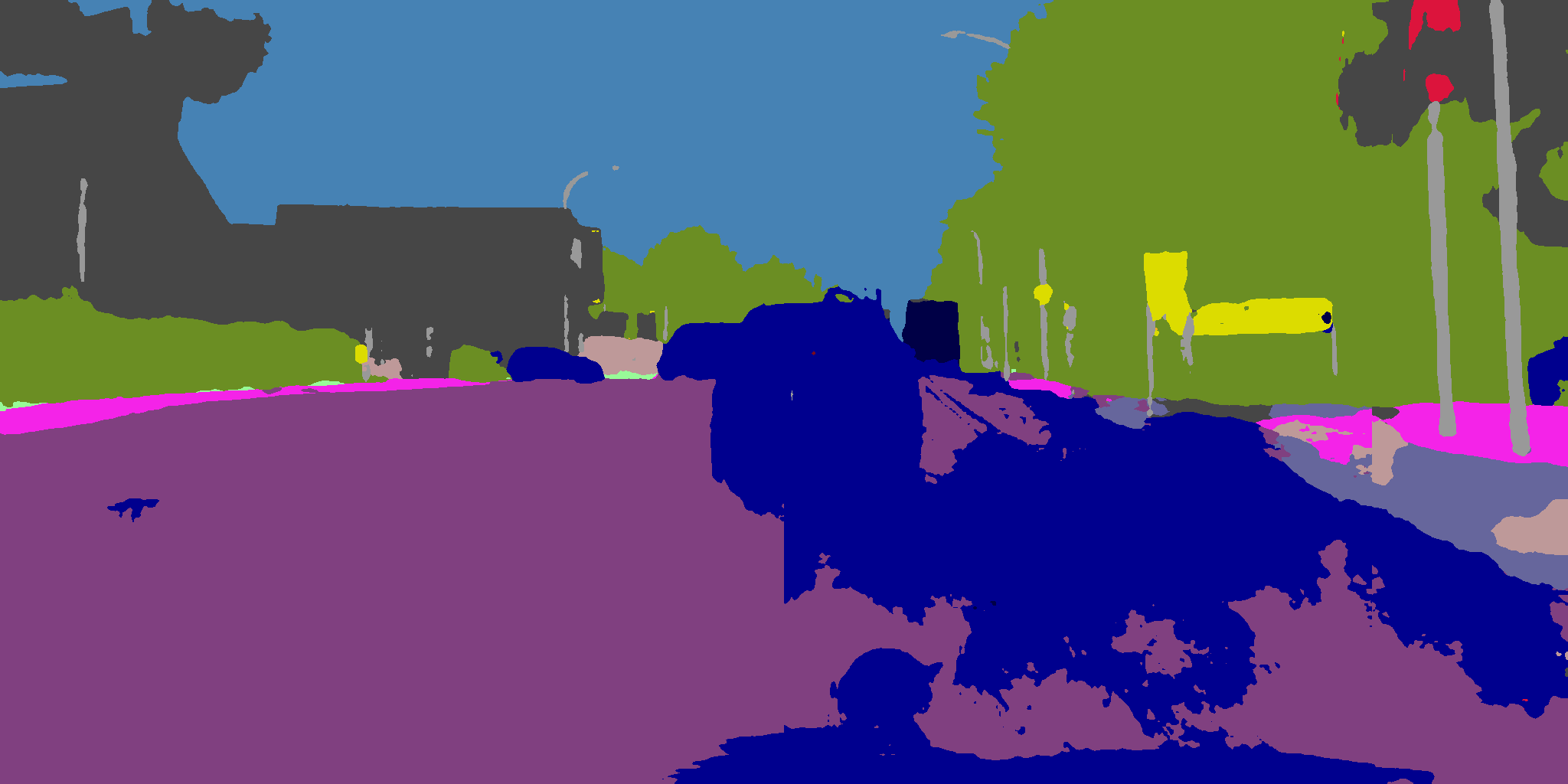}{0.65}{0.3}{0.25} &
\circledincludegraphics{\cityfigwidth}{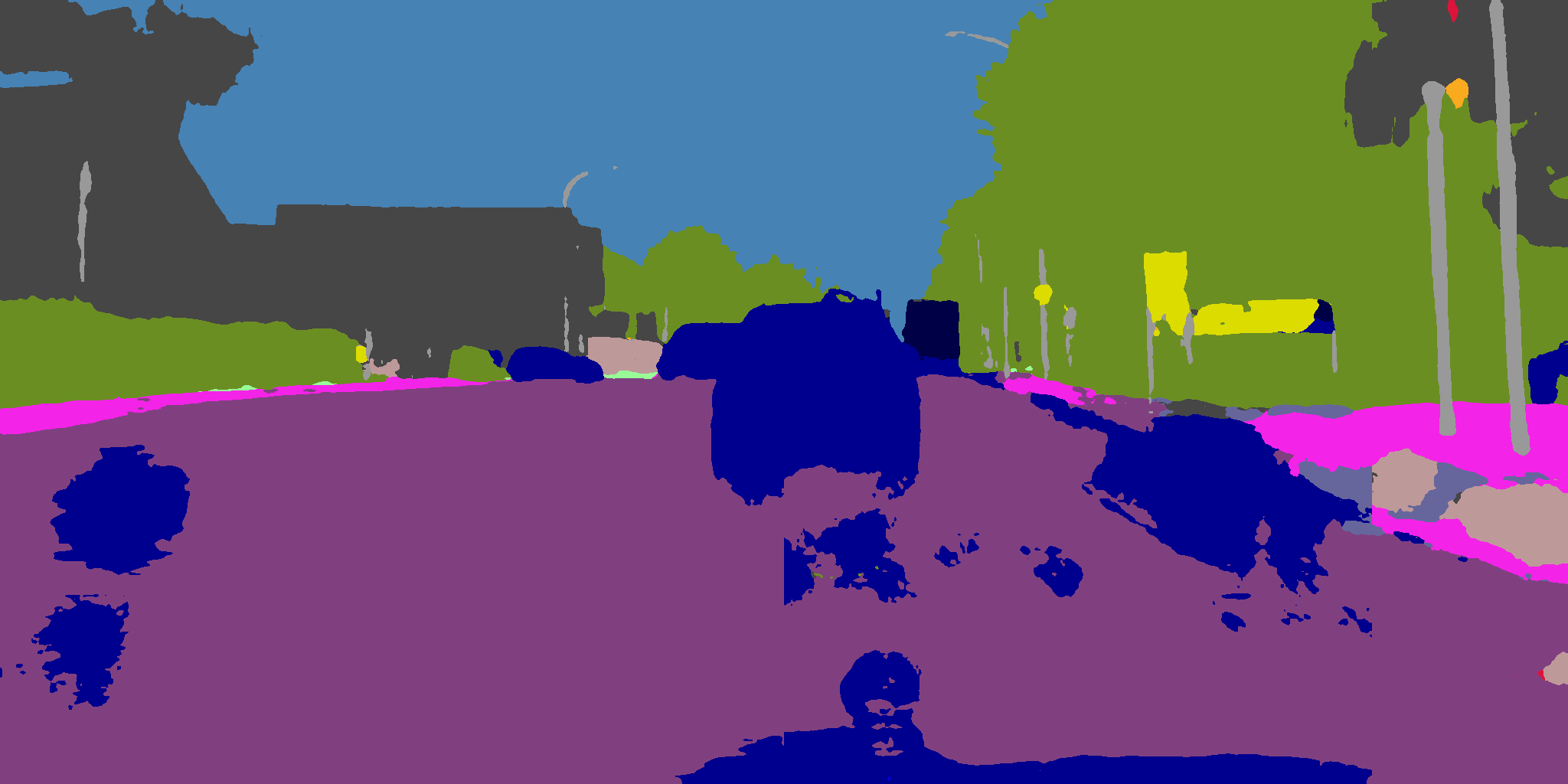}{0.65}{0.3}{0.25} &
\circledincludegraphics{\cityfigwidth}{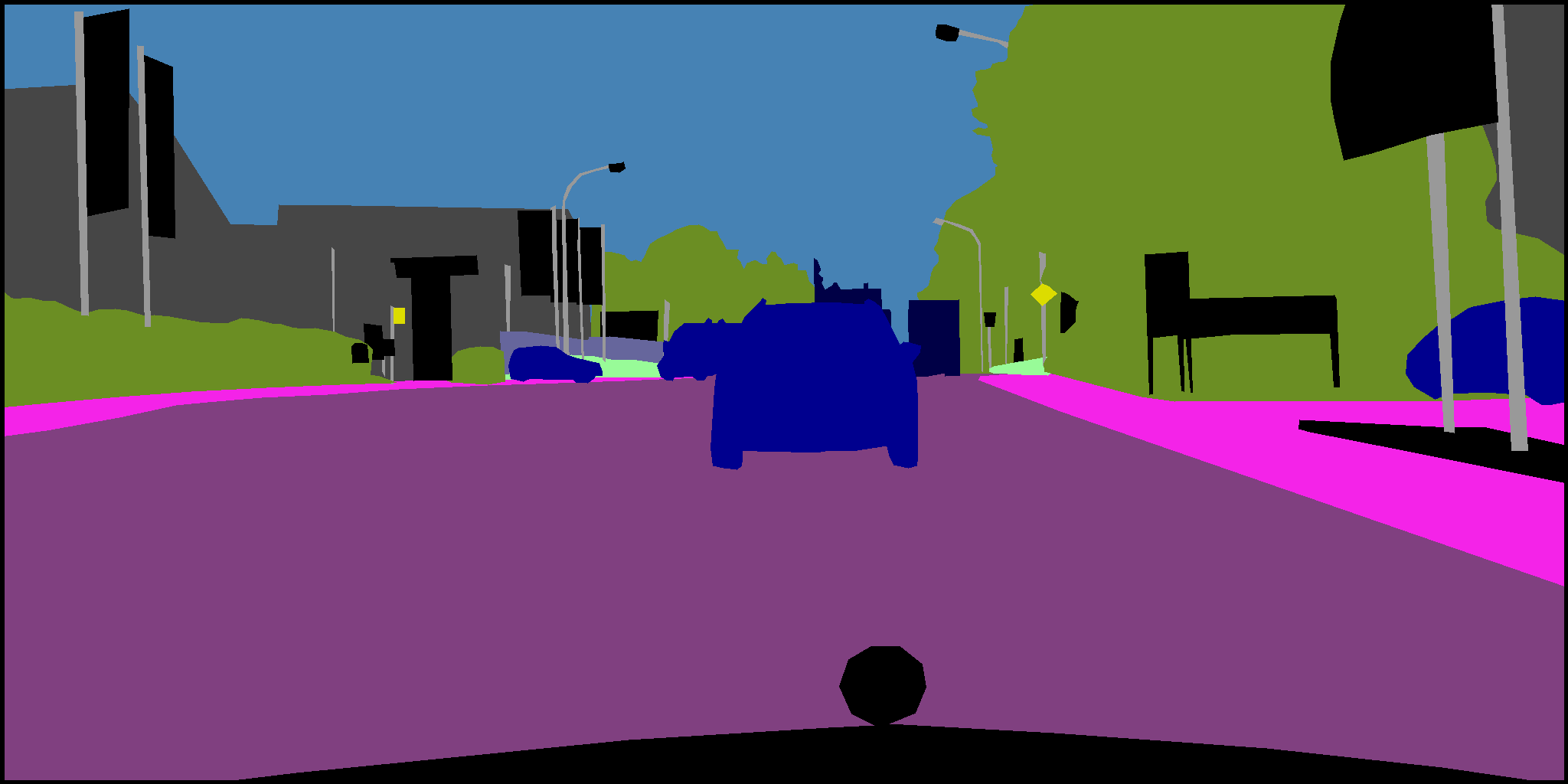}{0.65}{0.3}{0.25} \\

\circledincludegraphics{\cityfigwidth}{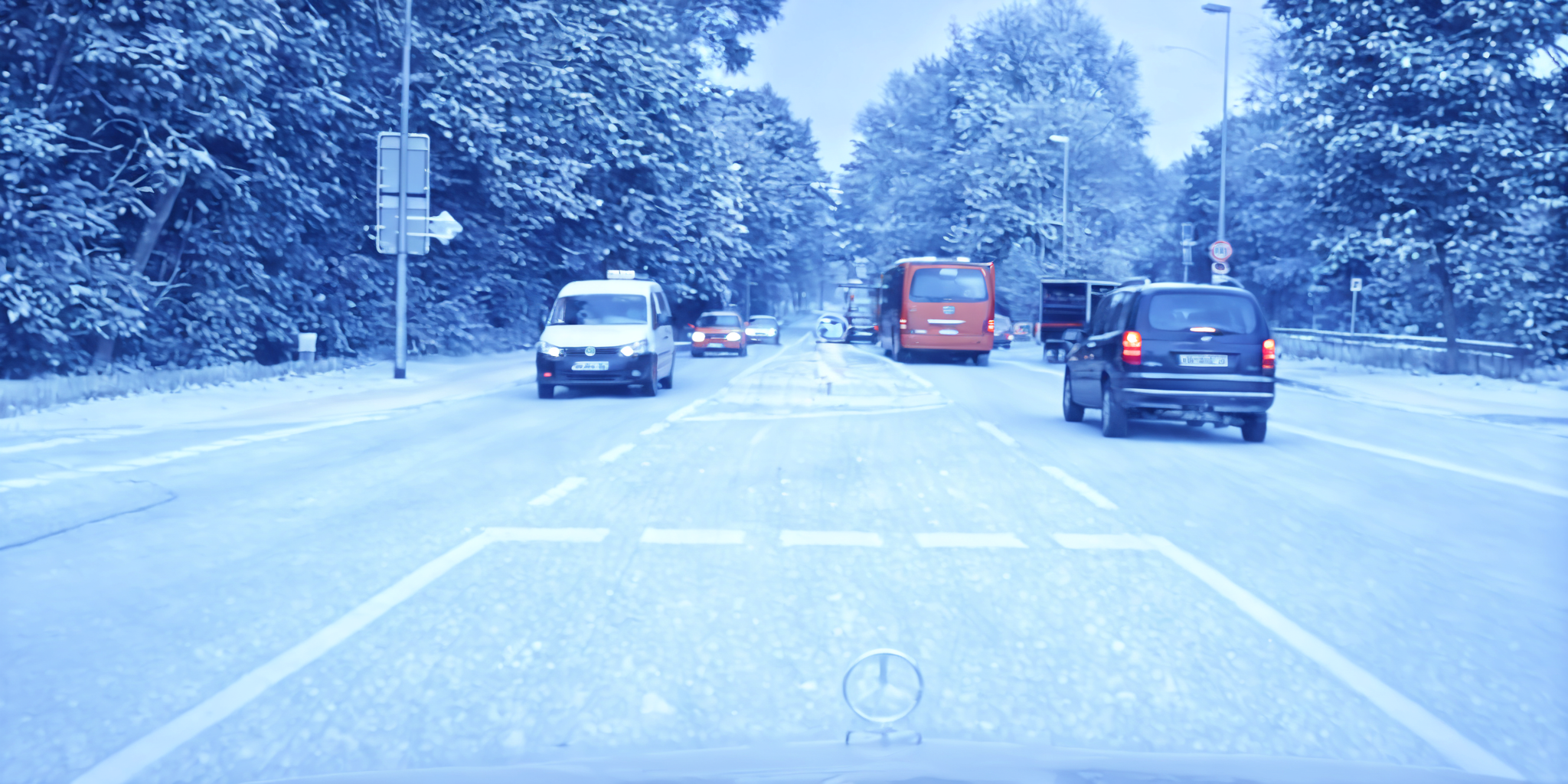}{0.25}{0.75}{0.25} &
\circledincludegraphics{\cityfigwidth}{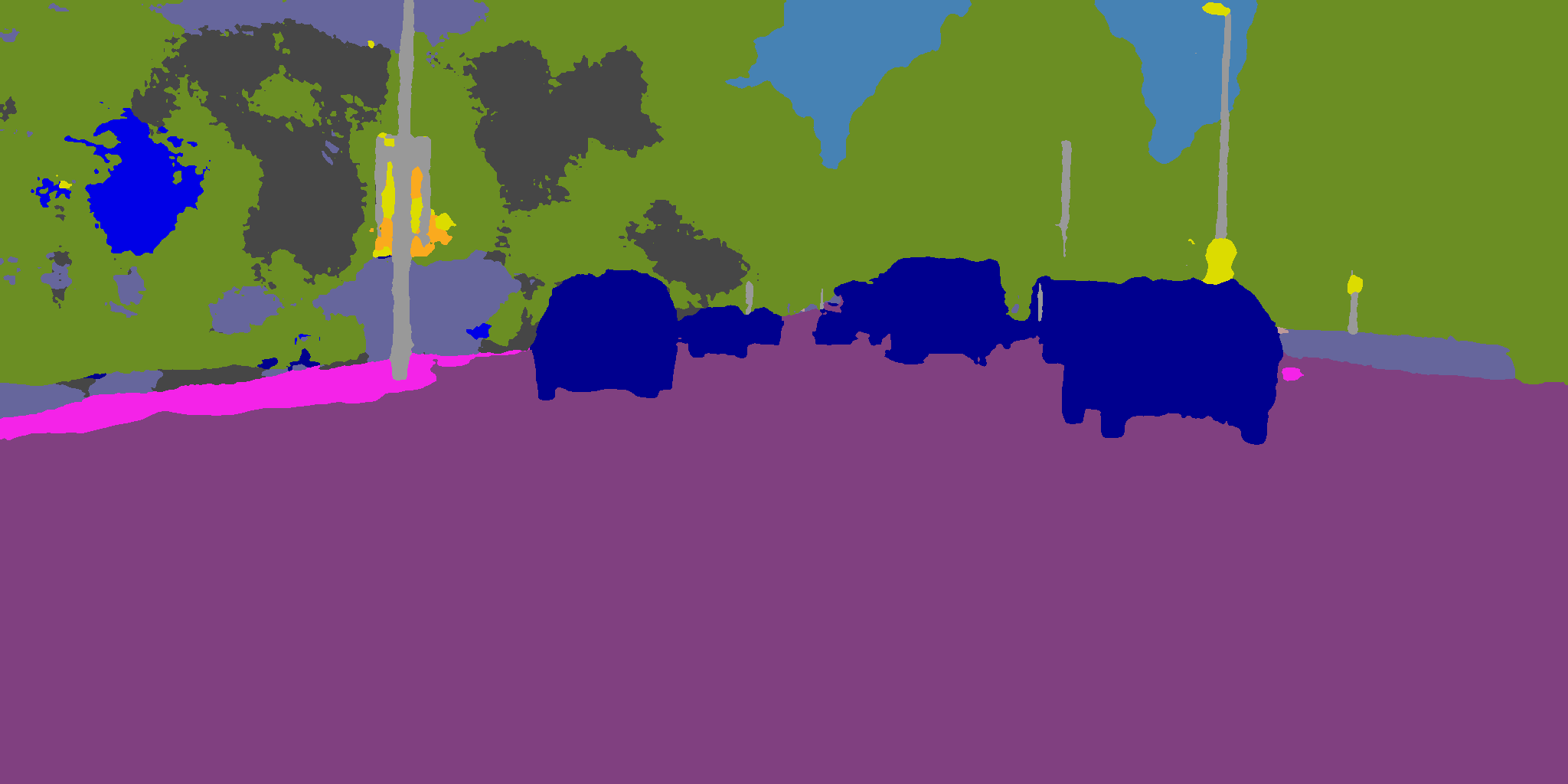}{0.25}{0.75}{0.25} &
\circledincludegraphics{\cityfigwidth}{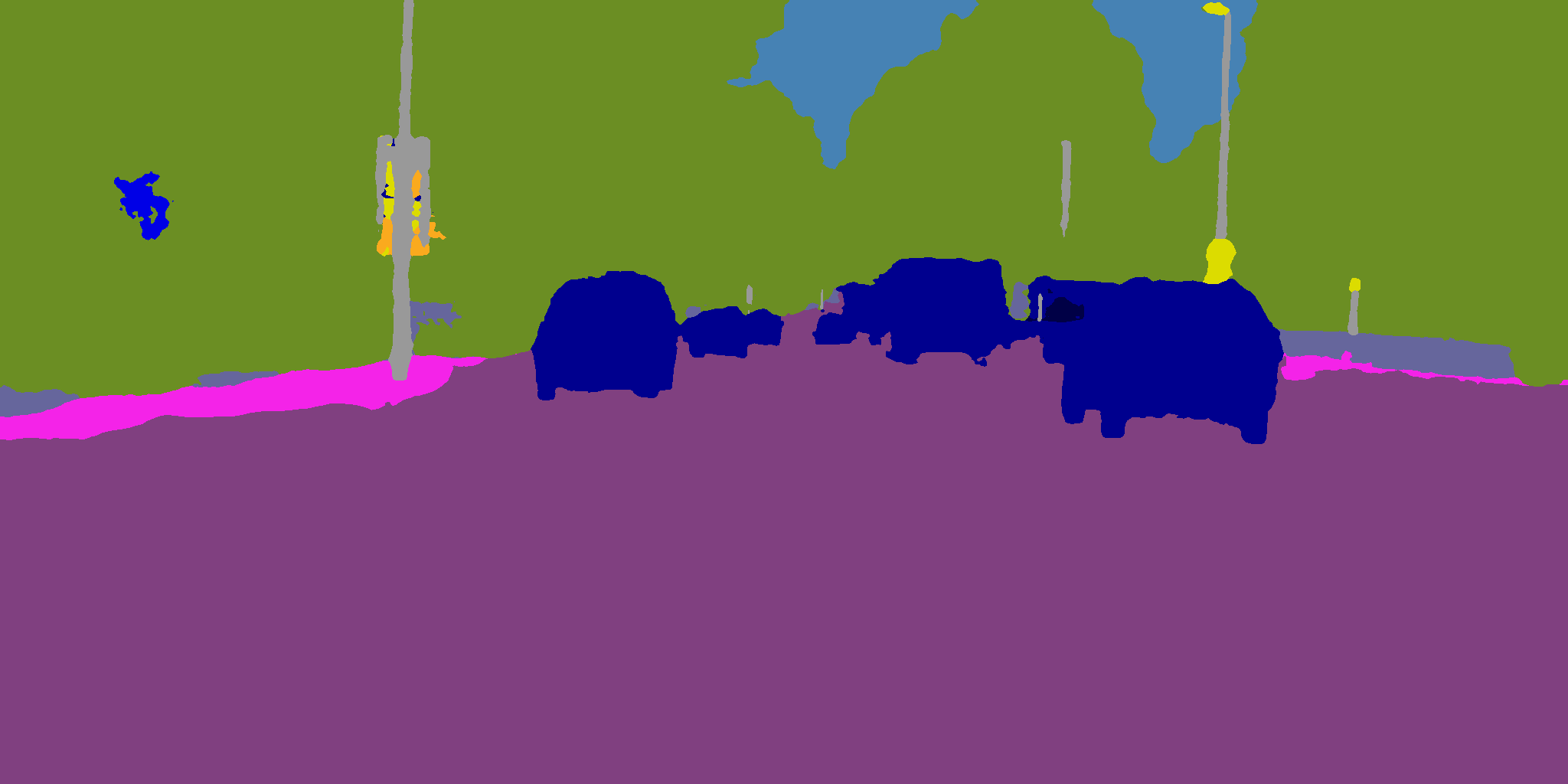}{0.25}{0.75}{0.25} &
\circledincludegraphics{\cityfigwidth}{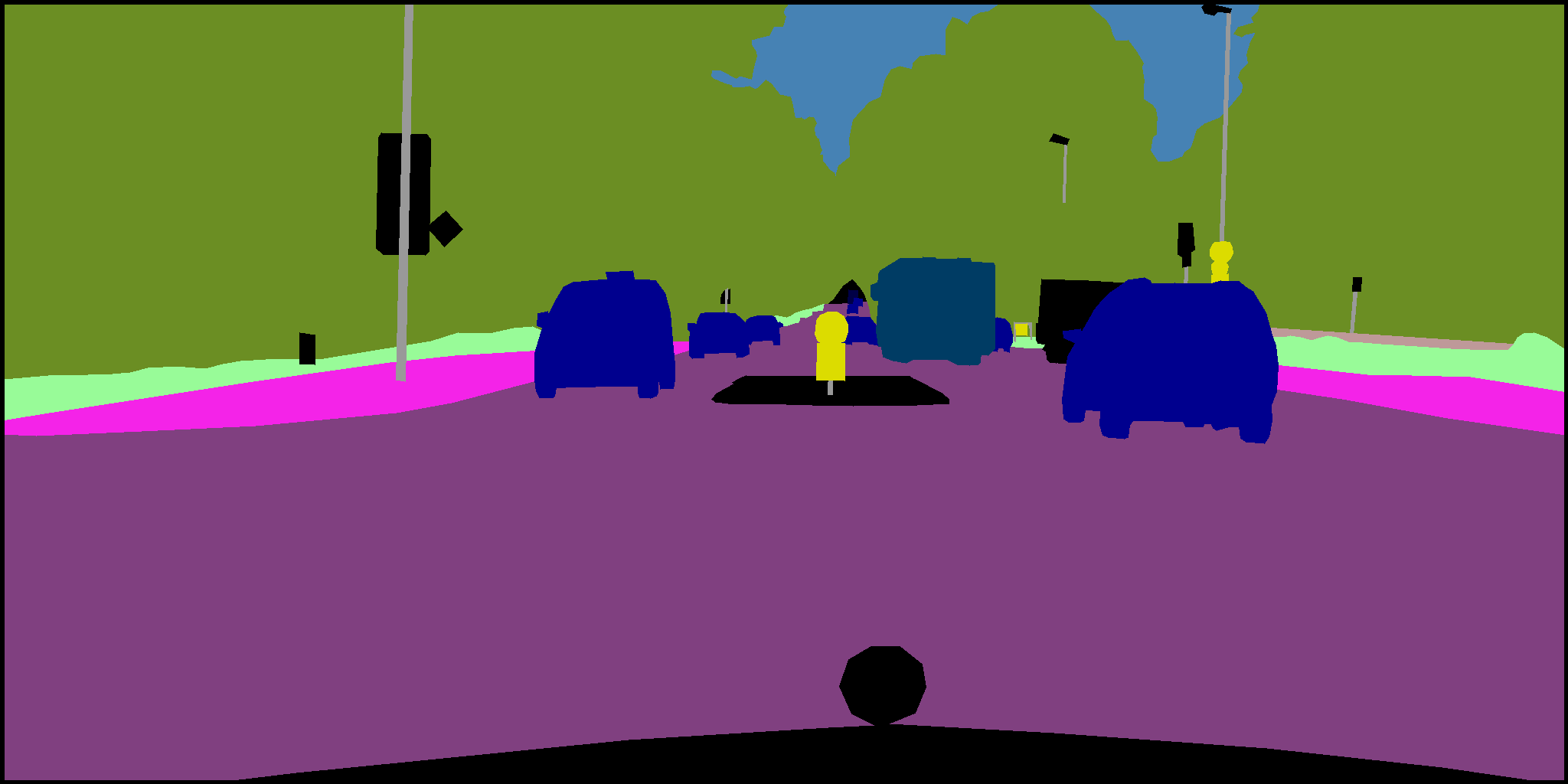}{0.25}{0.75}{0.25} \\

\circledincludegraphics{\cityfigwidth}{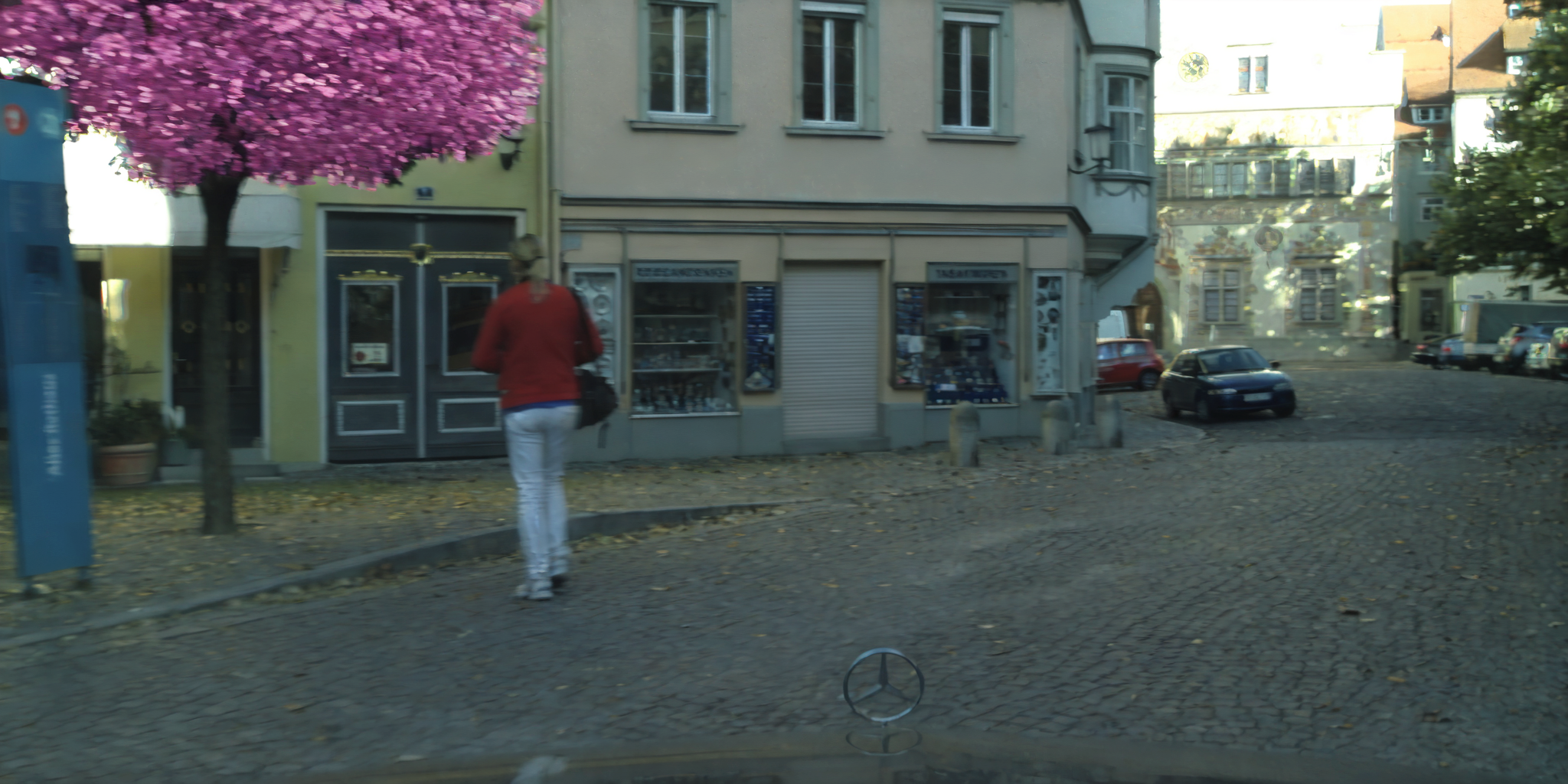}{0.65}{0.42}{0.15} &
\circledincludegraphics{\cityfigwidth}{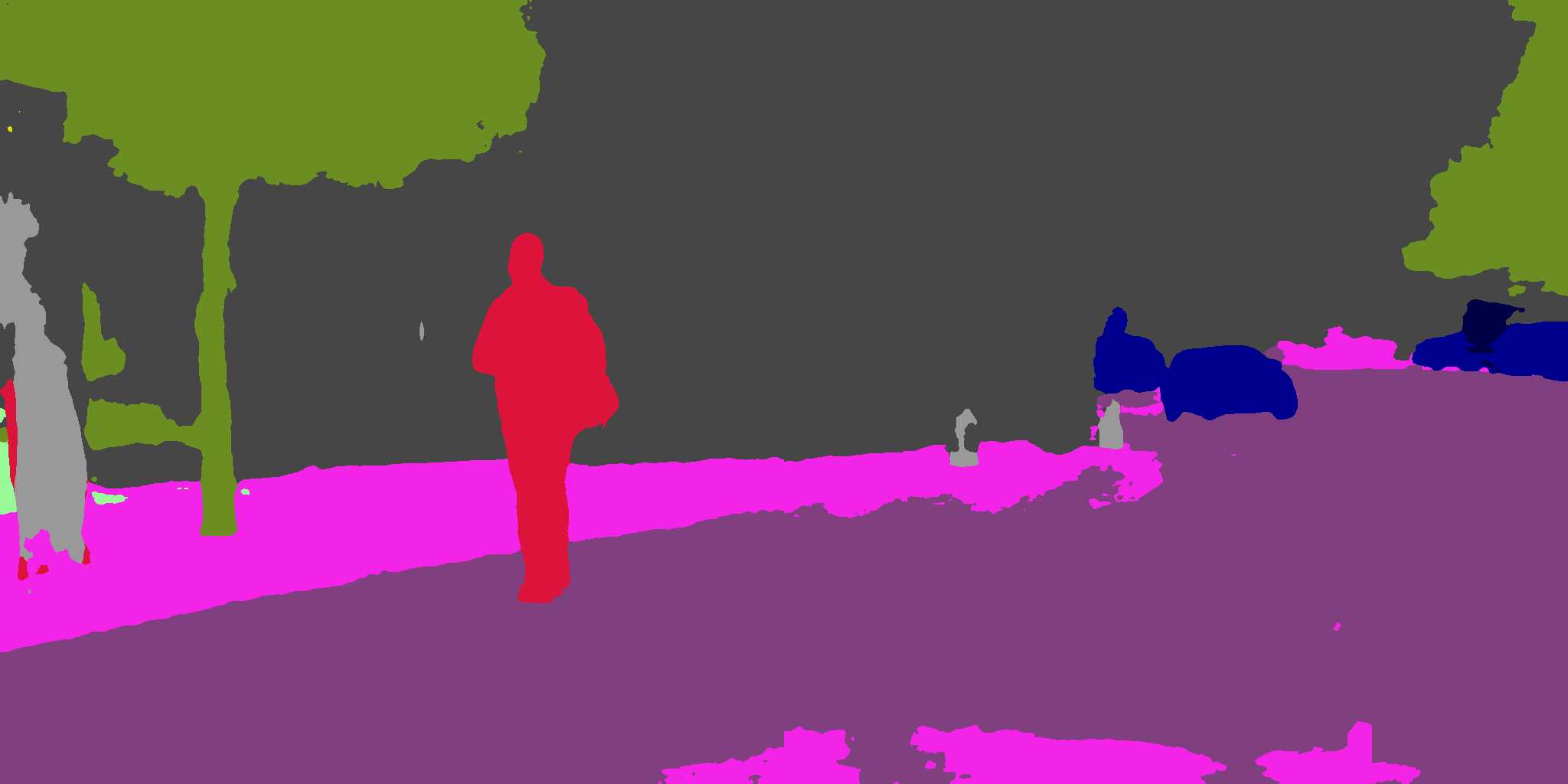}{0.65}{0.42}{0.15} &
\circledincludegraphics{\cityfigwidth}{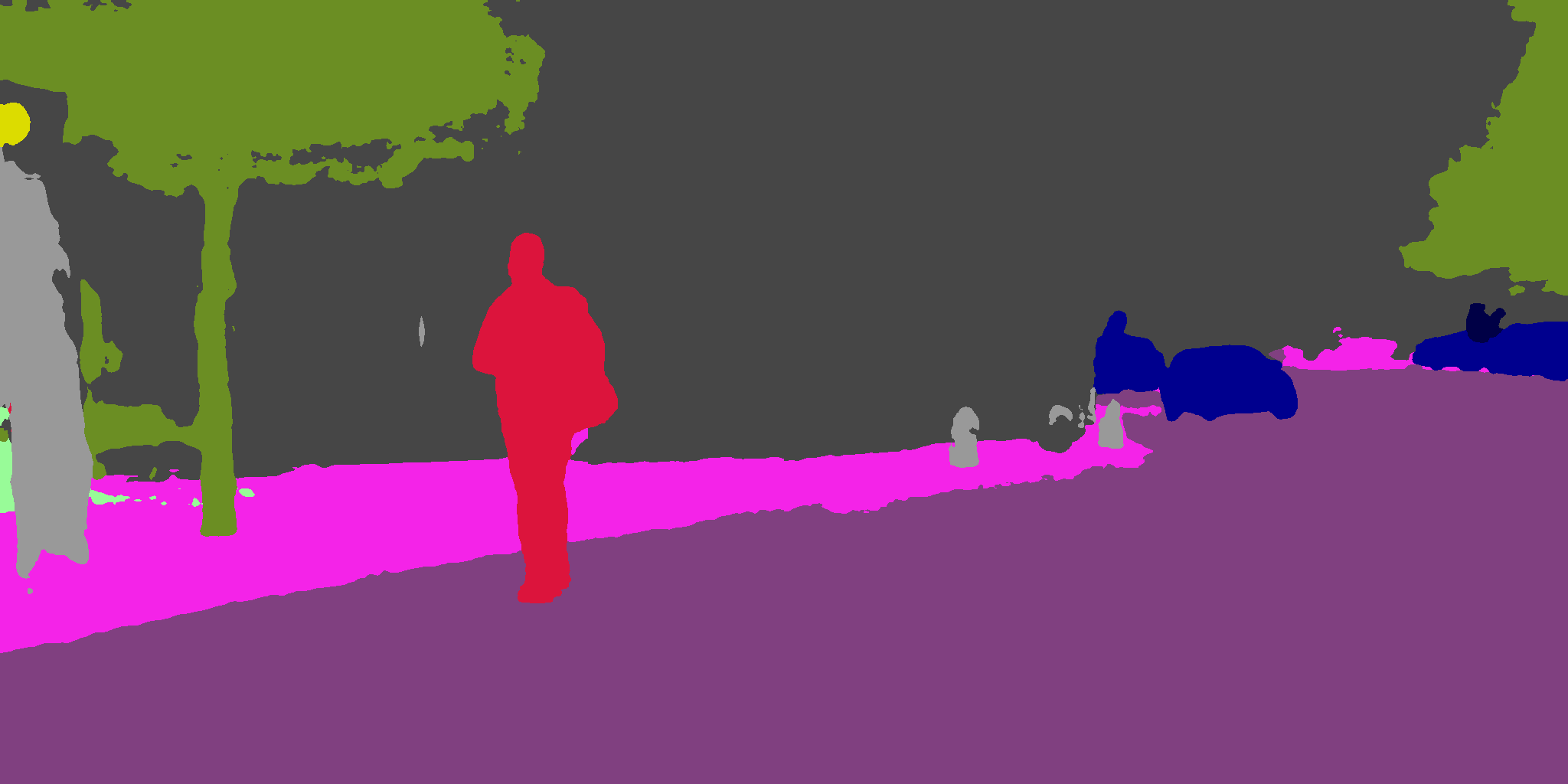}{0.65}{0.42}{0.15} &
\circledincludegraphics{\cityfigwidth}{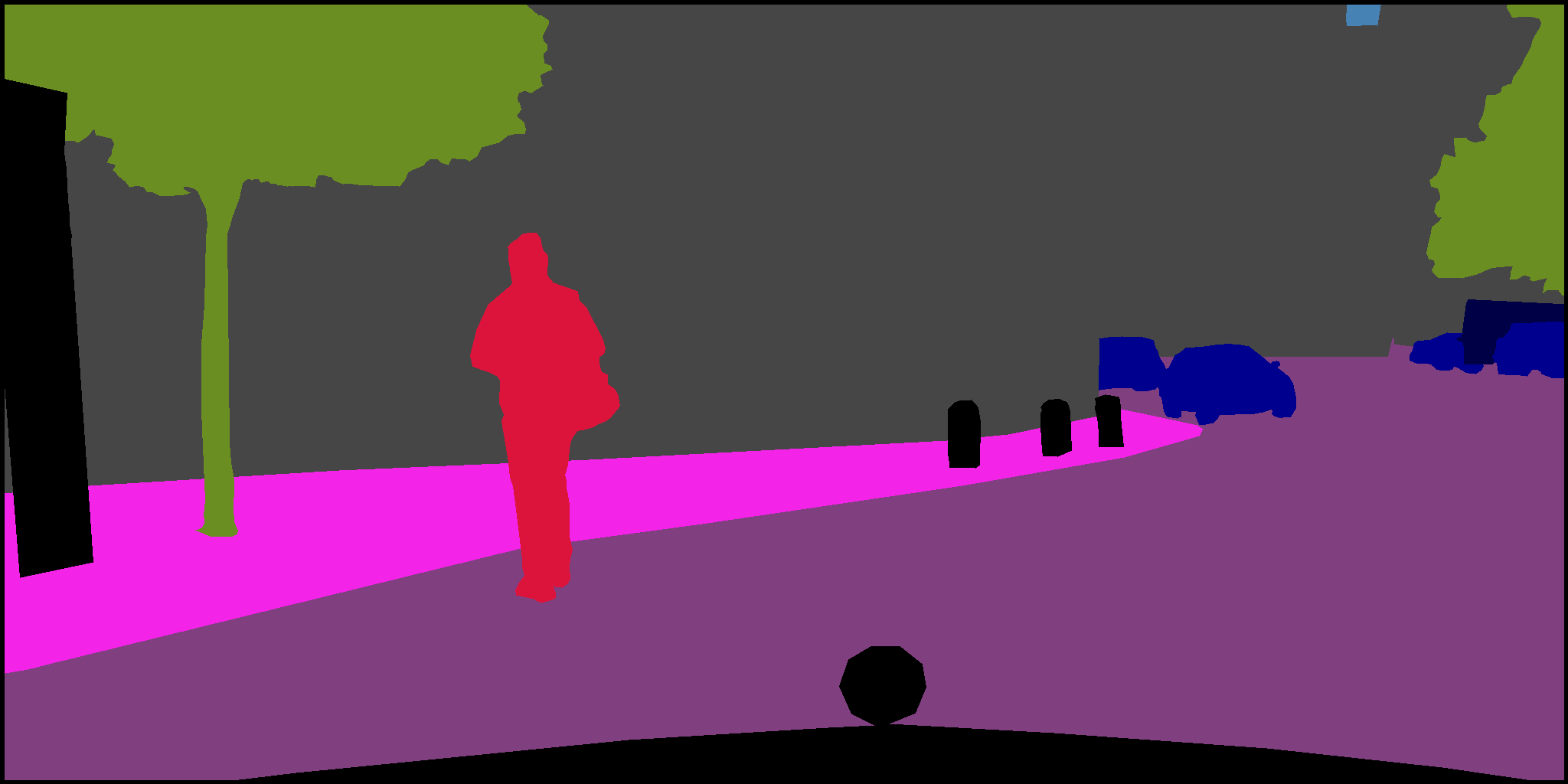}{0.65}{0.42}{0.15} \\

\circledincludegraphics{\cityfigwidth}{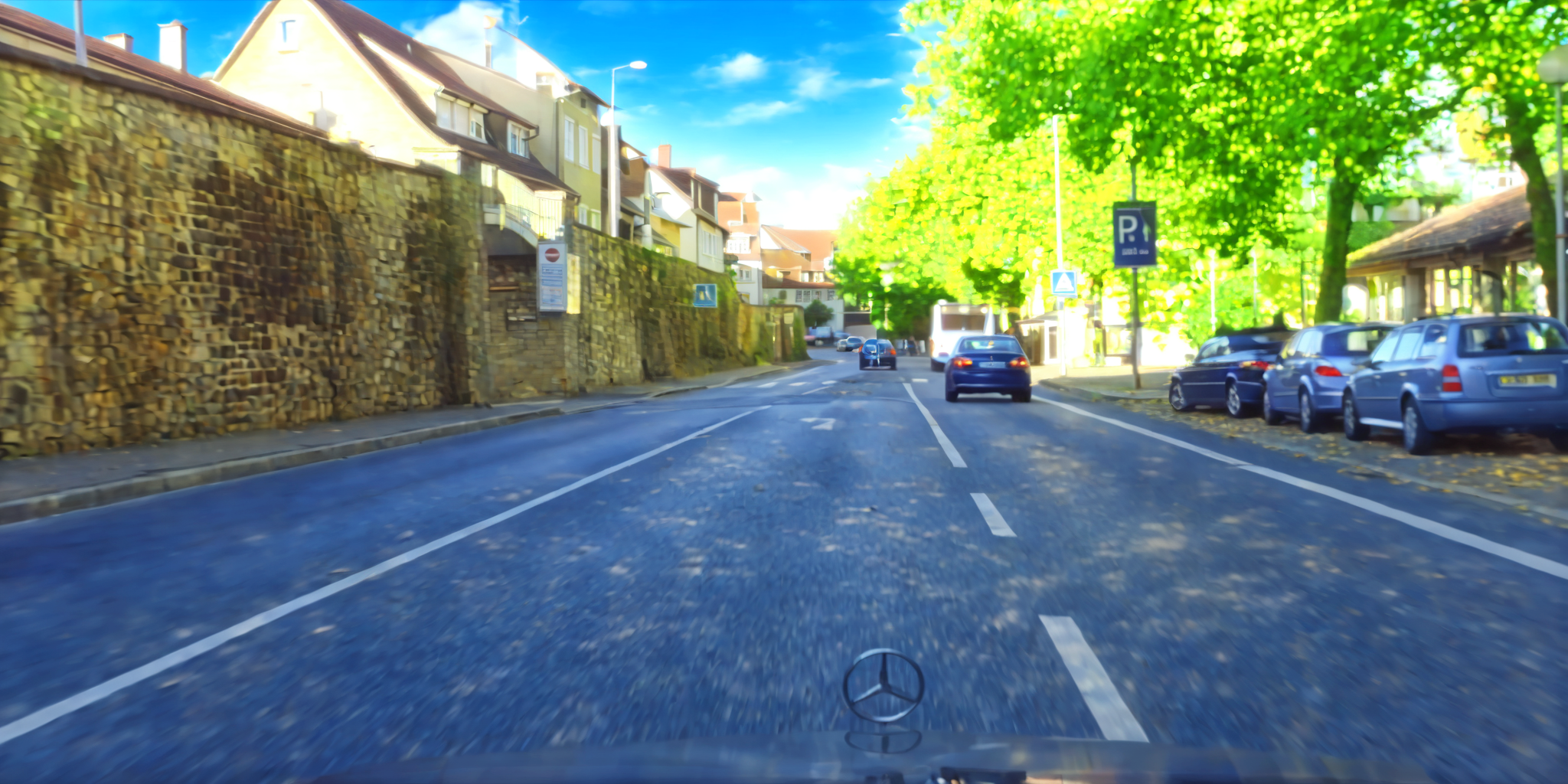}{0.25}{0.7}{0.25} &
\circledincludegraphics{\cityfigwidth}{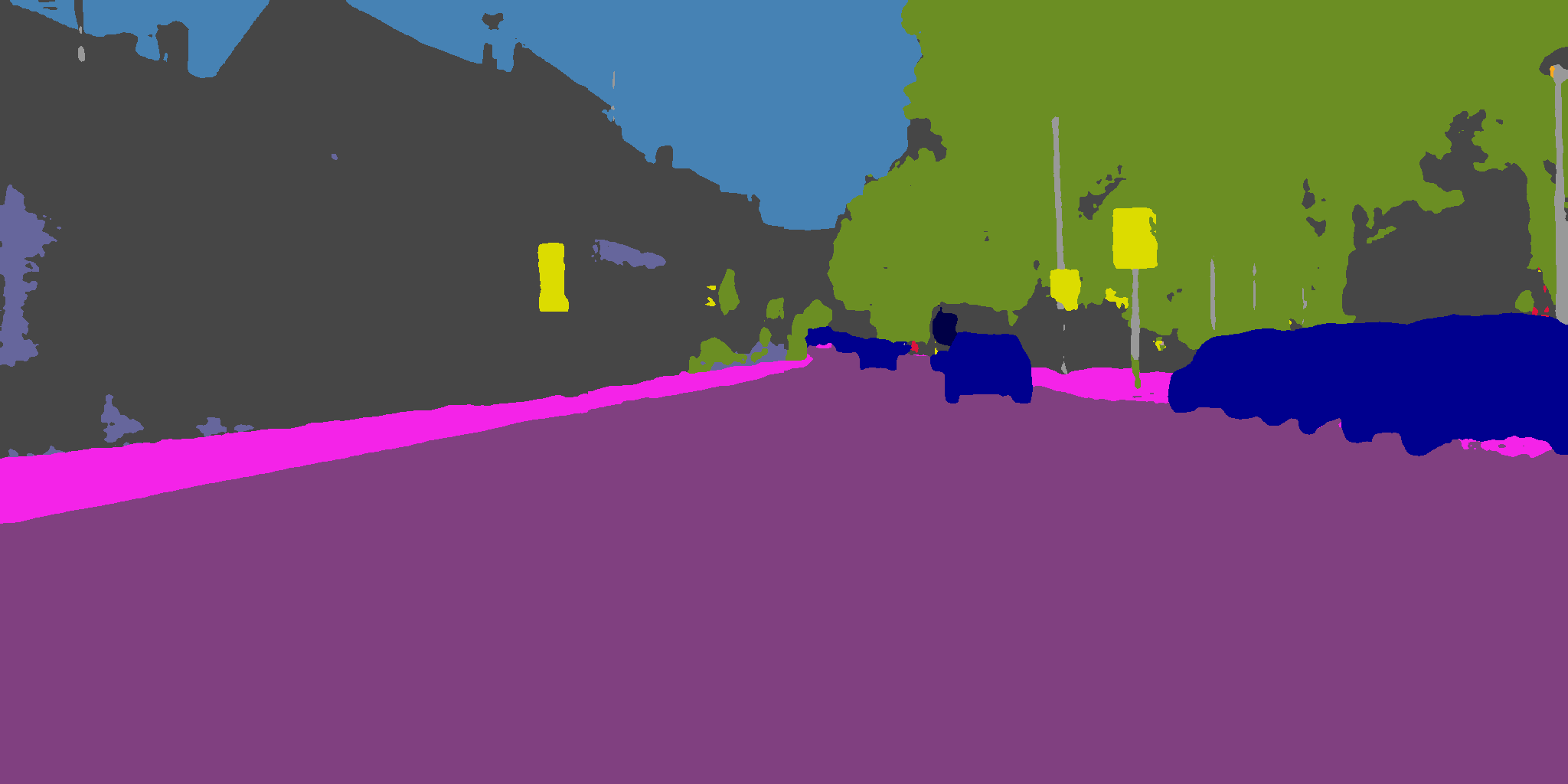}{0.25}{0.7}{0.25} &
\circledincludegraphics{\cityfigwidth}{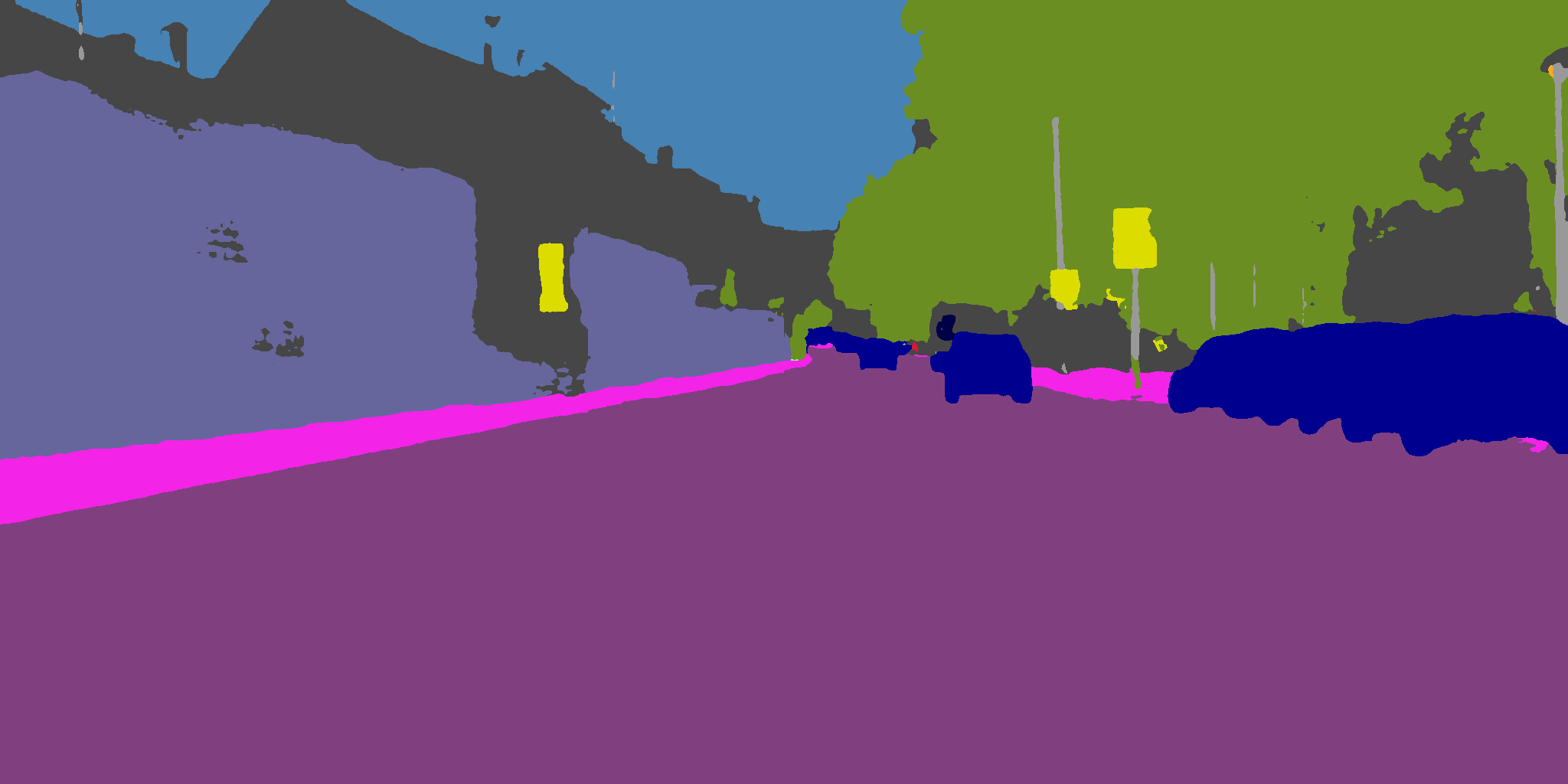}{0.25}{0.7}{0.25} &
\circledincludegraphics{\cityfigwidth}{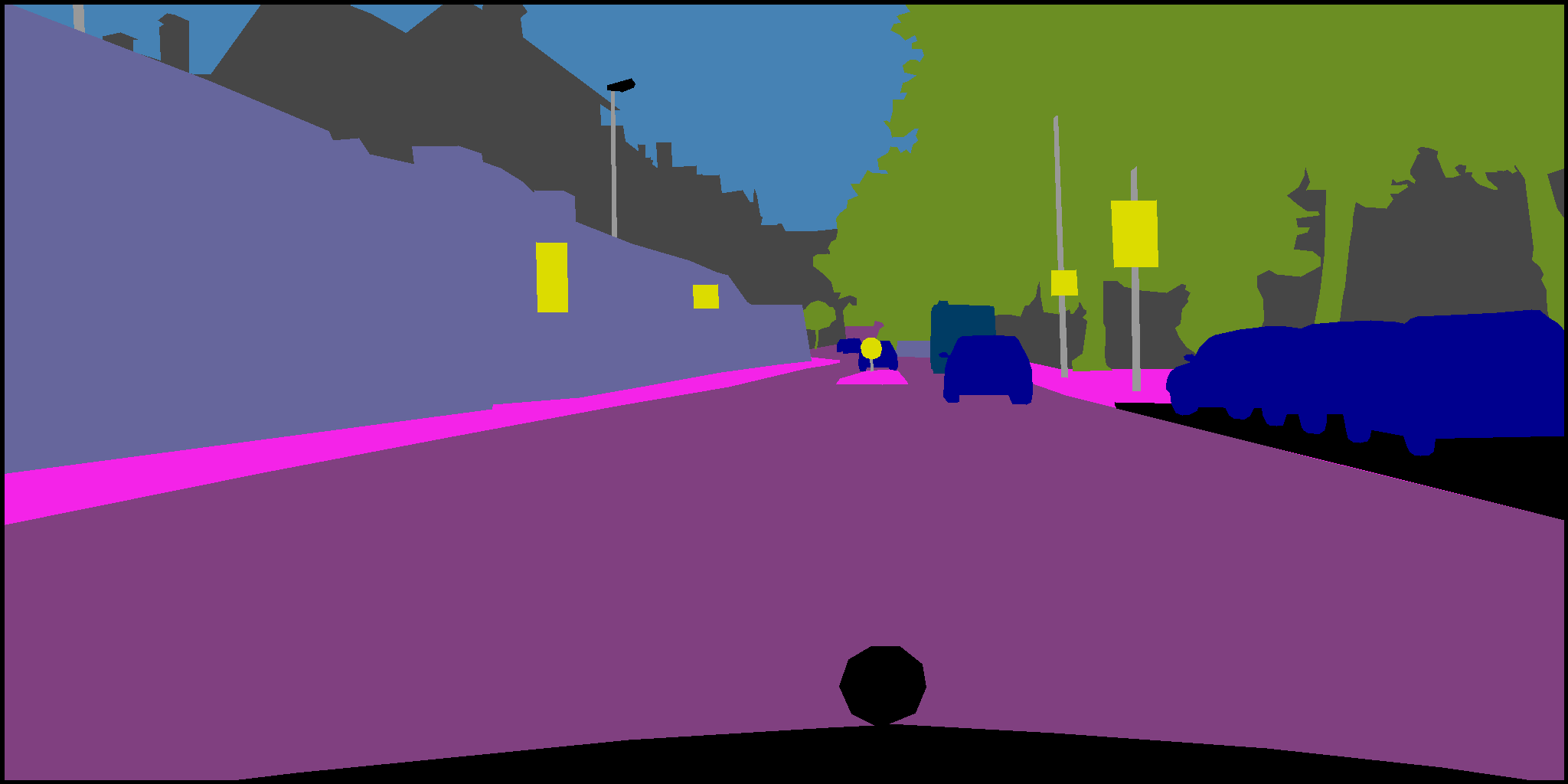}{0.25}{0.7}{0.25} \\
\circledincludegraphics{\cityfigwidth}{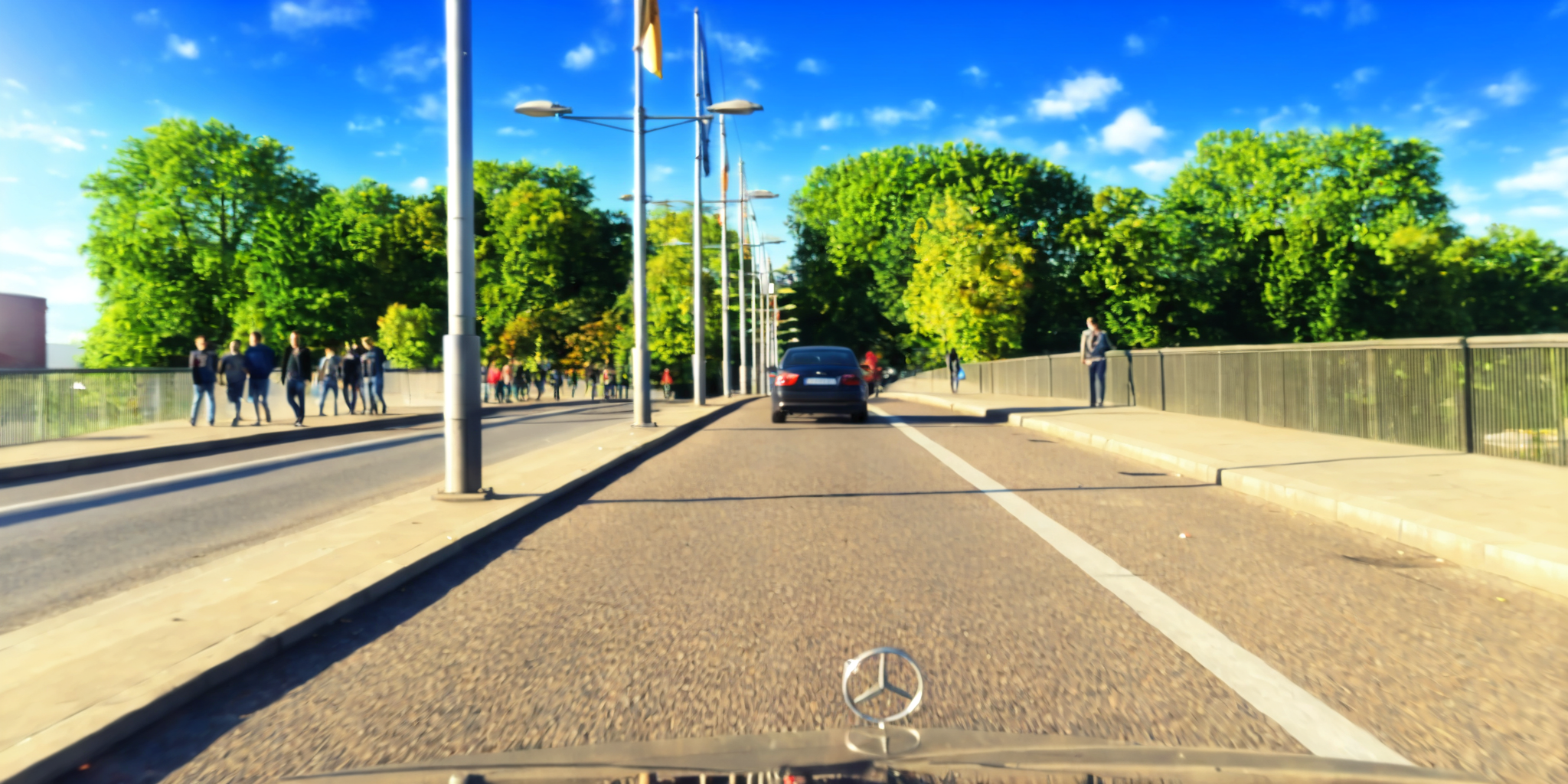}{0.2}{0.2}{0.2} &
\circledincludegraphics{\cityfigwidth}{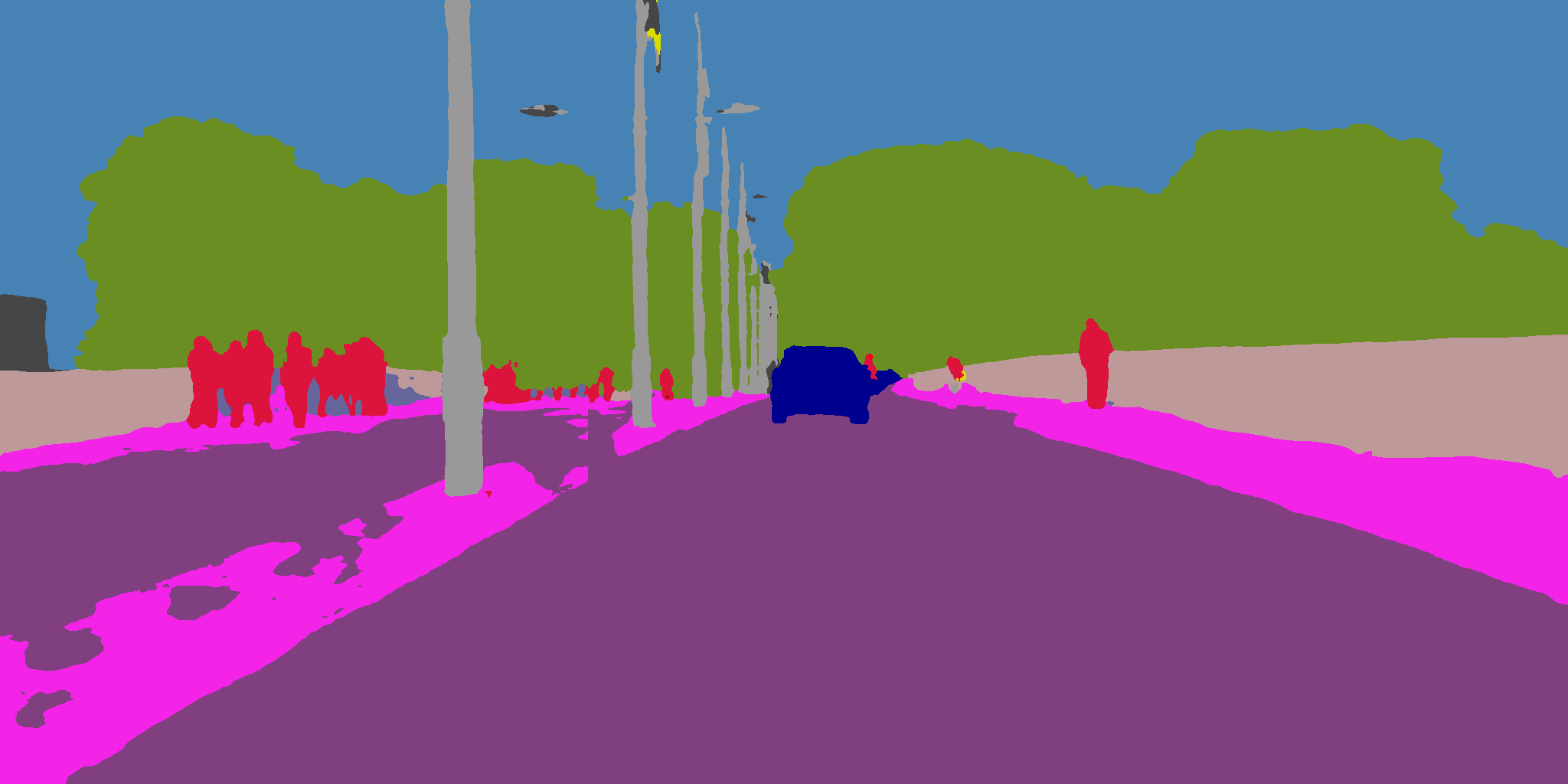}{0.2}{0.2}{0.2} &
\circledincludegraphics{\cityfigwidth}{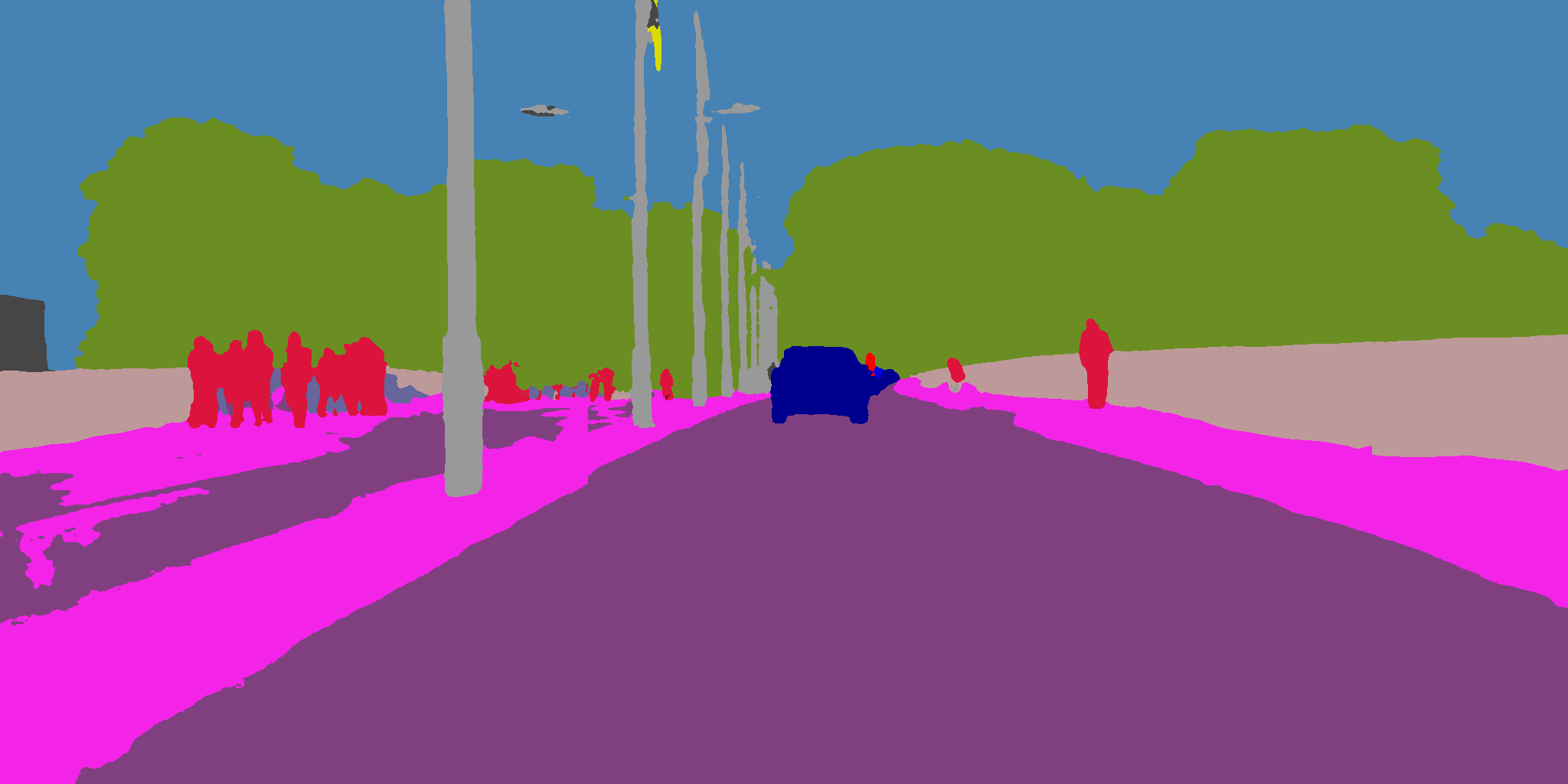}{0.2}{0.2}{0.2} &
\circledincludegraphics{\cityfigwidth}{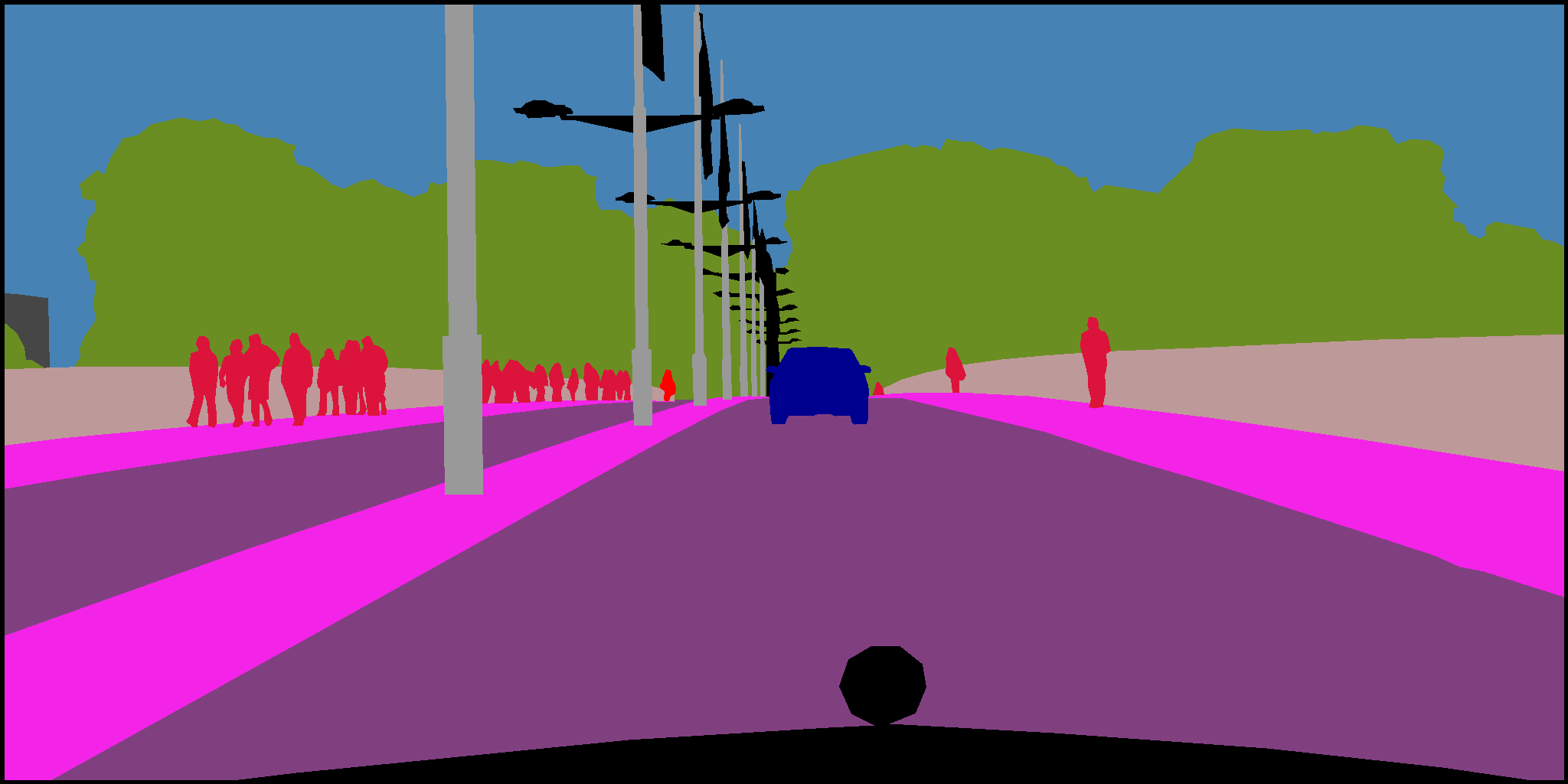}{0.2}{0.2}{0.2} \\

\end{tabular} 

\caption{
Qualitative results on Cityscapes Adverse. Details of interest are highlighted with yellow circles.
}
\label{fig:supplementary_segmentation}
\end{figure*}

\section{Limitations and Future Work} 
\label{sec:limitations}


While \method{} shows strong empirical performance, our study also has limitations that suggest natural directions for future research. Our experiments mainly target appearance shifts such as day-to-night and related changes in illumination and weather. We also observe gains in several adverse conditions, but there are regimes where the event signal can become noisy or less informative (for example, under rain or flickering neon lights), which may reduce the benefit of event-based supervision. In such cases, the robustness of \method{} is inherently bounded by the quality of the event stream.

Moreover, we currently treat events as the only privileged modality. This is a natural choice in scenarios where events remain relatively stable under domain shift, but it may be suboptimal when the event sensor itself is strongly affected by domain shift. 
In those settings, other sensing modalities (such as thermal imaging) or combinations of multiple privileged signals could provide a more reliable training signal than events alone. In addition, our implementation relies on a relatively simple frame-based representation of events. While this keeps the framework easy to integrate into existing architectures, it may not fully exploit the fine-grained spatio-temporal structure of the event stream. Alternative encodings, such as voxel grids or point-cloud–like representations, could provide richer supervision signals and further strengthen the predictive regularization effect.

These considerations point to several promising research directions. One avenue is to extend our predictive framework beyond domain generalization to settings such as (semi-)supervised domain adaptation and unsupervised domain adaptation, where unlabeled or sparsely labeled target data are available during training. 
Another is to explore multi-modal privileged supervision, leveraging multiple predictive targets (e.g., events and thermal) that can compensate for each other in challenging conditions where a single modality is unreliable. Finally, investigating more expressive event representations within our predictive framework is an interesting direction for better exploiting the temporal dynamics of event cameras in real-world deployments.

\end{document}